\documentclass[letterpaper,12pt]{report}	

\usepackage[authoryear, round]{natbib}
\usepackage{styles/utformat}  		

\usepackage{pdflscape}
\usepackage{acronym}

\usepackage{graphicx}
\usepackage{amsmath,amsfonts,amscd}
\usepackage{amsthm}
\usepackage{eucal} 	 	
\usepackage{verbatim}      	
\usepackage{styles/citesort}         	%
\usepackage{algorithm}
\usepackage{bibentry}
\usepackage{algorithmic}
\usepackage{xcolor}
\usepackage{markdown}
\usepackage{booktabs}
\usepackage{url}
\usepackage{svg}
\usepackage{tikz}
\usepackage{multirow}
\usepackage{makecell}
\usepackage{tabularx}
\usepackage{bbm}
\usepackage{mathtools}
\usepackage{hyperref}
\usepackage{wrapfig}
\hypersetup{colorlinks=false}
\usepackage{graphicx} 
\usepackage{subcaption} 
\usepackage{enumitem}
\usepackage{rotating}
\usepackage[htt]{hyphenat}
\usepackage{pgfplots}
\usepackage{tablefootnote}

\DeclareMathOperator*{\argmax}{argmax}
\DeclareMathOperator*{\argmin}{argmin}
  
\oneandonehalfspacing

\theoremstyle{definition}

\theoremstyle{remark}

\newcommand{\latexe}{{\LaTeX\kern.125em2%
                      \lower.5ex\hbox{$\varepsilon$}}}

\chardef\bslash=`\\	

\makeatletter		
\def\square{\RIfM@\bgroup\else$\bgroup\aftergroup$\fi
  \vcenter{\hrule\hbox{\vrule\@height.6em\kern.6em\vrule}%
                                              \hrule}\egroup}
\makeatother		

\nobibliography*

\author{Haresh Karnan}  	

\address{haresh.miriyala@utexas.edu}  

\title{Aligning Robot Navigation Behaviors with \\ Human Intentions and Preferences} 

\supervisor[Ashish Deshpande]
	{Peter Stone}

\committeemembers
	[Joydeep Biswas]
	[Garrett Warnell]
	[Anca Dragan]
    [Farshid Alambeigi]
    {Junmin Wang} 

\degree{Doctor of Philosophy}
\degreeabbr{PhD} 

\dissertation  

\graduationmonth{May}
\graduationyear{2024}

\begin{document}


\newcommand{\haresh}[1]{\textcolor{blue}{[Haresh: #1]}}
\newcommand{\xuesu}[1]{\textcolor{violet}{[Xuesu: #1]}}
\newcommand{\peter}[1]{\textcolor{green}{[Peter: #1]}}
\newcommand{\garrett}[1]{\textcolor{red}{[Garrett: #1]}}
\newcommand{\TODO}[1]{\textcolor{red}{[TODO: #1]}}

\newcommand{\xset}{\ensuremath{\mathcal{X}}}    

\newcommand{\gaifo}{\textsc{gai}f\textsc{o}}
\newcommand{\garat}{\textsc{garat}}
\newcommand{\gail}{\textsc{gail}}
\newcommand{\gat}{\textsc{gat}}
\newcommand{\ifo}{\textsc{i}f\textsc{o}}
\newcommand{\voila}{\textsc{voila}}
\newcommand{\vtr}{\textsc{vt}\&\textsc{r}}
\newcommand{\rae}{\textsc{rae}}
\newcommand{\vrlpap}{\textsc{vrl}-\textsc{pap}}
\newcommand{\scand}{\textsc{scand}}
\newcommand{\irl}{\textsc{irl}}
\newcommand{\rl}{\textsc{rl}}
\newcommand{\bc}{\textsc{bc}}
\newcommand{\lfd}{\textsc{l}f\textsc{d}}
\newcommand{\eg}{\textit{e.g.,}}
\newcommand{\movebase}{\texttt{move\_base}}
\newcommand{\virtue}{\textsc{virtue}}
\newcommand{\medirl}{\textsc{medirl}}
\newcommand{\vicreg}{\textsc{vic}reg}
\newcommand{\naturl}{\textsc{naturl}}
\newcommand{\sterling}{\textsc{sterling}}
\newcommand{\patern}{\textsc{patern}}
\newcommand{\lfp}{\textsc{lfp}}
\newcommand{\ganav}{\textsc{ganav}}
\newcommand{\rellis}{\textsc{rellis}}
\newcommand{\rugd}{\textsc{rugd}}
\newcommand{\rca}{\textsc{rca}}
\newcommand{\ser}{\textsc{ser}}
\newcommand{\psd}{\textsc{psd}}
\newcommand{\vi}{\textsc{vi}}
\newcommand{\mm}{\textsc{mm}}

\copyrightpage  

\commcertpage   

\titlepage      

\begin{dedication}		
\vspace*{187pt}
\begin{center}
To my dearest parents, Karnan and Sumathy; my loving wife, Gayatri;\\
and my dear brother, Dinesh.
\end{center}
\vspace*{\fill}
\end{dedication}



\begin{acknowledgments}		

I'm immensely grateful to several people who have helped me get where I am today. First, I begin by thanking my advisor, Peter Stone. Despite his several commitments, Peter has always found the time to advise me in my research journey. His immense interest in AI and Robotics and enthusiasm for exploring research ideas have helped propel my research journey to the fullest during my Ph.D., and I could not be more grateful and proud to have him as my advisor. I would also like to thank Garrett Warnell, who has been another pillar of great support throughout my PhD. My weekly interactions with Garrett and his thoughtful guidance on research have helped shape me into a better researcher.  I would also like to thank all of my dissertation committee, Ashish Deshpande, Joydeep Biswas, Junmin Wang, Farshid Alambeigi, and Anca Dragan, for their support and being a part of my journey. 

Reflecting on my journey, I vividly recall the day in the Fall of 2017 when I first came across Peter's website, sending him an email wishing to be a part of the Learning Agents Research Group. Fast forward, I'm immensely proud to call myself a member of this group at UT Austin, and I'm grateful for all the personal and professional growth I've had during my time here. I'm grateful to Josiah Hanna and Siddharth Desai, my first collaborators at LARG who were warm and welcoming, introducing me to sim-to-real research, which was immensely rewarding to me. I would also like to thank my RoboCup$@$Home team at UT, namely Rishi Shah, Justin Hart, Yuqian Jiang, Gilberto Briscoe-Martinez, and Dominick Mulder for an enriching RoboCup experience. I would like to thank everyone from the Learning Agents Research Group family, especially, Ishan Durugkar, Harel Yedidsion, Anirudh Nair, Reuth Mirsky, Xuesu Xiao, Yoonchang Sung, Chen Tang, Faraz Torabi, Sanmit Narvekar, Elad Liebman, Patrick MacAlpine, Piyush Khandelwal, Jinsoo Park, Eddy Hudson, Bo Liu, Yifeng Zhu, Jiaxun Cui, Michael Munje, Rolando Fernandez, Caroline Wang, Zizhao Wang, Zifan Xu,  Jiaheng Hu, Siddhant Agarwal, Bharath Masetty, Brad Knox, Brahma Pavse, William Macke, Sai Kiran Narayanaswami, and Shahaf Shperberg. I would also like to thank all members of the Autonomous Mobile Robotics Laboratory, another family I'm immensely proud to be a part of and to have conducted research with, namely, Joydeep Biswas, Elvin Yang, Daniel Farkash, Rohan Chandra, Zichao Hu, Arthur Zhang, Sadegh Rabiee, Jarrett Holtz, Luisa Mao, Kavan Sikand, Corrie Van Sice, Amanda, Noah, Sadanand Modak, Rahul Menon, and Max Svetlik. 

In particular, I would like to thank my amazing collaborators and friends at UT Austin: Siddharth Desai, Garrett Warnell, Josiah Hanna, Ishan Durugkar, Prasoon Goyal, Justin Hart, Elvin Yang, Daniel Farkash, Kavan Sikand, Eddy Hudson, Faraz Torabi, Yuqian Jiang, Luisa Mao, Zichao Hu, Rohan Chandra, and Xuesu Xiao. I would like to thank my former ``RLHouse"mates Prasoon Goyal, Ishan Durugkar, and Sandhya for all the board game sessions and spicy Indian food we relished! I also wish to thank my incredible friends Sanjana Rajendran, Sravan Devanathan, Shilpa, Neha Dipali, Mukesh, Eddy, Anup, Madhuri, Chandana, Nithin, Mythreyi, Pritesh, Karthikeya, Manish, and Karishma for making my time in Austin worth cherishing.

I kickstarted my journey in robotics during my undergraduate days with the Robotics and Machine Intelligence club at NIT Trichy, and I would like to thank the club, and my dearest friends Akshay P Roy, Surya Teja Golkonda, Hariharan, Keshav Rai Goud, Prajval Kumar, Pranav Sundaram, Sripad, Adarsh Jagan Krishnamurthy, and Prakash Baskaran for being a part of that journey. 

Finally, I would like to extend my deepest gratitude to my family. A special note of thanks goes to my partner and best friend, Gayatri, who has been an unshakeable foundation of strength, courage, and support in my life, uplifting my spirits through every challenge we have encountered together. I also thank her family, Sivaraman, Rajeswari, Harisree, and Ganesh for their encouragement and support. My heartfelt thanks to my parents, Karnan and Sumathy, my brother, Dinesh, my late grandfather Gopalakrishnan, and my grandmother Sakunthala, for their boundless love and support. Furthermore, I am immensely thankful to my closest friends Sravan, Sanjana, Mukesh, Keshav, Suraj, Pranav, Somya, Prajval, Ashwin, Hani, Surya, and Garan for their encouragement and love. Without their collective support, I could not have achieved this milestone.

\begin{flushright}
\large Haresh Karnan
\end{flushright}

\begin{flushleft}
The University of Texas at Austin \\
May 2024
\end{flushleft}

\addcontentsline{toc}{chapter}{Acknowledgments}
\end{acknowledgments}


%
\clearpage
\utabstract
\phantomsection
\addcontentsline{toc}{chapter}{Abstract}

Recent advances in the field of machine learning have led to new ways for mobile robots to acquire advanced navigational capabilities \citep{bojarskie2e, kahnssldeeprlnav, kendall2019learning, byronboots, silver2010learning}. However, these learning-based methods raise the possibility that learned navigation behaviors may not align with the intentions and preferences of people, also known as \textit{value misalignment}. To mitigate this danger, this dissertation aims to answer the question ``How can we use machine learning methods to align the navigational behaviors of autonomous mobile robots with human \textit{intentions} and \textit{preferences}?" 

First, this dissertation answers this question by introducing a new approach to learning navigation behaviors by imitating human-provided demonstrations of the intended navigation task. This contribution allows mobile robots to acquire autonomous visual navigation capabilities through imitating human demonstrations using a novel objective function that encourages the agent to align with the navigation objective of the human and penalizes for misalignment. Second, this dissertation introduces two algorithms to enhance terrain-aware off-road navigation for mobile robots through learning visual terrain awareness in a self-supervised manner. This contribution enables mobile robots to obey a human operator's preference for navigating over different terrains in urban outdoor environments and extrapolate these preferences to visually novel terrains by leveraging multi-modal representations.
Finally, in the context of robot navigation in human-occupied environments, this dissertation introduces a dataset and an algorithm for robot navigation in a socially compliant manner in both indoor and outdoor environments. In summary, the contributions in this dissertation\footnote{A recording of the defense talk can be accessed here: \href{https://youtu.be/MssiO6g0Gb8}{https://youtu.be/MssiO6g0Gb8}} take a significant step towards addressing the value alignment problem in autonomous navigation, enabling mobile robots to navigate autonomously with objectives that align with the intentions and preferences of humans.    

\tableofcontents   

\listoftables      
\listoffigures     

%
%

\chapter{Introduction}

Advances in the field of machine learning and artificial intelligence for robotics have enabled mobile robots to become increasingly intelligent, efficient, and autonomous \citep{bojarskie2e, kahnssldeeprlnav, kendall2019learning, byronboots, silver2010learning}. Coupled with a leap in compute capabilities, especially of hardware accelerators enabling deep learning-based methods to learn from huge amounts of data \citep{epoch2022trendsingpupriceperformance}, autonomous mobile robots have gained improved capabilities such as real-time perception \citep{redmon2016yolo, maskrcnn, clip}, planning \citep{spatialplanningtransformers, vin}, and control \citep{bojarski, byronboots, viikd, xuesusurvey, xuesusurveymotionplanning,helicopterrl}. These advances in data-driven algorithms, combined with improved hardware and availability of large datasets \citep{barndataset, scand, biswas2013longterm} have contributed to increasingly intelligent and efficient autonomous mobile robotic agents \citep{Amazonscout,khandelwal2017bwibots}. 

As these agents become more capable, there is a danger that their internal objectives may result in behaviors that do not align with the intentions and preferences of humans, also known as \textit{value misalignment}.\footnote{Note that throughout this dissertation, we also refer to ``alignment in behavior'' which means the same as \textit{value alignment}.} A well-known story related to value misalignment dates to King Midas from Greek mythology, who wished that everything he touched would be turned into gold, but failed to anticipate that food and drink touched by him for consumption also would turn into gold, making it a regrettable wish that did not align with his intentions. Similarly, in robotics, value misalignment has been observed as a consequence of the effects of \textit{reward misdesign} \citep{knox2022reward} and \textit{reward hacking} \citep{hadfield2017inverse}; for example, a vacuum cleaning robot rewarded for collecting dust might begin to eject collected dust to accumulate more rewards \citep{russell2003p}. Several other instances of value misalignment in robotics have been observed in prior work \citep{amodei-safety, everitt_safe_ai, reward-hacking-deepmind, lehmann_etal}. 
Therefore, it is necessary to develop new methods that ensure that the behavior of autonomous driving agents aligns with the intentions and preferences of their human operators or owners. 

This dissertation is especially concerned with the problem of autonomous robot navigation, i.e., to enable a robot to navigate autonomously, with minimal or no human supervision during deployment in ways that align with the intentions and preferences of people. Classical heuristic-based approaches \citep{dijkstra, a_star, xuesusurveymotionplanning} to this problem use hand-designed cost functions, planners, and kinodynamic controllers; and work well in structured environments such as industries and warehouses that are fully mapped, standardized for robot usage, and have minimal human presence. However, these approaches for navigation often struggle in unstructured, real-world conditions, including---but not limited to---navigation in unmapped novel environments, navigation on non-standard off-road terrains, and safe navigation in the presence of people.  

To overcome the limitations of classical approaches to navigation, many in the research community have started using tools from machine learning. Unlike classical approaches that are heuristic-based and designed for specific environments \citep{xuesusurveymotionplanning}, learning-based approaches offer the promise of learning patterns from data that may help the robot continually reason and adapt \citep{bojarski, byronboots, alvinn, tai2018socially, xiao2020appld}. However, the success of learning-based approaches is dependent on designing the right objective function for navigation that reflects the human-intended task, since a poorly-defined objective function could lead to unintended (misaligned) behaviors \citep{amodei-safety, everitt_safe_ai, reward-hacking-deepmind, lehmann_etal}. For example, a mobile robot tasked with reaching a goal location outdoors in a collision-free route may choose to trample over a bed of flowers, which may be undesirable for its human operator. In an alternate scenario, when tasked with navigating in the presence of people while avoiding collisions, a mobile robot may choose to impolitely cut through groups of people, or move dangerously close to a human,  behaviors that most humans would avoid. Conversely, at the other extreme, these robots may become overly cautious, leading to situations where they stop entirely and cease to make further progress, especially in complex or unpredictable environments. When mobile robots navigate in unexpected ways that are not generally preferred by humans, it may sometimes lead to catastrophic events, such as accidents with elderly people, visually impaired people, or children. Such public safety concerns in the past have led to autonomous mobile robots being banned from public roads and sidewalks \citep{Guardian2017DeliveryRobots, Sidewalkrobotban}.

To address the challenge of value alignment in autonomous navigation, the method of \textit{Learning from Demonstration} (\lfd{})---as discussed by \cite{argall2009survey} and \cite{chernovail}---emerges as a powerful approach. By inferring either the task objective or an imitating policy from such demonstrations, \lfd{} enables the robot's navigation behaviors to align with the intentions and preferences of the human operator. Alignment through \lfd{} is predicated on the assumption that human demonstrations themselves contain such ``aligned" behavior that reflects the intentions and preferences of the demonstrator. Given such a demonstration, one can infer the objective function of the intended task through inverse reinforcement learning (\irl{}) \citep{andrewng_irl, meirl}, or recover the underlying navigation policy through behavior cloning (\bc{}) \citep{behaviorcloning_framework, behaviorcloning_no_regret_learning, bco, xiao2020appld} or reinforcement learning (\irl{}+\textsc{rl}) \citep{finn2016connection, ho2016generative}.


Learning from Demonstrations (\lfd{}) employs expert human task demonstrations to facilitate imitation learning of robot behaviors. However, alternative forms of human feedback also exist and have been instrumental in teaching robots to behave in ways that resonate with human intentions and preferences. For instance, \cite{wang2021apple} explored learning from evaluative feedback, a less demanding form of human input compared to task demonstrations. Other feedback types have also been explored in the literature, such as interventions \citep{wang2021appli} and disengagements \citep{kahn2021land}. In this dissertation, I explore \textit{Learning from Preferences} (\lfp{}), another paradigm utilizing preference rankings as the feedback mechanism, to achieve operator terrain preference-aligned navigation. In \lfp{}, an operator reviews and ranks instances from the robot's experience based on their preferences. Utilizing these preference rankings, \lfp{} either learns an objective function followed by a policy \citep{browntrex, bobu2020less} or directly deduces a policy \citep{directpreferenceoptimization} that aligns with the operator's specified preferences.

Building on these insights, the primary aim of this dissertation is to address the value alignment problem in robot navigation by leveraging human feedback through task demonstrations and preference queries to improve navigation in unstructured static and dynamic environments. Specifically, in static environments, this dissertation introduces algorithms that enable autonomous navigation in unmapped indoor and outdoor environments, while adhering to human intentions and preferences. For dynamic environments, a dataset and algorithms are introduced to enable a mobile robot to navigate in a \textit{socially compliant} manner, while recognizing and reacting to the objectives of humans in the scene, also known as \textit{socially compliant navigation}.  Concisely, this dissertation will investigate the following question: 

\vspace{0.35cm}
\noindent\fbox{%
    \parbox{\textwidth}{%
       How can machine learning methods be applied to the task of autonomous navigation in unstructured environments such that the learned navigation behaviors of mobile robots align with the intentions and preferences of humans?
    }%
}


\section{Contributions}
\label{sec:first_contribs}

\noindent This thesis answers the stated question through contributions on three key topics:

\subsubsection*{1. Visual Imitation Learning for Robot Navigation}
\label{contrib1}
 Towards learning navigation policies on a mobile robot that align in behavior as demonstrated by a physically different agent such as a human operator, this dissertation introduces an imitation learning algorithm called Visual Observation-only Imitation Learning for Autonomous navigation (\voila{}) \citep{voila}. \voila{} introduces a novel reward function that overcomes a significant limitation of existing visual imitation learning algorithms---learning using video-only demonstrations from a physically different agent in the presence of egocentric viewpoint mismatch. \voila{} was found to be successful at learning visual navigation policies end-to-end (mapping from raw sensor observations directly to low-level action commands) from human demonstrations in both simulated and physical robot experiments, that align in behavior with the human demonstrations. Chapter \ref{chap:imitation_learning_for_robot_nav} discusses this contribution in detail.

\subsubsection*{2. Preference-Aligned Off-Road Navigation}
\label{contrib2}

Aligning a robot's navigation behavior with an operator's terrain preferences is a major challenge in visual off-road navigation. Prior approaches require extensive human labeling efforts, which are expensive and may not be scalable. To address this problem, this dissertation introduces two novel algorithms. 

First, this thesis introduces Self-supervised TErrain Representation LearnING from unconstrained robot experience (\sterling{}) \citep{sterling}, a self-supervised terrain representation learning algorithm that learns relevant terrain representations from easy-to-collect, unconstrained, and multimodal robot experience collected with any navigation policy. Following a two-step procedure, \sterling{} learns relevant terrain representations in a self-supervised manner followed by operator preference querying, used to identify an operator's preferences over traversed terrain examples. Given such preferences, a preference utility function can be learned, which can be used in real-time for preference-aligned off-road navigation on the robot. Through physical robot experiments in diverse outdoor environments, we find that \sterling{} leads to navigation behaviors that are in alignment with an operator's terrain preferences.

Second, this dissertation introduces an extension of \sterling{} to extrapolate operator preferences for visually novel terrains. More often than not, a mobile robot might encounter novel terrains outside the training distribution for which the operator's terrain preference is unknown, causing the robot to deal with uncertain situations. In certain cases, however, although a terrain looks visually novel, the proprioceptive feedback of the robot's interaction with the terrain might feel similar to a previously traversed terrain. Leveraging this intuition, this dissertation introduces Preference extrApolation for Terrain awarE Robot Navigation (\patern{}) \citep{patern} which extends the capabilities of \sterling{} by extrapolating operator preferences from known to visually novel terrains. \patern{} enhances the robot's adaptability to unseen terrains, further aligning its navigation behaviors with human preferences. The methodology and effectiveness of \patern{} are comprehensively discussed in Chapter \ref{chap:pref_learning_offroad_nav}, alongside \sterling{}.

\subsubsection*{3. Socially Compliant Robot Navigation}
\label{contrib3}

While the above contributions advance the state of the art in aligning a robot's navigation behavior with human \textit{intentions} and \textit{preferences}, the environments considered above are mostly static. However, an autonomous mobile robot deployed in urban environments might also need to deal with dynamic objects such as pedestrians, for which it is essential to act in a socially compliant manner to ensure politeness and safety. Towards addressing the social robot navigation problem through learning from demonstrations, this thesis introduces Socially CompliAnt Navigation Dataset (\scand{}) \citep{scand}, a large-scale dataset of demonstrations for socially compliant robot navigation. We show that learning a policy end-to-end through behavior cloning on \scand{} leads to behaviors that are perceived by human participants as more socially compliant than a classical navigation stack (\movebase{}). Additionally, this thesis introduces a novel hybrid algorithm \citep{raj2023targeted} that utilizes both the classical navigation stack and an end-to-end learned policy, switching between the two using a learned policy classifier. Both the dataset and the novel hybrid algorithm are detailed in Chapter \ref{chap:socially_compliant_nav}.

\begin{sidewaysfigure}
    \centering
    \includegraphics[width=\columnwidth]
    {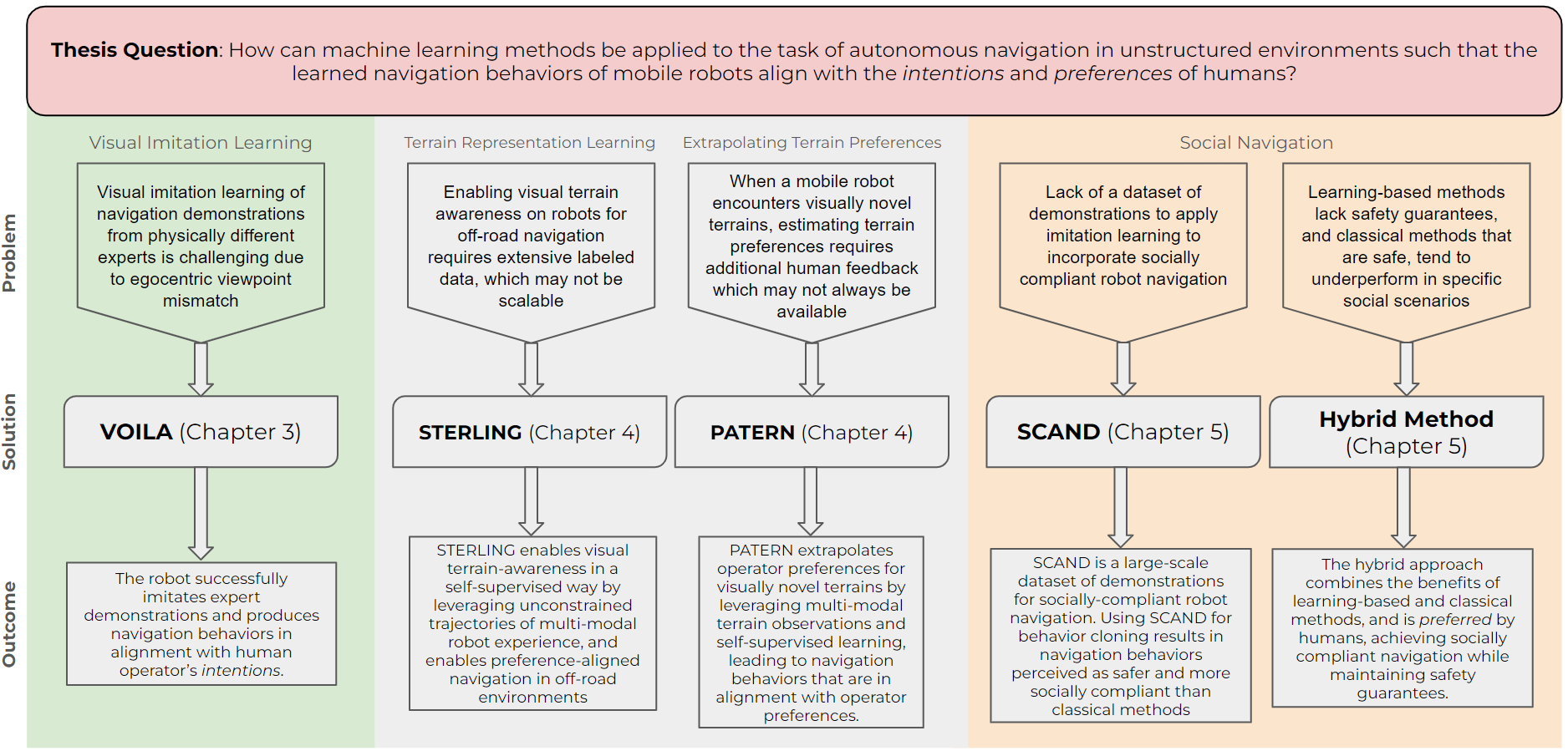}
    \caption{Flowchart summarizing the thesis question, problems, and their respective solutions proposed as contributions in individual chapters in this dissertation.}
    \label{fig:flowchart_thesis}
\end{sidewaysfigure}

Figure \ref{fig:flowchart_thesis} includes a flowchart summarizing the thesis question, problems, and their respective solutions proposed as contributions in individual chapters in this dissertation. 

\section{Reading Guide to the Thesis}

The rest of the thesis is organized as follows. Note that the contributions in Chapters \ref{chap:imitation_learning_for_robot_nav}, \ref{chap:pref_learning_offroad_nav}, and \ref{chap:socially_compliant_nav} in this thesis are not necessarily written to be read in that order and readers can feel free to read specific chapters of interest. 

\begin{itemize}
    \item Chapter 2 - \hyperref[chap:related_work]{Related Work}. This chapter reviews prior work that is closely related to the dissertation question and the three contributions of this thesis. 
    \item Chapter 3 - \hyperref[chap:imitation_learning_for_robot_nav]{Visual Imitation Learning for Robot Navigation}. In this chapter, I introduce the \voila{} algorithm for visual imitation learning from physically different expert's video-only demonstrations for robot navigation. This chapter addresses Contribution 1 of this thesis.
    \item Chapter 4 - \hyperref[chap:pref_learning_offroad_nav]{Preference Learning for Off-Road Navigation}. In this chapter, I introduce the \sterling{} and \patern{} algorithms for aligning off-road mobile robot navigation behaviors with operator preferences via self-supervised terrain representation learning and extrapolating operator preferences to novel terrains, respectively. This chapter addresses Contribution 2 of this thesis. 
    \item Chapter 5 - \hyperref[chap:socially_compliant_nav]{Socially Compliant Robot Navigation}. In this chapter, I introduce the Socially CompliAnt Navigation Dataset (\scand{}), a large-scale dataset of demonstrations for socially compliant robot navigation. I additionally introduce the hybrid algorithm for socially compliant robot navigation, leveraging classical and learned motion planners. This chapter addresses Contribution 3 of this thesis.
    \item Chapter 6 - \hyperref[chap:conclusion_and_future_work]{Conclusion and Future Work}. In this chapter, I conclude by providing a recap of the three contributions of this thesis in addressing the central thesis question. I additionally describe some ongoing work that is derived from contributions of this thesis and present ideas for future work. 
\end{itemize}


\chapter{Background}
\label{chap:related_work}

In this chapter, I provide an overview of the main areas of prior work related to this dissertation's topic of aligning robot navigation behaviors with human intentions and preferences, while also highlighting the gaps in existing literature that this dissertation addresses. The \textit{related work} sections in subsequent chapters will then delve into specific existing algorithms that are more closely related to each respective contribution. I begin this chapter in Section \ref{sec:learningfromdemo} with an overview of existing literature on imitation learning for robot navigation. Section \ref{sec:learningforoffroadnav} explores relevant studies in learning-based methods for off-road navigation, followed by Section \ref{sec:ml4socnav} surveying related work on machine learning for social robot navigation. Section \ref{sec:valuealignmentverification} extends the discussion to prior work on value alignment verification, particularly in the context of this dissertation's contributions, and describes how it is brought to bear in each of the contributions. Note that, in this chapter, the scope is restricted to learning-based approaches to robot navigation. For a more in-depth survey of classical and learning-based approaches to robot navigation and to understand their limitations and trade-offs, readers are directed to the works of \cite{xuesusurvey}, \cite{tai2016deepsurvey} and \cite{tai2016survey}. 

\section{Machine Learning for Robot Navigation}
\label{sec:learningfromdemo}

In this section, I explore prior approaches that utilize machine learning to enable autonomous navigation in mobile robots, specifically, imitation learning for robot navigation. I begin by exploring end-to-end approaches in machine learning for robot navigation, followed by reinforcement learning for robot navigation. Finally, I explore hybrid methods that combine machine learning with classical methods. 

The use of machine-learning-based approaches to learn end-to-end autonomous navigation policies goes back several decades \citep{alvinn, offroadoa}, though recent years have seen a spike in interest from the research community \citep{navigationbyimitation, cil, byronboots, xuesusurvey, semanticviz}.
One of the earliest successes was reported by Pomerleau et al. \citep{alvinn}, in which a system called \textsc{alvinn} used imitation learning to train a policy represented by an artificial neural network end-to-end (mapping from raw sensor observations directly to low-level control signals) to perform lane keeping based on demonstration data generated in simulation.
Since then, several improvements, both in the amount and type of demonstration data and in network architecture and training, have been proposed in the literature.
In particular, \cite{offroadoa} proposed the use of a convolutional neural network (CNN) to better process real demonstration images for an off-road driving task.
\cite{bojarskie2e} reported that gathering a large amount of real-world human driving demonstration data and applying data augmentation made it possible to train even more complex CNN architectures to perform lane keeping.
Recently, \cite{shah2023vint} and \cite{sridhar2023nomad} show that utilizing diverse open-source datasets from many robots to train a goal-conditioned navigation policy end-to-end serves as a good foundational control policy for goal-oriented navigation.
\cite{loquercio2018dronet} proposed the DroNet architecture for learning to fly a UAV based on driving datasets collected on the road. Using supervised learning, DroNet was able to control a UAV autonomously in unstructured urban environments. 

While the aforementioned approaches each use end-to-end imitation learning through supervised learning to find autonomous navigation policies, other machine learning for autonomous navigation work has adopted the alternative training paradigm of reinforcement learning.
\cite{semanticviz} propose using off-policy Q-learning from video demonstration data to learn goal-conditioned hierarchical policies for semantic navigation. \cite{contexttranslation} propose a context translation network to imitate an expert demonstration in the presence of viewpoint mismatch. While \cite{contexttranslation} effectively address viewpoint mismatches, their method primarily handles third-person views and does not extend to the egocentric viewpoint mismatches central to the contribution in Chapter \ref{chap:imitation_learning_for_robot_nav} of this dissertation. Similarly, \cite{gasketvisualservo} propose a visual servoing algorithm that uses image template-matching to provide rewards for their reactive agent. The work by \cite{byronboots} proposes using demonstrations from a privileged expert agent equipped with expensive sensors, to imitate the behavior on physically different hardware with cheaper sensors. The work closest to ours is that of \cite{kendall2019learning}, in which the proposed system learns a navigation policy using RL, where the reward function is the total distance traveled by their autonomous vehicle before a human driver intervenes (to, e.g., prevent collisions). 

An alternative approach to end-to-end controller learning is using inverse reinforcement learning to recover a cost function for navigation, that can be used in conjunction with a classical planner to navigate. \cite{wulfmeier2015maximum, wulfmeier2017large} extend the Maximum Entropy Inverse Reinforcement Learning (\textsc{meirl}) \citep{meirl} approach to deep cost function learning from lidar scans using a non-linear, neural network-based cost functions.  

In the studies referenced above, while successful navigation behaviors are achieved through the methodologies employed, these outcomes are predicated on the assumption of similar physical forms between the imitator and the expert. This assumption of requiring similar physical forms between the expert and imitator agents becomes problematic particularly when the expert is a human and the imitator is a ground vehicle, leading to significant discrepancies in embodiment. Such disparities often result in viewpoint mismatches when using an egocentric camera as the primary sensor, rendering existing visual imitation learning algorithms ineffective. The discussion in Chapter \ref{chap:imitation_learning_for_robot_nav} of this dissertation delves into this issue, introducing \voila{}, a visual imitation learning algorithm that remains effective despite the viewpoint mismatches caused due to physical embodiment mismatch between the demonstrator and imitator agents.

\section{Machine Learning for Off-Road Navigation}
\label{sec:learningforoffroadnav}

The problem of autonomous navigation in unstructured environments is well-studied, and several classical approaches have been proposed to address it \citep{alvinn, a_star, rosmovebase, biswas2013longterm, khandelwal2017bwibots}. While successful in long-term deployment scenarios, such classical heuristic-based approaches treat off-road navigation as geometric obstacle-avoidance (i.e., obstacles that are untraversable) problem. However, a mobile robot navigating in unstructured off-road environments may often experience a variety of terrains to traverse over with specific terrain traversability costs for navigation based on factors such as human preferences, ride comfort, or social norms. Classical approaches that solely perform geometric navigation will not be able to navigate in a terrain-aware manner, or worse, in some environments such as tall grass, a classical approach may choose not to navigate over it to avoid collisions, whereas, tall grass could be navigable for a mobile robot, and could be a human-preferred terrain to drive over. This lack of terrain-awareness of classical approaches to robot navigation has created a need for more advanced approaches to terrain-aware off-road navigation. 

To overcome existing limitations of classical heuristic-based approaches to off-road navigation, recently, several learning-based approaches have been proposed that are promising for terrain-aware navigation in unstructured off-road environments. \cite{kahn2021land} propose \textsc{land} for learning to navigate from disengagements. \textsc{land} takes a reinforcement learning-based approach where disengagements due to sub-optimal behavior of the agent are recorded to train the agent using reinforcement learning. \cite{kahn2021badgr} propose \textsc{badgr}, in which available sensory redundancy on a mobile robot is leveraged to learn behaviors on different types of terrains. \textsc{badgr} learns a planner from off-policy trajectories collected onboard during exploration. This learned planner can safely ignore geometric obstacles (i.e., obstacles that are untraversable) and navigate through safe-to-travel regions avoiding regions that are lethal and bumpy. \cite{yao2022rca} propose the \textsc{rca} algorithm for terrain-aware navigation to maximize ride comfort. In \textsc{rca}, an unsupervised representation learning algorithm is used to learn visual-inertial features from data collected on the robot.   

The work most relevant to preference learning for navigation is \vrlpap{} by \cite{vrlpap} in which a representation of visual patches of the terrain is learned and then the preference function is learned from the partial order rankings of terrains. While \vrlpap{} was shown to be successful at learning preference costs of terrains that align with the human operator's interests, it requires structured \textit{concave} demonstration trajectories such that the terrain of higher preference is the one being traversed and the lesser preferred terrain is present at the midpoint of the shortest path connecting the start and end location of the robot. This limitation of \vrlpap{} prohibits its applicability in situations where we have a large amount of unconstrained trajectory data, such as driving straight where the shortest path is indifferent structurally from the path traversed. 

Each of the approaches above is limited in that it cannot learn from unconstrained robot trajectories for terrain-aware off-road navigation, aligning with human preferences. While \vrlpap{} \citep{vrlpap}, the closest approach to this work can learn from geometrically constrained trajectory data, it requires data gathered on specific pairs of terrains encountered which may not be available in the environment. The \sterling{} \citep{sterling} algorithm introduced in Chapter \ref{chap:pref_learning_offroad_nav} uses self-supervised non-contrastive representation learning from unconstrained robot experiences, enabling it to learn relevant terrain representations from robot trajectories of any geometric shapes and sizes, which can be utilized for operator preference-aligned off-road navigation. Additionally, Chapter \ref{chap:pref_learning_offroad_nav} introduces \patern{} \citep{paternarxiv} that enables extrapolating operator preference to visually novel terrains. 


\section{Machine Learning for Social Navigation}
\label{sec:ml4socnav}

Recently, several algorithms have emerged that show the potential of applying learning to address challenges in robot navigation \citep{xuesusurvey}. Broadly speaking, in the robot navigation literature, learning-based approaches are successful in problems such as adaptive planner parameter learning \citep{xiao2020appld}, overcoming viewpoint invariance in demonstrations \citep{voila}, and end-to-end learning for autonomous driving \citep{bojarski, pfeiffer2018reinforced, wang2021agile, wurman2022outracing}. 

Specifically in applying imitation learning for social navigation, the work by \cite{tai2018socially} is the most relevant to this dissertation. They provide a simulation framework in Gazebo along with a dataset generated using the same where virtual human agents navigate following the social force model \citep{socialforce}. They additionally train a social navigation policy using the Generative Adversarial Imitation Learning algorithm assuming the social force model as the \textit{expert} demonstrator and show a successful deployment of the learned policy in the real-world on a turtlebot robot. While their work has shown that imitation learning can be applied to address the social navigation problem, they do so assuming the social force model in simulation as the \textit{expert}, socially-compliant policy. Although simulated environments enable fast and safe data collection for online learning, they lack the naturally occurring social interactions seen in the wild.

Other learning paradigms such as \rl{} have also been applied to address the social navigation problem. \cite{collavoideverett} present \textsc{ca}-\textsc{drl}, a multi-agent collision avoidance algorithm learned using \textsc{rl} that shows impressive results in the real-world on specific social navigation scenarios. \cite{soc_com_robot_nav_paper} use inverse reinforcement learning on human demonstrations to learn cost functions for the socially compliant navigation task. \cite{baghisesno} propose the \textsc{sesno} algorithm for sample efficient inverse reinforcement learning for social navigation using the UCY dataset \citep{ucydataset}.

Both imitation learning and reinforcement-learning-based approaches discussed above have either used inaccurate simulators with simplified motion models or human-only pedestrian motion tracking datasets to learn socially compliant navigation behaviors. In this dissertation, I posit that a large-scale dataset of socially compliant navigation demonstrations involving a robot in human-occupied environments would help learn relevant socially compliant navigation policies for autonomous mobile robots. Towards addressing the problem of lack of large-scale datasets for social robot navigation, in Chapter \ref{chap:socially_compliant_nav}, I present Socially CompliAnt Navigation Dataset (\scand) \citep{scand}, an extensive collection of robot social navigation demonstrations featuring multi-modal data captured in diverse real-world settings. This dataset, encompassing several hours of recordings from both indoor and outdoor environments, leverages two distinct robot morphologies and four different human demonstrators. To demonstrate the utility of this dataset, I employed an imitation learning algorithm to train an end-to-end policy.   Experimental results affirm the feasibility of learning robot navigation policies, that are perceived by human participants as being more safe and socially compliant in comparison to a classical heuristic-based navigation policy. A notable insight from analyzing \scand{} reveals that in approximately 80\% of cases, a traditional heuristic-based planner aligns with a learning-based method. This observation prompts a reevaluation of learning-based strategies in social navigation. Consequently, I propose a new hybrid approach for social robot navigation, detailed in Chapter \ref{chap:socially_compliant_nav}.

\section{Value Alignment Verification}
\label{sec:valuealignmentverification}
The preceding sections of this chapter have reviewed various learning-based algorithms documented in the literature that help learn robot navigation behaviors, that align with human intentions and preferences. This section extends the discourse to survey related work on the topic of value alignment verification---specifically, verifying whether the learned navigation policy aligns with the human operator's intentions and preferences. Note that the core contributions of this dissertation are centered around devising algorithms and datasets that enable mobile robots to navigate in accordance with a human operator's intentions and preferences, and not to propose new ways to verify value alignment. This is still an active area of research in the field of machine learning and falls beyond the scope of this dissertation.

\cite{brown2021value} formally introduce the value alignment verification problem and propose designing a \textit{driver's test} to efficiently verify whether an AI system is aligned with a human's values through a minimal number of queries. Exact value alignment involves sampling the policy at all possible states which can be expensive and are often impossible to evaluate since the number of states can be prohibitively large for some tasks, such as autonomous driving. To tackle this, \cite{brown2021value} propose heuristic and approximate value alignment verification for grid-world and continuous action domains. \cite{hadfield2016cooperative} propose \textsc{cirl} which introduces a cooperative game between the agent and the human, where both agents are rewarded according to the human's reward function, but unlike the human, the robot does not initially know the reward function. While \textsc{cirl} ensures the robot's learned policy asymptotically converges to the human intended values, it does not propose a way to verify this alignment. Evaluating social compliance in robot navigation is still an unsolved problem in the navigation community. For further details, \cite{francis2023principles} recently published a comprehensive survey on principles and guidelines for evaluating social robot navigation. There have been several proposed approaches both using simulation \citep{tsoi2020sean, tsoi2021approach, kastner2023arena} and physical robot experiments \citep{svensson2003defining, soreneval}. The inaccuracy of long-term human motion models, the wide variety of social navigation scenarios, and the subjectivity of ``socialness" in navigation make evaluating social compliance a challenging task in robot navigation. \cite{knox2022reward} highlight major flaws in existing human-designed reward functions for autonomous driving, and propose eight sanity checks for a human-designed reward function, underscoring the importance of addressing the value alignment problem in robot navigation by alleviating reward misdesign. 

In this dissertation, we take different approaches in each contribution to verify the alignment of the learned navigation behavior to a human operator's intentions and preferences. To verify the alignment of the policy learned using \voila{}, we report the Hausdorff distance of the imitated trajectory to the ground truth trajectory demonstrated by the human. We verify that the learned policy imitates the intended navigation behavior by checking that the trajectory traced by the learned policy closely matches the human-demonstrated trajectory in the physical robot experiments. Additionally, in simulation experiments, in novel environments unseen by the agent, we report the Hausdorff distance of the robot's states between the human-demonstrated trajectory and the trajectory traced by the learned policy and find that the imitated policy closely follows the unseen human demonstrated trajectory, verifying that \voila{} has indeed learned the intended navigation behavior. In Chapter \ref{chap:pref_learning_offroad_nav}, to validate the preference alignment of the trajectories traced by \sterling{} and \patern{} algorithms, we utilize a success metric in which a trajectory traced by the robot using any algorithm from start to goal locations are considered successful if it adheres to the operator's ranked preferences over the terrains. In Chapter \ref{chap:socially_compliant_nav}, to verify alignment with human preferences for socially compliant navigation, we perform a human participant study with post-experience questionnaires requesting feedback on a Likert-scale, evaluating social compliance, and safety of the learned controller. 

\chapter{Visual Imitation Learning for Robot Navigation}
\label{chap:imitation_learning_for_robot_nav}

In this chapter, I introduce the first contribution of this thesis, Visual\hyp{}Observation\hyp{}only Imitation Learning for Autonomous navigation (\voila{}) \citep{voila}, a visual imitation learning algorithm for autonomous navigation in the presence of egocentric viewpoint mismatch between a physically different demonstrator such as a human, and a mobile robot as the imitator. This chapter is organized as follows. The introduction, Section \ref{sec:intro_voila}, provides an overview of this contribution, followed by background and related work in Section \ref{sec:bg_and_rl_voila}. Section \ref{sec:voila} introduces \voila{} and the novel reward function for visual imitation learning. Sections \ref{sec:implementation_voila} and \ref{sec:experiments_voila} provide details on the implementation and the experiments performed for evaluation, respectively. 

\section{Introduction}
\label{sec:intro_voila}
Enabling vision-based autonomous robot navigation has recently been a topic of great interest in the robotics and machine learning community \citep{bojarski, cil, byronboots}.
As discussed in Section \ref{sec:learningfromdemo}, imitation learning in particular has emerged as a useful paradigm for designing vision-based navigation controllers. Using this paradigm, the desired navigation behavior is first demonstrated by another agent (usually a human), and then a recording of that behavior is supplied as training data to a machine learner that tries to find a control policy that can mimic the demonstration.
To date, most approaches in the navigation domain that use imitation learning require demonstration recordings that contain both state observations (e.g., images) and actions (e.g., steering wheel angle or acceleration) gathered onboard the deployment platform \citep{bojarski, cil, offroadoa,  end2endsurvey}.

As explored in Section \ref{sec:learningfromdemo}, while these existing imitation learning approaches have proved successful in certain scenarios, there are situations in which it would be beneficial to relax the requirements they impose on the demonstration data.
For example, if we wish to collect a large number of demonstrations from many experts, it may prove too difficult or costly to arrange for each expert to operate specific deployment platforms, which are often expensive or difficult to transport. Additionally, it might be costly to outfit all demonstration platforms with instrumentation to record the control signals with the demonstration data.
However, due to the low cost and portability of video cameras, it may still be feasible to have demonstrators record ego-centric video demonstrations of their navigation behaviors while operating a different platform. 
Demonstrations of this nature would consist of video observations only (i.e., they would not contain control signals), and, because of the difference in platform, the videos would likely exhibit ego-centric viewpoint mismatch compared to those that would be captured by the deployment platform.
One example of such data is the plethora of vehicle dashcam videos available in publicly accessible databases \citep{dashcamvids} or on YouTube.
Another example is video demonstrations of robot behaviors generated by proprietary code that one would like to mimic on the same or different robot hardware. Unfortunately, to the best of our knowledge, there exist no current imitation learning techniques for vision-based navigation that can leverage such demonstration data.

Fortunately, recent work in \textit{Imitation from Observation} (\ifo{}) \citep{farazijcaisurvey}---imitation learning in the absence of demonstrator actions---has shown a great deal of success for several related tasks. For example, work in this area has been able to learn from video-only demonstrations for both simulated and real limbed robots \citep{farazridm, gaifopaper, TCN, zeroshotil}.
However, no literature of which we are aware has considered whether these \ifo{} techniques can be applied to the vision-based autonomous navigation problem outlined above. This problem is especially challenging since physical differences in the demonstration platform introduce viewpoint mismatch in the video demonstrations.

In this Chapter, I present our hypothesis that it is possible to learn visual robot navigation policies that are aligned with a demonstrator's intentions, even if such demonstrations contain significant viewpoint differences due to embodiment mismatch between the demonstrator and the robot. To this end, we introduce a new \ifo{} technique for vision-based autonomous navigation called {\em Visual-Observation-only Imitation Learning for Autonomous navigation} (\voila{}).\footnote{A preliminary version of this work was presented at the 2021 AAAI Spring Symposium on Machine Learning for Navigation. The final version was published at the 2022 ICRA conference.} 
To overcome viewpoint mismatch, \textsc{voila} uses a novel reward function that relies on off-the-shelf keypoint detection algorithms that are themselves designed to be robust to egocentric viewpoint mismatch. This novel reward function is utilized to drive a reinforcement learning procedure that results in navigation policies that imitate the demonstrator.

We experimentally confirm our hypothesis both in simulation and on a physical Clearpath Jackal robot.\footnote{The research described in this chapter was done in collaboration with Garrett Warnell, Xuesu Xiao, and Peter Stone.} We compare \voila{} against a state-of-the-art \ifo{} algorithm \gaifo{} \citep{gaifopaper}, and show that \voila{} can learn to imitate an expert's visual demonstration in the presence of viewpoint mismatch while also generalizing to environments not seen during training. Additionally, we demonstrate the flexibility of \voila{} by showing that it can also support vision-based training of navigation policies with observation inputs other than camera images.

\section{Background and Related Work}
\label{sec:bg_and_rl_voila}
The proposed approach, \textsc{voila}, performs reinforcement learning (RL) using a novel reward function based on image keypoints in order to accomplish imitation from observation for autonomous navigation with viewpoint mismatch. In this section, I review related work that is more closely related to \voila{}, such as autonomous robot navigation, imitation from observation, and computer vision techniques for visual feature extraction. For a more general literature review on machine learning for robot navigation, refer to Section \ref{sec:learningfromdemo}.

\subsection{Machine Learning for Autonomous Navigation}

The use of machine learning methods in the design of autonomous navigation systems goes back several decades, though recent years have seen a spike in interest from the research community \citep{navigationbyimitation, cil, byronboots, xuesusurvey, semanticviz}.
One of the earliest successes was reported by \cite{alvinn}, in which a system called \textsc{alvinn} used imitation learning to train an artificial neural network that could perform lane-keeping based on demonstration data generated in simulation.
Since then, several improvements, both in the amount and type of demonstration data and in network architecture and training, have been proposed in the literature.
In particular, \cite{offroadoa} proposed the use of a convolutional neural network (CNN) to better process real demonstration images for an off-road driving task, and, more recently, \cite{bojarski} reported that gathering a large amount of real-world human driving demonstration data and applying data augmentation made it possible to train even more-complex CNN architectures to perform lane keeping.


The work closest to ours is that of \cite{kendall2019learning}, in which the proposed system learns a navigation policy using RL, where the reward function is the total distance travelled by their autonomous vehicle before a human driver intervenes (to, e.g., prevent collisions).
However, unlike \textsc{voila}, \cite{kendall2019learning} utilize experience gathered exclusively by the learning platform itself, considering neither imitation from observation nor the particular problem of viewpoint mismatch. 

\subsection{Imitation from Observation}

Recently, there have been a number of imitation from observation (\ifo{}) techniques introduced in the literature, including an adversarial approach called \gaifo{} \citep{gaifopaper, vgaifoso} that we use here as a baseline. 
In \gaifo{}, the reward signal is provided by a learned discriminator network that seeks to reward state transitions similar to those present in the demonstration and penalize---if it can tell the difference---state transitions that come from the imitator.
While \textsc{gaifo} \citep{gaifopaper} has been shown to be successful in both low- and high-dimensional observation spaces, it has thus far only been applied to continuous control tasks for limbed agents.
Moreover, as we will show in our experiments in Section \ref{sec:experiments_voila}, while \gaifo{} can imitate the expert when the egocentric viewpoints between the expert and the imitator match, it is unable to do so in the presence of viewpoint mismatch.

Viewpoint mismatch in \ifo{} has been previously considered in the work by \cite{TCN}, which proposes Time Contrastive Networks (\textsc{tcn}s). \textsc{tcn}s use a triplet loss metric to learn a feature space embedding which is then used for rewarding the agent to imitate the expert. While both \voila{} and \textsc{tcn}s are robust to viewpoint mismatch, \textsc{tcn}s require demonstration data with multiple viewpoints in the same timestep to learn an embedding space robust to viewpoint mismatch, whereas \voila{} achieves this robustness by leveraging feature detection algorithms (e.g \textsc{sift} by \cite{sift}) commonly used in \textsc{slam} that are themselves designed to be robust to viewpoint mismatch.

\subsection{Feature Detection and Matching}
To overcome viewpoint mismatch, \voila{} utilizes a novel reward function that relies on local image features such as keypoints and their descriptors.
Keypoints have been used for decades to solve challenging tasks such as image verification, matching, and retrieval.
More recently, deep-learning-based keypoint extractors such as \textsc{superpoint} by \cite{superpoint} have been shown to be more successful than classical approaches.
In this work, we use \textsc{superpoint} to detect keypoints and their corresponding descriptors, and we determine keypoint matches between two images using the typical method based on the two nearest neighbors in descriptor space \citep{superpoint} using the $\ell_2$ distance metric. However, in principle \voila{} can be used with any local feature detector or feature-matching algorithm. In our experiments, we use the simplest image retrieval algorithm of local keypoint detection followed by nearest neighbor matching in the descriptor space to identify the visually closest image with maximum keypoint matches. However, there exist more sophisticated algorithms such as \textsc{n}et\textsc{vlad} \citep{netvlad, hfnet} that are more robust to visual aliasing errors for the task of image retrieval which can be used with \voila{} with no changes to the underlying algorithm.

Several works have proposed learning a keypoint detector specific to the imitation learning task \citep{mbrlkeypoint, keypointsintofuture}. Unlike such approaches, \voila{} uses an off-the-shelf keypoint extractor that is not trained specifically for the navigation task.

\subsection{SLAM-based Approaches for Navigation}
Visual Teach and Repeat (\textsc{vtr}) methods such as the one proposed by \cite{timbarfootvtr} follow a two-step approach to imitating a navigation demonstration. First, a SLAM map of the demonstration environment is built in the \emph{teach} phase. In the subsequent \emph{repeat} phase, the robot localizes within a submap of the environment and follows a desired trajectory. \voila{}, on the other hand, learns a reactive navigation policy directly from demonstrations, sidestepping the SLAM problem. 

\section{Visual Imitation Learning for Autonomous Navigation}
\label{sec:voila}
In this section, I formulate the visual imitation learning problem for the task of autonomous visual navigation, which we pose as a reinforcement learning problem with a demonstration-dependent reward.
The critical contribution of \voila{} is the development of this particular reward function, as described in detail below.

\subsection{Preliminaries}
We treat autonomous visual navigation as a RL problem where the environment is a Markov decision process. 
At every time step $t$, the state of the agent is described by $s_t\in\mathcal{S}$, the observation of the agent is described by $O_t\in\mathcal{O}$, and an action $a_t \in \mathcal{A}$ is sampled from the agent's policy $a_t \sim \pi(\cdot|O_t)$.\footnote{While \voila{}'s reward function depends on camera images, the imitation policy can actually be learned over {\em any} appropriate state representation---vision-based or otherwise. We show one such example using LiDAR scans as the state representation in Figure~\ref{fig:voila_trajs_w_mismatch}. }
A single expert demonstration is represented as a set of $n$ sequential observations $\mathcal{D}^e = \{I_1, I_2, \dots ,I_n\}$. Performing this action in the environment leads to a next state $s_{t+1} \sim T(\cdot|s_t,a_t)$, where $T$ is the unknown transition dynamics of the agent in the environment. For this specific transition, the agent receives a reward, $r_{t+1} \in \mathbb{R}$, which is a function of both the agent's transition tuple and the demonstration, i.e., $r_{t+1} = R(O_t, a_t, O_{t+1}; \mathcal{D}^e)$. The relative utility of near-term and long-term reward is controlled using the discount factor $\gamma \in (0, 1]$.
The RL objective is to find a policy $\pi$ that maximizes the expected sum of discounted rewards $\mathbb{E}[\Sigma_{t=0}^{\infty} \gamma^t R(O_t, a_t, O_{t+1}; \mathcal{D}^e)]$. 

\subsection{Reward Formulation for Imitation Learning}


For each transition $(O_t, a_t, O_{t+1})$ experienced by the learner, we require a reward $r_{t+1}$ such that the learner, by optimizing the RL objective with this reward, can learn to imitate the demonstration.
In particular, because we wish to perform learning in real-time, we seek a dense reward function that provides feedback at each timestep without delay.
Since the expert demonstrations are from a physically different agent, there can be significant ego-centric viewpoint mismatch between the observation spaces of the learner and the demonstrator as shown in Figure~\ref{fig:voila_trajs_w_mismatch}. Such a mismatch poses a challenge to designing a good reward function since it is not immediately clear how to compare images from different viewpoints. Hence, we introduce here a novel reward function based on keypoint feature matches between the expert and the imitator's ego-centric observations for the task of imitation learning for visual navigation. Keypoint detectors have been extensively used in the computer vision community for several decades to solve challenging tasks like structure-from-motion (\textsc{s}f\textsc{m}), visual \textsc{slam} and hierarchical localization. Recent keypoint detection algorithms such as \textsc{superpoint} by \cite{superpoint} provide invariance to perspective distortion, scaling, translation, rotation, viewpoint mismatches, and varied lighting conditions between the key and query images. Hence, we use keypoint detectors to help define the reward function to learn visual navigation policies from demonstrations provided by any other agent. 

The reward function we propose relies on a quantity that we call {\em match density}. We define the match density $d(O_1, O_2)$ between two images $O_1$ and $O_2$ as the ratio of the number of keypoint matches between $O_1$ and $O_2$, and the total number of detected keypoints in $O_2$.
$d(O_1, O_2) \in [0, 1]$, assuming there is always a non-zero number of keypoints detected in an image.
Additionally, instead of imposing a temporal alignment constraint, we define the reward for a particular transition by searching the demonstration for the image which is most visually similar to the learner's current observation.
Here, we define the most visually similar image in the expert demonstration to be the one that has the highest match density with $O_t$, which we denote as $I_t$. Figure~\ref{fig:voila_trajs_w_mismatch} shows the imitator's current observation $O_t$ and corresponding visually closest expert image $I_t$.
For convenience of notation, we denote the next image after $I_t$ in the demonstration sequence as $J_t$.

Using the concepts described above, we now define the proposed reward function for \textsc{voila}:

\begin{equation}
   R(O_t, a_t, O_{t+1}; \mathcal{D}^e) = 
   \begin{cases} 
      F + V - \lambda ||a^{steer}_t|| &, alive \\
      -10 &,  done \\
   \end{cases} \; ,
  \label{rewardfunc}
\end{equation}
where $F=d(O_{t+1}, I_{t+1})$ and $V=\gamma * d(O_{t+1}, J_t) - d(O_t, J_t)$. 
If the robot is in the {\em done} state, i.e., it has crashed (as detected in AirSim, or by the trainer in physical experiments) or the number of keypoint matches drop below 10, the agent receives a penalty reward of $-10$.
Otherwise, the agent is in the $alive$ state, and we assign a reward that depends on terms $F$ and $V$.
The $F$ term assigns reward value based on the match density encountered at the next observation $O_{t+1}$ that the agent ends up in the transition.
This component encourages the agent to stay on the demonstrated trajectory.
The $V$ term is similar to a potential-based shaping term, and rewards a transition based on the difference in the match densities with the next expert observation $J_t$ and the imitator's observations.
This component encourages the imitator to find a policy that exhibits similar state transitions to those experienced by the expert.
We additionally found that adding the action penalty term with a $\lambda$ of $0.01$ penalizes the agent for making large steering changes. Since the maximum value of the reward function in the \textit{alive} state is 2, we heuristically picked a value 5 times higher and arrived at -10 for the negative penalty. This value worked well empirically in our experiments; we do not have any reason to believe that the results are particularly sensitive to this exact value.
The expert image retrieval step is performed in real-time using feature matching and is outlined in the implementation section.

\section{Implementation}
\label{sec:implementation_voila}
In this section, I provide specific implementation details of \voila{} including those related to representation learning, keypoint feature extraction, and the network architectures.

\subsection{Representation Learning}

Representation learning using unsupervised learning is a powerful tool to improve the sample efficiency of deep RL algorithms. Instead of learning a navigation policy over high dimensional image space, \textsc{voila} uses a latent representation of the image and learns the navigation policy over this latent code as input to the policy. Specifically, \textsc{voila} uses a Regularized Auto Encoder (\textsc{rae}) \citep{rae} to learn a latent posterior of the visual observations of the imitator.
The imitating control policy is then learned using RL with the latent code $z_t = g_\phi(O_t)$ as the input to the policy network, where $g_\phi$ is the encoder of the \textsc{rae} with weights $\phi$.
A ResNet-18 encoder-decoder network architecture is used for the \textsc{rae} and is trained for the task of image reconstruction, with data collected from the imitator using random rollouts.
The input images are of size $256 \times 256$ and the size of the latent dimension is $512$.
Random cropping and random affine image augmentations are utilized to regularize training.

\subsection{Keypoint Feature Extraction}
As a preprocessing step, all \textsc{superpoint} features detected from expert observations are stored in a buffer. At the start of an episode (for the first frame), the nearest expert observation $I_t$ to $O_t$ is retrieved by linearly searching for the closest expert observation in $\mathcal{D}^e$ with the maximum feature matches. As the episode unfolds, instead of exhaustively searching for the closest expert image at every transition, the search is restricted over the three next expert observations forward in time from the previous closest expert image in $\mathcal{D}^e$. At each transition, \textsc{superpoint} keypoints and descriptors are extracted and used to retrieve the closest expert image and compute the reward according to Equation \ref{rewardfunc}. Note that the \textsc{superpoint} keypoint descriptors extracted in an image are the local features and not the global features for that image. Hence, we explicitly train an \rae{} to learn a compressed global representation of the image for training the navigation policy, as described in the previous section. The \textsc{superpoint} descriptor for every keypoint is a vector of 512 float values. These descriptors are learned from data and correspond to different features of a keypoint including---but not limited to---lighting, location, appearance, and texture.

\subsection{Navigation Policy Architecture}
\label{experimental_setup}

As shown in Figure~\ref{fig:voila_arch}, we model the navigation policy $\pi_\theta$ using a 3-layer, fully-connected neural network, with 256 neurons per layer. We frame-stack three consecutive latent codes of observations in time to alleviate the effects of partial observability in the environment. We also additionally include the most recent action performed by the agent as a part of the state. We use Soft Actor-Critic (\textsc{sac}) by \cite{haarnoja2018softsac}, an off-policy RL algorithm, to learn $\pi$. 

\begin{figure*}[!tb]
\centering
\includegraphics[width=\columnwidth]{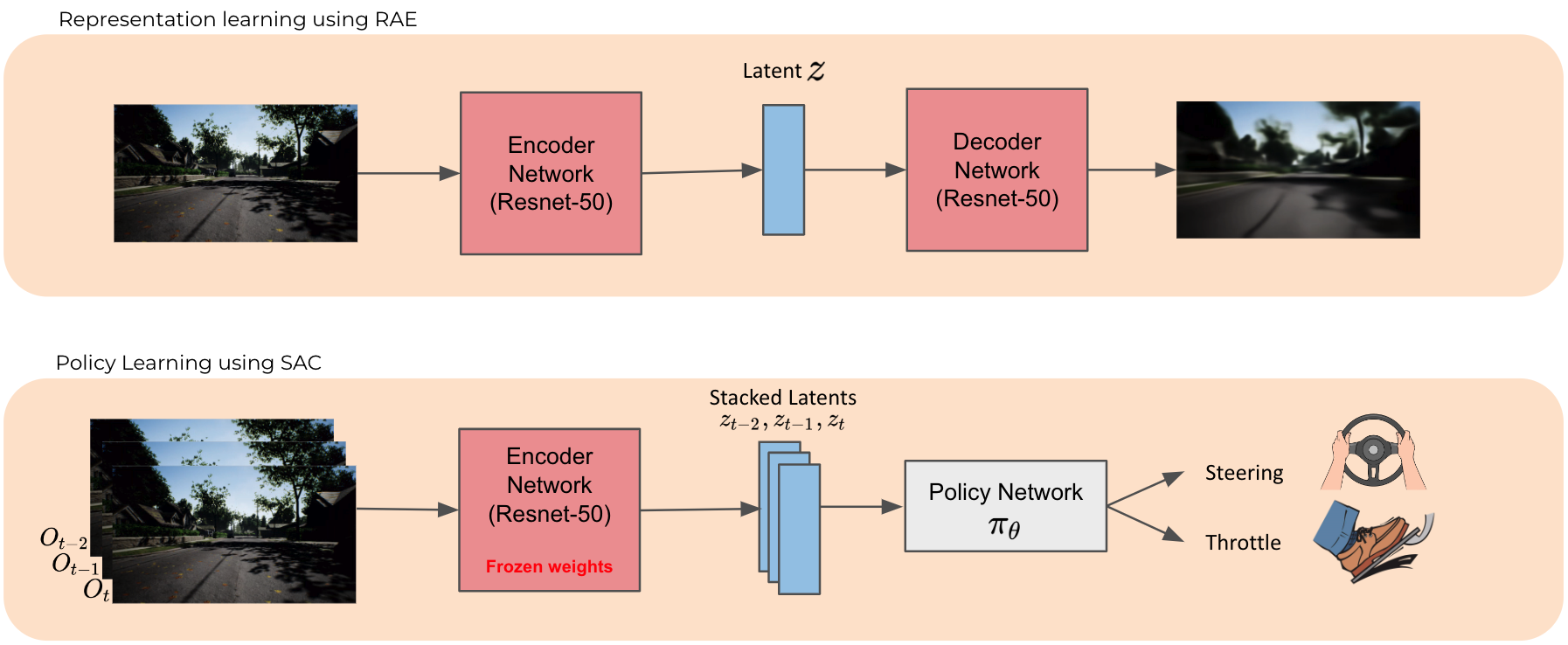}
\caption{Overview of the representation learning and reinforcement learning training approaches in \voila{}. To improve sample efficiency in the RL training step, we first learn a latent representation of the visual observations using an \textsc{rae}\citep{rae}. We then frame-stack the latent representations $z_{t-2}, z_{t-1}, z_t$ of three consecutive observations as state inputs to the policy network $\pi_\theta$, which is then trained using \textsc{sac} \citep{haarnoja2018softsac}, with the visual encoder network's weights frozen.}
\label{fig:voila_arch}
\end{figure*}


\section{Experiments}
\label{sec:experiments_voila}

I now describe the experiments that we performed to evaluate \voila{}.
The experiments are designed to answer the following questions:
\begin{enumerate}[label=($Q_\arabic*$)]
    \item Is \voila{} capable of learning imitative policies from video demonstrations that exhibit viewpoint mismatch?
    \item How well do policies learned using \voila{} generalize to environments unseen during training?
    \item Does \voila{} work with sensor modalities other than vision?
\end{enumerate}

\noindent To answer the questions above, we performed experiments using both a simulated autonomous vehicle and a real Clearpath Jackal robot on the task of imitating ``road following" and ``hallway patrol" tasks, respectively. The objective of the \voila{} agent is to imitate the expert's visual demonstration by learning an end-to-end navigation policy, even in the presence of viewpoint mismatch. To quantify the performance of imitation policies, we compute the Hausdorff distance metric (lower is better) between trajectories generated on a held-out set of environments.

\subsection{Simulation Experiments in AirSim}
In our simulation experiments, we answer questions $Q_1$ and $Q_2$ using the outdoor `Neighborhood' environment in AirSim \citep{airsim} and learn the task of road following, i.e., driving on a straight road avoiding collisions with obstacles such as parked cars along the curb while dealing with varied lighting conditions along its path. To this end, we pick 12 straight road segments (tracks) in the AirSim environment, as shown in Figure~\ref{fig:airsim_top}. Two tracks (Track 1 and 2) were used for training the agent, and the learned policy was deployed on all 12 tracks. The other 10 tracks and their expert demonstrations were not seen by the agent before evaluation, and so we use them to test the generalizability of the learned policy to unseen environments. In each episode, the car begins at a randomized initial position near the start of the track, so the agent cannot trivially solve the task by learning to drive straight without having to steer. The expert demonstrations consist of a single trajectory (egocentric, front-facing images) for all tracks provided by a human (the first author) controlling the demonstration vehicle with the objective of navigating from start to end of the tracks, driving straight, in the middle of the road, and avoiding collisions with obstacles such as parked cars.

\begin{figure}[!tb]
    \centering
    \includegraphics[scale=0.25]{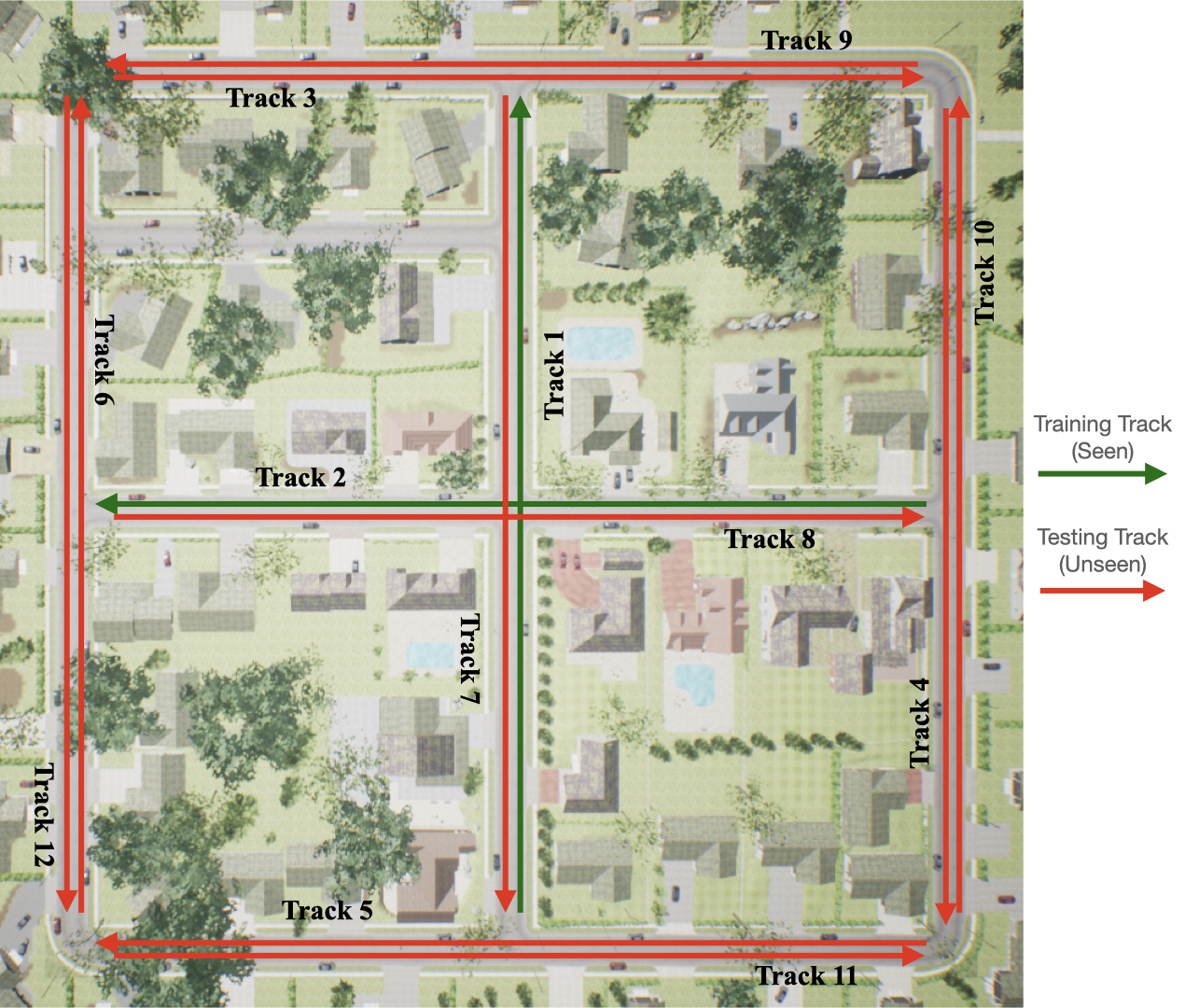}
    \caption{Aerial image of the AirSim simulation environment. Green lines show the tracks used to train the agent and red lines show the tracks unseen by the agent.}
    \label{fig:airsim_top}
\end{figure} 

We use the latent vector of the \textsc{rae} as the state representation, and the action space consists of changes in steering and throttle values. Note that expert demonstrations were required only during training to compute the reward. At test time, the agent imitates the expert without requiring access to expert demonstrations. 

\begin{figure*}[!tb]
\begin{subfigure}{0.5\columnwidth}
    \centering
    \includegraphics[scale=0.45]{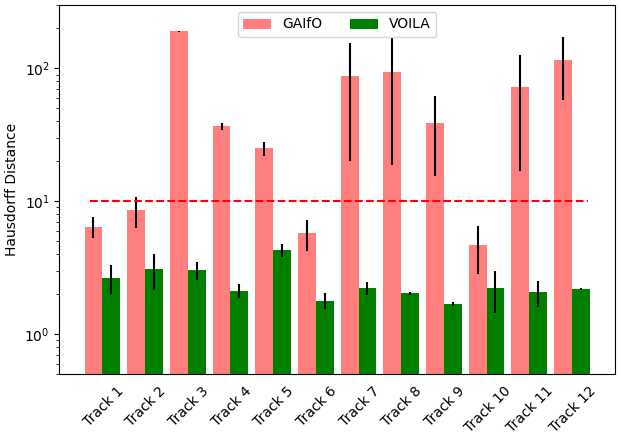}
    \caption{Without viewpoint mismatch between expert and imitator}

    \label{fig:bar_wo_vp}
\end{subfigure}
\begin{subfigure}{0.5\columnwidth}
    \centering
    \includegraphics[scale=0.45]{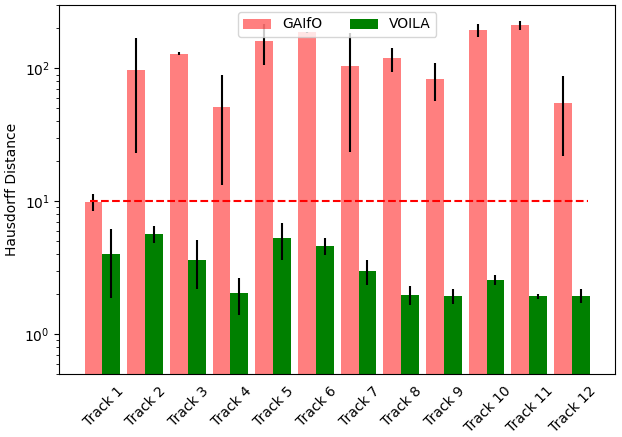}
    \caption{With viewpoint mismatch between expert and imitator}
    \label{fig:bar_w_vp}
\end{subfigure}
\label{fig:bargraphs}

\caption{Imitation performance of policies learned using \voila{} and \gaifo{} in AirSim. The y-axis shows the Hausdorff distance between the expert and imitator's trajectories, averaged across five trials (lower distance indicates behavior more similar to the expert). A Hausdorff distance greater than 10.0 (marked by the red line) indicates a failure in imitating the demonstration. We see that with viewpoint mismatch, the \gaifo{} agent is unable to imitate the expert successfully on all tracks, whereas \voila{} is unaffected by viewpoint mismatch and results in policies that induce behavior closer to that of the demonstrator. Tracks 1 and 2 were used for training, and other tracks were unseen by the agent while learning.}
\end{figure*}

We compare \voila{} against \gaifo{} \citep{gaifo, gaifopaper}, a state-of-the-art \ifo{} algorithm that does not explicitly seek to overcome viewpoint mismatch. While both \gaifo{} and \voila{} are \ifo{} algorithms that can imitate from video-only demonstration data, \gaifo{} has been evaluated predominantly in domains such as limbed-robot locomotion and manipulation, whereas \voila{} has been designed specifically for vehicle navigation domains. Additionally, \gaifo{} uses a learned reward function whereas in \voila{}, we propose a manually defined, demonstration-dependent reward function that is not learned. 
To ensure a fair comparison, we provide each algorithm the same state representation, i.e., the latent code of the \textsc{rae}.
Further, since \gaifo{} is an on-policy algorithm whereas \voila{} relies on the off-policy \textsc{sac} algorithm, we allow \gaifo{} ten times more training timesteps than \voila{} (1 million vs. 100,000).
Finally, we report results for \gaifo{} using the policy that achieved maximum on-policy returns during training.

Figure~\ref{fig:bar_wo_vp} compares \voila{} and \gaifo{} without any viewpoint mismatch between the expert and imitator; we see that \gaifo{}, as expected, is able to imitate the expert demonstration on the training Tracks 1 and 2, but it fails to generalize to most unseen tracks. The policy trained with \voila{} performs better than \gaifo{} at imitating the expert demonstration on the training tracks, and also generalizes to unseen environments, addressing $Q_2$.
Figure~\ref{fig:bar_w_vp} addresses both $Q_1$ and $Q_2$. In Figure~\ref{fig:bar_w_vp}, in the presence of viewpoint mismatch, we see that, as expected, \gaifo{} is unable to imitate the expert on the training Track 2 and does not generalize to other environments. However, confirming our hypothesis, \voila{} is able to imitate the expert demonstration even in the presence of viewpoint mismatch on seen Tracks 1 and 2 and also generalizes to the other 10 unseen tracks. 

In summary, in the simulation experiments performed on AirSim, we observed that \voila{} is indeed capable of learning imitative policies from video-only demonstrations even in the presence of significant viewpoint mismatch (addressing $Q_1$), and a policy learned using \voila{} does generalize to novel environments (addressing $Q_2$).




\subsection{Physical Experiments on the Jackal}
To answer $Q_1$ and $Q_2$ on a physical robot, we performed experiments using a Clearpath Jackal---a four-wheeled, differential drive ground robot equipped with a front-facing camera. The environment considered is an indoor office space, shown in Figure~\ref{fig:voila_trajs_w_mismatch}, consisting of carpeted floors, straight hallways, intersections, and turns.
There are also static obstacles such as benches, chairs, whiteboards, pillars along the wall, and trashcans, all of which the robot needs to avoid colliding with.

We evaluated \voila{} on a hallway patrol task, in which the robot begins at a start state (shown in Figure~\ref{fig:voila_trajs_w_mismatch}) and patrols around the building clockwise by taking the first right at intersections and driving straight in the hallways.
To obtain a video demonstration of this task from a physically different agent, a human (the first author) walked the patrol trajectory once while recording video using a mobile phone camera held approximately 4 feet above the ground (the robot's camera is at approximately 0.8 feet from the ground).
To contrast the imitation learning performance of \voila{} with and without any viewpoint mismatch, we performed additional experiments, henceforth called \voila{}-w/o-mismatch in which the expert demonstrations were collected onboard the deployment platform itself. These demonstrations were collected using the \textsc{ros} \texttt{move\_base} \citep{rosmovebase} navigation stack with a pre-built map of the environment and waypoints to patrol the environment while recording the egocentric visual observations from the front-facing camera.

A \voila{} training episode consists of the robot starting at approximately the same start state (as shown in Figure~\ref{fig:voila_trajs_w_mismatch}) and exploring in the training environment until the agent reaches the \textit{done} state.
After each training episode, the robot was manually reset back to the start state by a human operator, and a new training episode began.
Training the navigation policy happened onboard the robot on a \textsc{gtx} 1050Ti \textsc{gpu}.

The \voila{} agent trained using demonstrations with and without viewpoint mismatch learned to imitate the expert within 90 minutes (120 episodes) and 60 minutes (100 episodes), respectively, of experiment time (including time taken to reset the robot at the end of an episode).
Figure~\ref{fig:voila_trajs_w_mismatch} shows the trajectory rollout (in green) of the policy learned using \voila{}, imitating the expert demonstration in the presence of viewpoint mismatch. Addressing $Q_1$, we see that \voila{} is able to successfully patrol the indoor environment in a real-world setting, as demonstrated by a physically different expert agent, in the presence of viewpoint mismatch.
 
\begin{figure*}[!tb]
    \centering
    \includegraphics[scale=0.25]{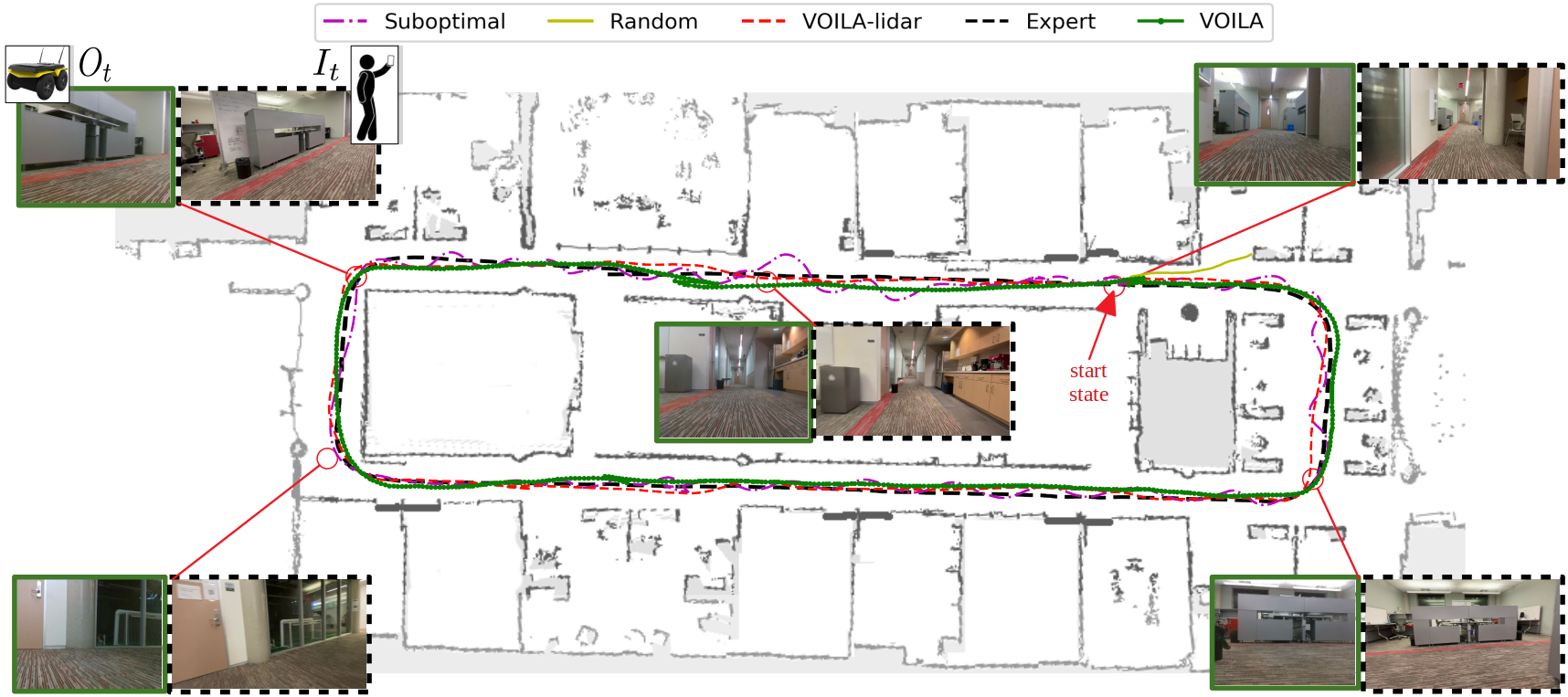}
    \caption{Policy rollout trajectories of the \voila{} agent (green) successfully imitating a demonstration behavior (black) of patrolling a rectangular hallway clockwise. The demonstration consists of a video gathered by a human walking while using a handheld camera that is considerably higher than the robot's camera (introducing significant viewpoint mismatch). We see that the \voila{} agent is able to successfully imitate the expert demonstration even in the presence of this egocentric viewpoint mismatch.}
    \label{fig:voila_trajs_w_mismatch}
\end{figure*} 

 \begin{figure}[!tb]
    \centering
    \includegraphics[scale=0.475]{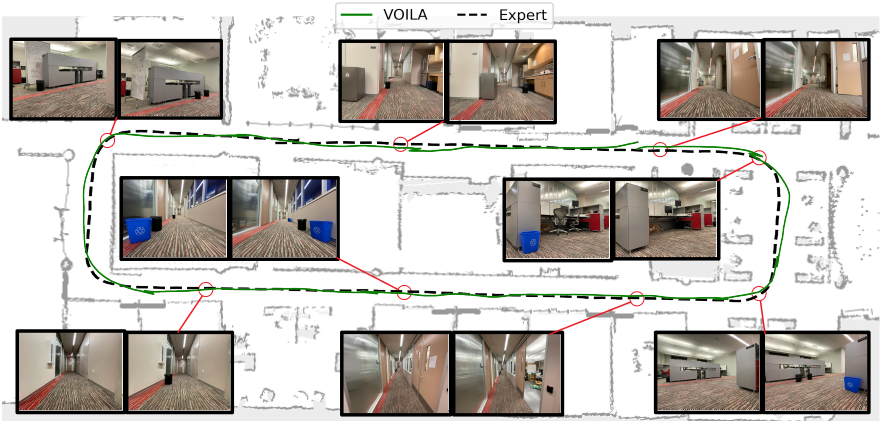}
    \caption{The \voila{} agent (green), trained in the unperturbed training environment (left), deployed here in the perturbed environment (right). We see that the learned policy is robust to the visual differences between the training and deployment environment, examples of which are provided as image pairs.}
    \label{fig:voila_perturbed}
\end{figure}


To evaluate the generalizability of policies learned using \voila{} to unseen real-world conditions ($Q_2$), first, we deploy the policy learned by \voila{} in a `perturbed environment', in which positions of movable objects such as trashcans, doors, whiteboards, chairs, and benches in the training environment are perturbed as shown in Figure~\ref{fig:voila_perturbed}. We see that, with such environmental changes, \voila{} is able to successfully patrol the hallway without any collisions. Second, we deploy \voila{} on a different floor within the same building, with major visual and structural differences from the training environment as shown in Figure~\ref{fig:voila_4th_floor}. While the robot succeeds for much of the trajectory, this experiment demonstrates the limitations of the current approach, as the robot collides with the walls in two places where there are large visual differences from the training environment. 

To quantify the imitation learning performance of \voila{}, we compute the Hausdorff distance between the human demonstrated trajectory and the trajectory generated by the policy learned using \voila{}, provided in Table \ref{table:voila_1}. The Hausdorff distance between the human demonstration and \voila{} is 0.783, whereas the \voila{}-w/o-mismatch agent trained using demonstrations without any egocentric viewpoint mismatch achieves a Hausdorff distance of 0.665. To provide context for the Hausdorff distance metric, we also compute the Hausdorff distance for suboptimal and random trajectories. The suboptimal trajectory was collected by navigating along the hallway in a zig-zag route using \texttt{move\_base} by setting waypoints closer to the walls and achieves a Hausdorff distance of 0.806. The random trajectory was collected using a randomly initialized policy, which fails quickly by crashing in the environment, achieving a Hausdorff distance of 1.192. We see that the \voila{} agent is able to successfully imitate the expert's video-only navigation demonstration and patrol the hallway without any collisions, addressing $Q_1$ and $Q_2$.

\begin{figure}[!tb]
    \centering
    \includegraphics[width=\columnwidth]{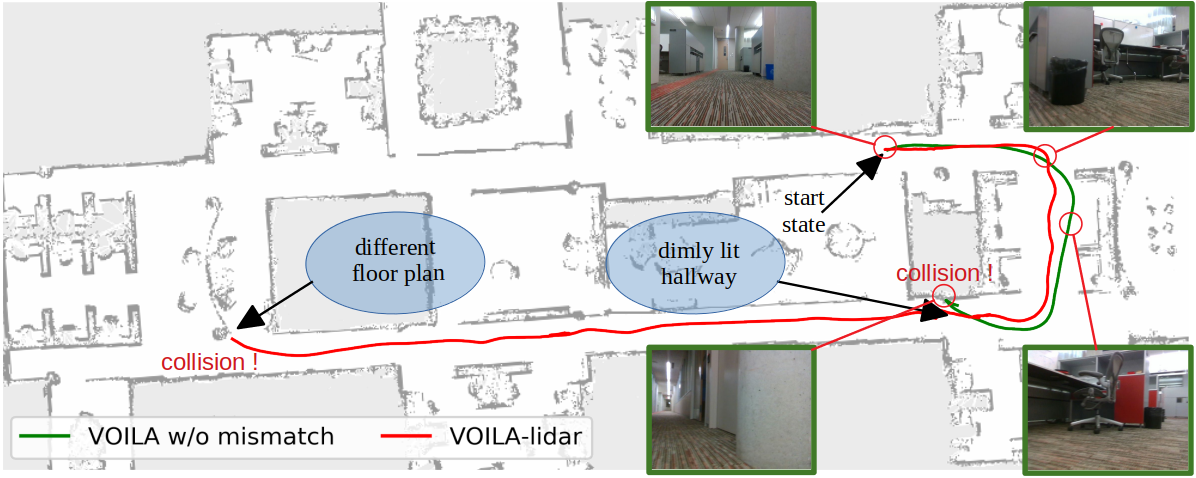}
    \caption{Deploying the policies learned using \voila{} without viewpoint mismatch in an environment with major visual (dimly-lit) and structural (floor plan) differences. While the agent succeeds for much of the trajectory, the \voila{} policy fails to fully generalize and patrol the hallway.}
    \label{fig:voila_4th_floor}
\end{figure}



\subsubsection{Learning a LiDAR-conditioned policy using VOILA}
To evaluate whether \voila{} supports sensing modalities other than vision in the navigation policy's observation space ($Q_3$), we train \voila{} with \textsc{lidar} range scan as the policy's input, and observed that the agent learns to imitate the expert within 30 minutes of experiment time. As shown in Figure~\ref{fig:voila_lidar}, in red, \voila{}-lidar shows the rollout trajectory of the policy learned using \voila{} with \textsc{lidar} range scans. 
The Hausdorff distance between trajectories of the corresponding expert demonstration and the different policies learned using \voila{} are shown in Table~\ref{table:voila_1}. To provide context for the Hausdorff distance metric, we also show results for suboptimal and random trajectories. The suboptimal trajectory was collected by navigating along the hallway in a zig-zag route using \texttt{move\_base}, and the random trajectory was collected using a randomly initialized policy $\pi$, which fails quickly. We see that both \voila{} policies performed well, and that the \textsc{lidar}-based policy provides better performance, which is consistent with findings by \cite{gaifopaper} using \gaifo{}.

In summary, in the physical experiments performed on the Jackal robot, we observed that 1) \voila{} is indeed capable of learning imitative policies from video-only demonstrations even in the presence of significant viewpoint mismatch (addressing $Q_1$), 2) a policy learned using \voila{} does generalize to unseen environments (addressing $Q_2$), and 3) \voila{} does work with sensor modalities other than vision, such as lidar range scans (addressing $Q_3$).

 \begin{figure*}[!tb]
    \centering
    \includegraphics[width=\columnwidth]{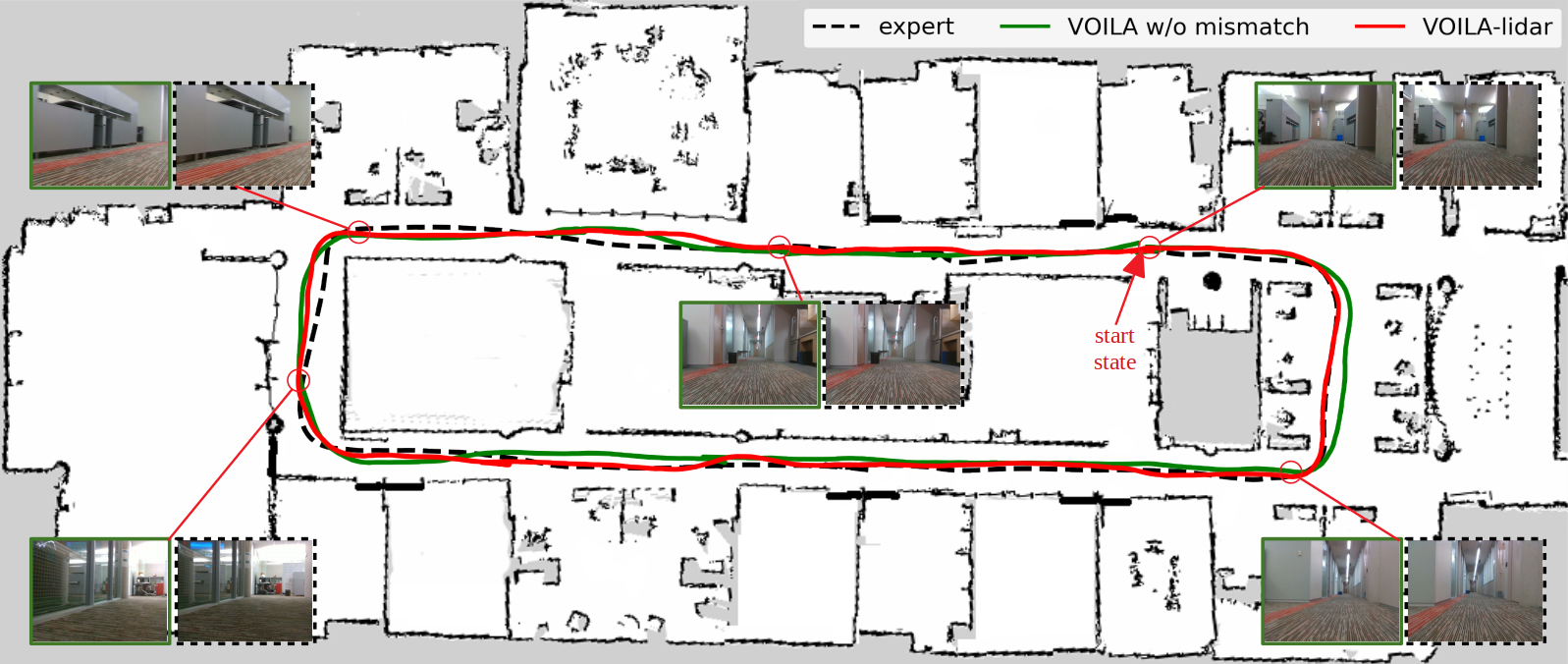}
    \caption{Policy rollouts of the \voila{}-lidar agent trained with lidar range scans as the state input to the policy. We see that the \voila{}-lidar agent succeeds in imitating the expert demonstration, by patrolling the hallway without collisions.}
    \label{fig:voila_lidar}
\end{figure*}

\begin{table}
\centering
\caption{Hausdorff distance between the expert trajectory and the policy rollout trajectory of \voila{}. \voila{}-lidar performs the best, followed by \voila{}-vision.}

\begin{tabular}{ ||c|c|c|| } 
 \hline
 Expert & Policy & Hausdorff Distance \\ 
 \hline \hline
  human demo & \voila{} w/ mismatch &  \textbf{0.783} \\ 
  \hline \hline
 \texttt{move\_base} & \voila{}-lidar w/o mismatch &  \textbf{0.487} \\ 
  \hline
 \texttt{move\_base} & \voila{} w/o mismatch &  0.665 \\ 
  \hline
 \texttt{move\_base} & Suboptimal &  0.806 \\
  \hline
 \texttt{move\_base} & Random &  1.192 \\ 
 \hline
\end{tabular}
\label{table:voila_1}
\end{table}

\section{Summary}
\label{sec:summary_voila}

In this chapter, I introduced Visual-Observation-only Imitation Learning for Autonomous navigation (\textsc{voila}), a novel approach that enables imitation learning for autonomous robot navigation using a single, egocentric, video-only demonstration while being robust to egocentric viewpoint mismatch, thereby enabling navigation behaviors that are in alignment with the demonstrator.
\textsc{voila} formulates the imitation problem as one of reinforcement learning using a novel reward function that is based on keypoint matches between the expert and imitator's visual observations.
Through experiments, both in simulation and on a physical robot, it is observed that, by optimizing the proposed reward function using reinforcement learning, \voila{} could successfully find a good imitation policy that maps sensor observations directly to low-level action commands.
Additionally, I show experiments that were performed to test the generalizability of policies trained using \voila{} to unseen environments.

This chapter effectively concludes the first contribution of this dissertation, as outlined in Section \ref{contrib1}, on visual end-to-end imitation learning for autonomous navigation. In the subsequent chapter, I delve into the second contribution focusing on terrain representation learning and the alignment of robot navigation behaviors with operator preferences in off-road autonomy.

\chapter{Preference-Aligned Off-Road Navigation}
\label{chap:pref_learning_offroad_nav}

In this chapter, I introduce the second contribution of this dissertation, which comprises two algorithms, namely \sterling{} \citep{sterling} and \patern{} \citep{patern}. These algorithms enable aligning a mobile robot's off-road navigation behavior with an operator's terrain preferences in off-road conditions by leveraging self-supervised representation learning. The first algorithm, \sterling{}, is focused on learning relevant terrain representations in a self-supervised manner, enabling operator-preference-conditioned mobile robot navigation in off-road environments. The second algorithm, \patern{}, extends this concept through a self-supervised learning framework designed to extrapolate operator preferences onto visually novel terrains. The structure of this chapter is as follows. Section \ref{sec:intro_prefoffroad} introduces the specific problem this chapter addresses. Section \ref{sec:relatedwork_offroad} provides a review of relevant literature, followed by foundational concepts outlined in Section \ref{sec:prelimsoffroadnav}. Sections \ref{sec:sterling} and \ref{sec:patern} present the two algorithmic contributions, \sterling{} and \patern{} respectively.

\section{Introduction}
\label{sec:intro_prefoffroad}

\textit{Terrain awareness}, defined as the ability to identify distinct terrain features that are relevant to a wide variety of downstream tasks (e.g., changing preferences over terrain types) is a particularly difficult challenge in off-road autonomous navigation \citep{paternarxiv, paternicra, vrlpap, yao2022rca, ser, viikd, atreya2022highspeed, scand}. Existing approaches typically rely on difficult-to-collect curated datasets \citep{rugd, jiang2020rellis3ddataset, catdataset, ganav, triest2022tartandrive} or have been focused on particular tasks \citep{bojarskie2e, offroadoa, wulfmeier2015maximum, byronboots, kahn2021land, viikd} and is not amenable to downstream task changes \citep{kahn2021badgr, kahn2021land, yao2022rca}. These limitations prevent existing approaches from appropriately scaling to the vast distribution of terrains and navigation tasks in the real world.

This necessity for terrain awareness stems not only from direct implications for robot functionality but also from operator-indicated terrain preferences that the robot must adhere to. Often, these preferences are motivated by the desire the protect delicate landscapes, such as flower beds, or to mitigate potential wear and tear on the robot by avoiding hazardous surfaces. However, during autonomous operations, ground robots frequently face unfamiliar terrains~\citep{viikd, marcoanymal} and dynamic real-world conditions such as varied lighting that lie outside the distribution of visually recognized terrains where operator preferences have been pre-defined. This mismatch presents significant challenges for vision-based outdoor navigation \citep{xuesusurvey}.

Prior approaches to addressing the preference-aligned path planning problem include collecting expert demonstrations on diverse environments \citep{bojarskie2e, offroadoa, vrlpap}, gathering pixel-wise human annotations and labels for off-road data \citep{ganav, rugd, jiang2020rellis3ddataset}, and utilizing hand-coded reward functions to assign traversability costs \citep{yao2022rca, kahn2021badgr, terrapn}. While these approaches have been successful at visual navigation, collecting expert demonstration data in the field and labeling may be labor-intensive and expensive, and utilizing hand-coded reward functions may not always align with operator preferences. We posit that in certain cases, while the terrain may look visually distinct in comparison to prior experience, similarities in the inertial-proprioceptive-tactile space may be leveraged to extrapolate operator preferences over such terrains that the robot must adhere to. For instance, assuming a robot has traversed both $\texttt{concrete pavement}$ and $\texttt{marble rocks}$, and prefers the former over the latter (as expressed by the operator), when the robot experiences a visually novel terrain such as $\texttt{pebble pavement}$ which feels inertially similar to traversing over $\texttt{concrete pavement}$, it is more likely that the operator might also prefer $\texttt{pebble pavement}$ over $\texttt{marble rocks}$. While it is not possible to know the operator's true preferences without querying them, we submit that, in cases where the operator is unavailable, hypothesizing preferences through extrapolation from the inertial-proprioceptive-tactile space is a plausible way to estimate traversability preferences for novel terrains.

First, toward overcoming the scalability challenge in learning to be terrain aware, we introduce \textit{Self-supervised TErrain Representation LearnING} (\sterling{}),\footnote{A preliminary version of this work was presented at the PT4R workshop at ICRA 2023 \citep{sterlingicra}. The final version was presented at CoRL 2023 \citep{sterling}.} a novel approach to learning terrain representations for off-road navigation. \sterling{} learns an encoding function that maps high-dimensional, multi-modal sensor data to low-dimensional, terrain-aware representations that amplify differences important for navigation and attenuate differences due to extraneous factors such as changes in viewpoint and lighting. Importantly, \sterling{} works with easy-to-collect unconstrained and unlabeled robot data, thereby providing a scalable pathway to data collection and system improvement for the wide variety of terrain and downstream tasks that off-road robots must face. To evaluate \sterling{}, we apply it to the problem of preference-aligned off-road navigation and provide a detailed comparison to existing approaches to this problem, including \rca{} \citep{yao2022rca}, \ganav{} \citep{ganav}, \ser{} \citep{ser}, and a fully-supervised oracle. We find that \sterling{} enables performance on par with or better than these existing approaches without requiring any expert labels or demonstrations. Additionally, we report the results of a large-scale qualitative experiment in which \sterling{} enabled semi-autonomous robot navigation on a 3-mile long hiking trail.

Second, leveraging the intuition of extrapolating operator preferences for visually distinct terrains that are familiar in the inertial-proprioceptive-tactile space (collectively known as \textit{proprioceptive} for brevity), we introduce \textit{Preference extrApolation for Terrain-awarE Robot Navigation} (\patern{}), \footnote{\patern{} is accepted for publication at ICRA 2024.} a novel framework for extrapolating operator terrain preferences for visual navigation. \patern{} learns a proprioceptive latent representation space from the robot's prior experience and uses nearest-neighbor search in this space to estimate operator preferences for visually novel terrains. Figure \ref{fig:patern_intuition} provides an illustration of the intuition behind preference extrapolation in \patern{}. We conduct extensive physical robot experiments on the task of preference-aligned off-road navigation, evaluating \patern{} against state-of-the-art approaches, and find that \patern{} is empirically successful with respect to preference alignment and in adapting to novel terrains and lighting conditions seen in the real world. \footnote{The research described in this chapter was done in collaboration with Elvin Yang, Daniel Farkash, Garrett Warnell, Joydeep Biswas, and Peter Stone.} 


\noindent\textbf{Relation between STERLING and PATERN:} \sterling{} is a self-supervised representation learning algorithm designed to learn terrain representations solely from a robot's unconstrained experiences that can better enable the downstream task of operator preference-aligned off-road navigation. To gather preferences, however, \sterling{} necessitates gathering operator feedback for each new terrain that the robot encounters which may not always be feasible. \patern{}, in contrast, addresses this gap by leveraging a simple intuition. Instead of querying an operator for all new terrains encountered, \patern{} complements \sterling{} by inferring preferences through extrapolation from known to visually novel terrains, utilizing non-visual terrain data observed by the robot. \sterling{} is most effective in cases where the robot is unlikely to come across terrains beyond its training set, or where operator preferences are readily obtainable. However, in situations where the robot may face visually novel terrains or obtaining operator feedback for novel terrains is challenging, \patern{} is a necessary extension to \sterling{} that enables extrapolating operator preferences, thereby complementing \sterling{} and reducing the dependence on direct operator feedback. Therefore, \patern{} does not completely replace \sterling{}, however, it acts as a wrapper over \sterling{}, extending its capabilities beyond the observed set of terrains. To summarize, \sterling{} is a self-supervised terrain representation learning algorithm that utilizes operator feedback of preferences for all encountered terrains to enable preference-aligned off-road navigation, whereas \patern{} complements \sterling{} by deducing operator preferences for certain visually novel terrains, thereby reducing the dependency on direct operator feedback.

\begin{figure}
        \centering
        \includegraphics[width=\textwidth]{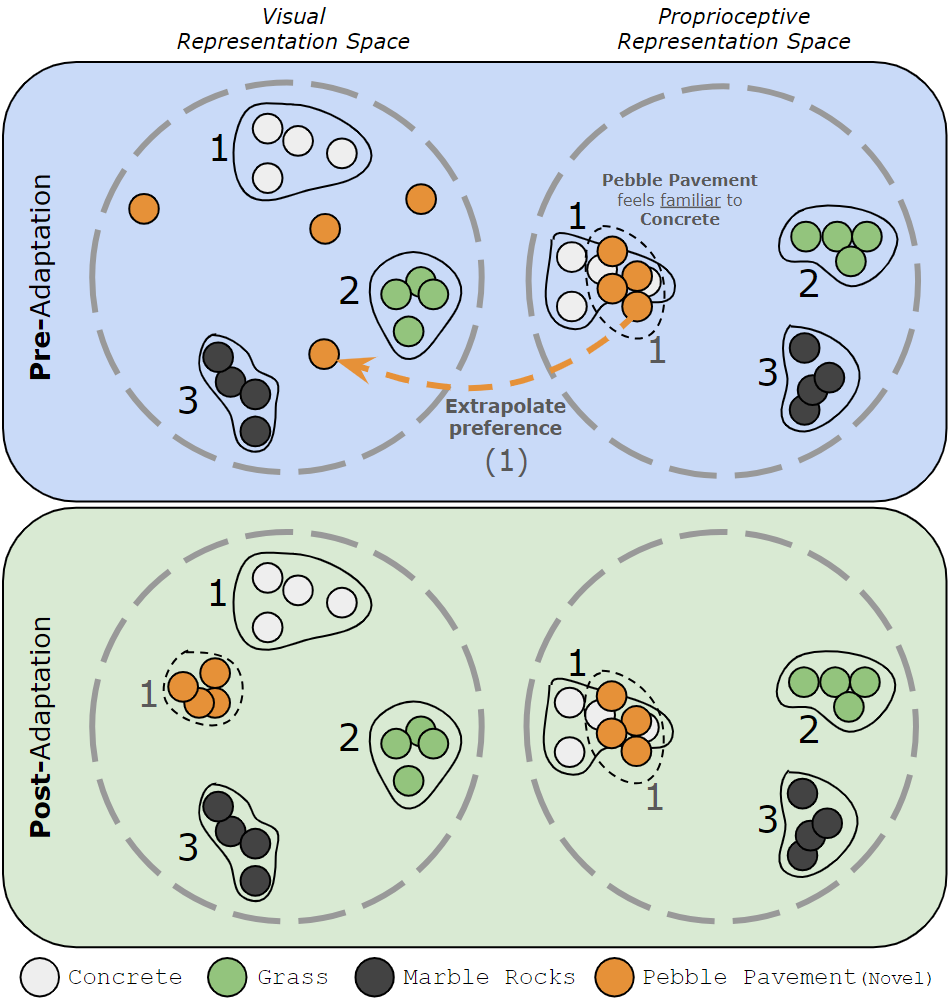}
        \caption{An illustration of the intuition behind preference extrapolation in \patern{}. Operator preferences of the three known terrains are numerically marked, with 1 being the most preferred and 3 being the least preferred. In the pre-adaptation stage, a novel terrain ($\texttt{pebble pavement}$) is encountered and the preference order of its nearest neighbor ($\texttt{concrete}$) inferred from proprioceptive representations is transferred (extrapolated) to the corresponding samples in the visual representation space. The extrapolated preference order is used to update both the visual representations and the visual preference function. In the post-adaptation stage, the figure illustrates the extrapolated preferences in the updated visual representation space for the novel terrain.}
        \label{fig:patern_intuition}
\end{figure}

\section{Related Work}
\label{sec:relatedwork_offroad}

In this section, I review related work on terrain-aware visual off-road navigation. I specifically focus on approaches that learn to navigate off-road conditions using supervised and self-supervised learning with a focus on operator-preference-aligned planning.

\subsection{Supervised Methods}
Several approaches in the past have proposed using supervised learning from large-scale data to navigate off-road environments. We divide them into two categories as follows.

\textbf{End-to-End Learning:}
The initial success of applying learning-based solutions to off-road terrain-aware navigation was by \cite{offroadoa} who used a convolutional network to learn to drive in off-road conditions. More recently, \cite{bojarskie2e} trained a deep neural network end-to-end using several miles of driving data collected on a vehicle in the real world. While both approaches were promising in urban and off-road environments, end-to-end methods require large amounts of data and are well-known to suffer from domain and covariate shifts \citep{xuesusurvey, voila, barnchallenge}.

\textbf{Image Segmentation:} Unlike end-to-end approaches that learn behaviors, segmentation-based approaches seek to characterize terrain using a set of known semantic classes, and the resulting semantic features are consumed by downstream planning and control techniques for navigation \citep{schilling2017geometric, ganav, wigness2018robot}. \cite{ganav} propose \ganav{}, a transformer-based architecture to pixel-wise segment terrains, trained on \rellis{} \citep{jiang2020rellis3ddataset} and \rugd{} \citep{rugd} datasets, with manually assigned terrain costs. While effective at terrain awareness, segmentation-based methods are fixed to the specific terrain types available in the datasets and require additional labeling effort to generalize to novel terrains. On the other hand, \sterling{} does not require semantically labeled datasets and learns terrain representations from unconstrained experience collected onboard a mobile robot.

\subsection{Self-Supervised Learning}
To alleviate the need for extensive human labeling, self-supervised learning methods have been proposed to either learn terrain representations or costs from data gathered onboard a mobile robot.

\textbf{Representation Learning:}
\cite{brooks2007self} utilize contact vibrations and visual sensors to classify terrains via self-supervision. \cite{loquercio2022learning} use proprioceptive supervision to predict extrinsic representations \citep{vgaifoso} of terrain geometry from vision, used as inputs to drive a Reinforcement Learning-based locomotion policy. In both contributions in this Chapter, we do not learn a robot-specific locomotion policy and instead learn relevant representations for off-road terrain awareness. \cite{ser} introduce \ser{} which utilizes acoustic and visual sensors on the robot to segment terrains using a self-supervised triplet-contrastive learning framework. Using triplet-based contrastive learning methods requires negative samples which may not be available when learning using unlabeled data. In \sterling{}, I propose using non-contrastive unsupervised learning approaches known as \vicreg{} \citep{bardes2021vicreg} that do not require any negative samples and instead rely on correlations between data modalities to learn relevant terrain representations.

\textbf{Cost Learning:}
Several methods have applied self-supervision to assign traversability costs for the downstream off-road navigation task \citep{yao2022rca, terrapn, kahn2021badgr, wher2walk, castro2023does, triest2023learning, chen2023learningonthedrive}. Specifically, these methods rely on inertial spectral features \citep{yao2022rca}, future predictive models \citep{kahn2021badgr}, inertial-odometry errors \citep{terrapn}, or force-torque values from foothold positions \citep{wher2walk, anomalydetect} as self-supervision signals to learn a traversability cost map, used to evaluate candidate actions. More recently, \cite{frey2023fast} proposed an online traversability estimation approach inspired by the above self-supervision schemes. Instead of inferring costs or rewards using self-supervision for a fixed task, the first contribution in this chapter, \sterling{}, focuses on learning relevant visual features from unconstrained robot experiences that could be used in downstream tasks. This framework allows a designer to reuse features across tasks without retraining entirely from scratch. Additionally, contrasting with prior methods using hand-coded reward/cost functions that may not adhere to operator preferences, \patern{} utilizes the prior experience of the robot and extrapolates operator preferences to novel terrains.

\textbf{Hybrid Methods:}
The approach closest to ours is \vrlpap{} 
\citep{vrlpap} which requires human expert teleoperated demonstrations of a particular trajectory pattern to both explicitly learn visual terrain representations as well as to infer terrain preference costs. However, the first contribution in this chapter, \sterling{}, focuses on learning terrain features from unconstrained robot experiences without requiring human experts in the field for demonstrations, which is a more general problem than the one considered by \vrlpap{}. Additionally, \patern{} focuses on extrapolating operator preferences without additional human feedback, which is not supported in \vrlpap{}.

\section{Preliminaries}
\label{sec:prelimsoffroadnav}
We formulate preference-aligned planning as a local path-planning problem in a state space $\mathcal{S}$, with an associated action space
$\mathcal{A}$. The forward kino-dynamic transition function is denoted as $\mathcal{T}:
\mathcal{S} \times \mathcal{A} \rightarrow \mathcal{S}$ and we assume
that the robot has a reasonable model of $\mathcal{T}$ (\eg{} using parametric system identification~\citep{seegmiller2013vehicle} or a learned kino-dynamic
model~\citep{xuesuikd,viikd,optimfkd}), and that the robot can execute actions
in $\mathcal{A}$ with reasonable precision. 
For ground vehicles, a common choice for $\mathcal{S}$ is $\mathrm{SE}(2)$, which represents the robot's x and y position on the ground plane, as well as its orientation $\theta$.

The objective of the path-planning problem can be expressed as finding the optimal trajectory
$\Gamma^* = \underset{\Gamma}{\arg \min}\ J(\Gamma, G)$ to the goal $G$, using any planner (e.g. a sampling-based motion planner like \textsc{dwa} \citep{dwa}) while minimizing an objective function $J(\Gamma, G)$, $J : (\mathcal{S}^N, \mathcal{S}) \rightarrow \mathbb{R}^+$. Here, $\Gamma = \{s_1, s_2, \dots, s_N\}$ denotes a sequence of states. The sequence of states in the optimal trajectory $\Gamma^*$ is then translated into a sequence of actions, using a 1-D time-optimal controller, to be played on the robot. For operator preference-aligned planning, the objective function $J$ is articulated as,
\begin{equation}
    J(\Gamma, G) = J_G(\Gamma(N), G) + J_P(\Gamma),
\end{equation}
\noindent

Here, $J_G$ denotes cost based on proximity of the robot's state to the goal $G$, while $J_P$ imparts a cost based on terrain preference. 
Crucially, $J_P$ is designed to capture operator preferences over different terrains; less preferred terrains incur a higher cost. Earlier studies such as \vrlpap{} \citep{vrlpap} leverage human feedback in the form of geometrically constrained trajectories to ascertain $J_P$. In the contributions of this chapter, I utilize a two-step framework to determine $J_P$. In the first contribution of this chapter as detailed in Section \ref{sec:sterling}, \sterling{} uses a self-supervised training recipe to first learn relevant terrain representations followed by operator preference queries to learn a mapping from visual observations of terrains to representations followed by a mapping from representations to preference cost value, used for planning. In the second contribution detailed in Section \ref{sec:patern}, \patern{} extrapolates preference values $J_P$ of known terrains to visually novel terrains through a self-supervised learning process.

\section{Self-Supervised Terrain Representation Learning}
\label{sec:sterling}
In this section, I introduce \sterling{}, a new approach for finding terrain representations using self-supervised learning. I first describe the offline pre-processing performed on unconstrained robot data and then summarize the self-supervision objectives. Finally, I describe the problem formulation for preference-aligned off-road navigation and present how features learned using \sterling{} can be utilized within a planner for terrain-aware and preference-aligned navigation.

\subsection{Data-Collection and Pre-Processing}
\label{sec:data_processing}

\sterling{} learns terrain representations from unconstrained, unlabeled robot experiences collected using any navigation policy. This policy may be, for instance, non-expert human teleoperation, curiosity-driven exploration \citep{pathak2017curiositydriven}, or point-to-point navigation using any underlying planner. Compared to requiring a human expert to provide teleoperated demonstrations and labels, collecting this type of robot experience is cheap and easy, thereby providing a scalable pathway to data collection and system improvement. We additionally assume that the robot is equipped with {\em multiple sensors}, e.g., an egocentric RGB camera, odometry sensors, onboard IMU, proprioceptive, and tactile sensors, that together provide rich multi-modal observations as the robot traverses over different terrains collecting experience. \sterling{} leverages this multi-modal data by using the correlation between different modalities to inform the learned terrain representations.

In order to learn terrain representations using \sterling{}, we begin by pre-processing the visual and non-visual observations, which are explained in detail below.

\noindent\textbf{Visual Patch Extraction: } The egocentric camera observations are homography-projected into a virtual bird's eye view (BEV) frame, assuming that the ground is a flat plane, using the intrinsic and extrinsic camera matrices. As shown in Figure \ref{fig:patch_extraction}, we project the robot's trajectory onto the BEV frame and extract 64-by-64 pixels (equivalent to the robot's footprint of 0.5-by-0.5 meters) square visual patches of terrain along with the corresponding inertial, proprioceptive, and tactile observations at the same location, along the trajectory.
Since the terrain at $s_k$ is unobservable when the robot itself is at $s_k$ (i.e., it is underneath the robot), we extract terrain image patches corresponding to $s_k$ from BEV observations at previous locations $s_{k-1}, s_{k-2}, \ldots$ along its trajectory. Figure \ref{fig:patch_extraction} illustrates the offline patch extraction process from two previous viewpoints, however, we extract patches from up to 20 previous viewpoints within 2 meters.
Although just one viewpoint is sufficient to learn the correlation between visual and other sensor observations, when planning to navigate, the robot will need to visually evaluate terrain at future locations, and therefore \sterling{} also seeks representations that are invariant to patch differences due to viewpoint, also known as \textit{viewpoint invariance}.

\begin{figure*}[!ht]
    \centering
    \includegraphics[width=\columnwidth]{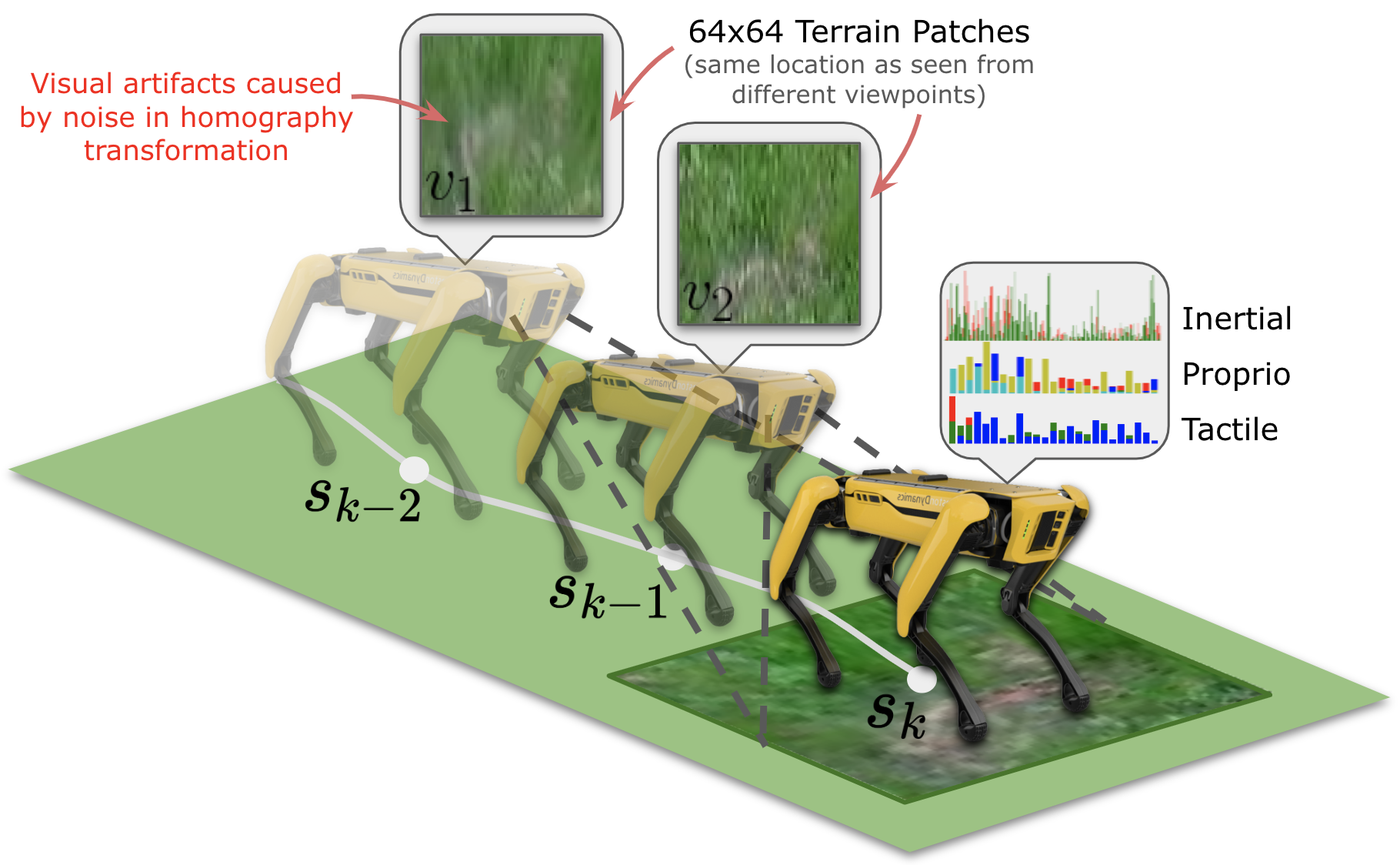}
        \caption{An illustration of the pre-processing performed on unconstrained robot experience. Image patches of traversed terrain at location $s_k$ are extracted from bird's eye view observations at prior locations $s_{k-1}, s_{k-2}$ along the trajectory. The corresponding \textsc{ipt} observations at $s_k$ are transformed from time series to \textsc{psd} signals. Note the visual artifacts caused by noise in homography transformation from viewpoints farther away from $s_k$.}
        \label{fig:patch_extraction}
\end{figure*}

\noindent\textbf{IPT Pre-Processing:} For the inertial, proprioceptive, and tactile (\textsc{ipt}) observations, we retain up to 2-second history and convert the time-series signals into power-spectral density (\psd{}) representation in the frequency domain. This ensures the \textsc{ipt} time-series data representations used as input to \sterling{} are invariant to differences in length and phase in the recorded signals. 

\noindent\textbf{Data Collection:}
In all experiments, we use a legged Boston Dynamics Spot robot and collect robot experiences on eight different types of terrain around the university campus that we labeled as \texttt{mulch}, \texttt{pebble sidewalk}, \texttt{cement sidewalk}, \texttt{grass}, \texttt{bushes}, \texttt{marbled rock}, \texttt{yellow bricks}, and \texttt{red bricks}. The data is collected through human teleoperation such that each trajectory contains a unique terrain throughout, with random trajectory shapes. Note that \sterling{} does not require a human expert to teleoperate the robot to collect robot experience nor does it require the experience to be gathered on a unique terrain per trajectory. We follow this data collection approach since it is easier to label the terrain for evaluation purposes. \sterling{} can also work with random trajectory lengths, with multiple terrains encountered along the same trajectory, without any semantic labels such as terrain names, and any navigation policy can be used for data collection. We record 8 trajectories per terrain, each five minutes long, and use 4 trajectories for training and the remaining for validation.

\subsection{Sampling-based Planning}

We assume access to a receding horizon sampling-based motion planner with a fixed set of constant-curvature arcs $\{\Gamma_0, \Gamma_1, \dots, \Gamma_{ns}\}$, $\Gamma \in \mathcal{S}^N$ which solves for the optimal arc $\Gamma^* = \underset{\Gamma}{\arg \min} [\mathcal{J}(\Gamma, G)]$, minimizing the objective function $\mathcal{J}(\Gamma, G), \mathcal{J}: (\Gamma, G) \longrightarrow \mathbb{R}^+$. For the task of preference-aligned off-road navigation, we assume the objective function is composed of two components $\mathcal{J}_{geom}(\Gamma, G)$ and $\mathcal{J}_{terrain}(\Gamma)$, and can be defined as $
    \mathcal{J}(\Gamma, G) = \alpha \mathcal{J}_{geom}(\Gamma, G) + (1-\alpha) \mathcal{J}_{terrain}(\Gamma)
$. $\mathcal{J}_{geom}(\Gamma, G)$ is the geometric cost that deals with progress towards the goal $G$ and avoiding geometric obstacles, whereas $\mathcal{J}_{terrain}(\Gamma)$ is the terrain cost associated with preference-alignment. We utilize the geometric cost as defined in AMRL's graph navigation stack \footnote{\href{https://github.com/ut-amrl/graph_navigation}{https://github.com/ut-amrl/graph\_navigation}}. The multiplier $\alpha \in [0, 1]$ trades off relative contributions of the geometric and terrain preference components of the path planning objective. A 1D time-optimal controller translates the sequence of states in the optimal trajectory $\Gamma^*$ to a sequence of receding horizon actions $(a_0, a_1, \dots, a_N)$. For a given arc $\Gamma = \{s_0, s_1, \dots, s_N\}$, such that state $s_0$ is closest to the robot, the terrain-preference cost can be computed as follows.
\begin{equation}
    \mathcal{J}_{terrain}(\Gamma) = \sum\limits_{v_i \sim \Gamma, i=0}^{N} \frac{\gamma^{i} C(u(f_v(v_i)))}{N+1}
    \label{eq:terrain_cost}
\end{equation}
The function $f_v(.)$ maps from RGB space of a visual patch of terrain $v_i$ at a specific state $s_i$, to its visual representation $\phi_v \in \Phi_v$. For instance, $f_v$ can be the visual encoder learned using \sterling{}, as described in Section \ref{sec:sterling_learning}. The utility function $u(.)$ maps the visual representation $\phi_v$ of a patch of terrain to a real-valued utility of preferences. We follow the utility function formulation of \cite{zucker2011optimization} and assume the terrain preference cost follows a multiplicative formulation such that given a utility value $x \in \mathbb{R}^+$, the traversability cost is $C(x)=e^{-x}$. The discount factor $\gamma$ weighs the terrain cost proportional to its proximity to the robot. We set $\gamma$ to $0.8$, which we find to work well in practice. 

\label{sec:sampling_based_planning}
\begin{figure*}[!ht]
    \centering
        \includegraphics[width=\columnwidth]{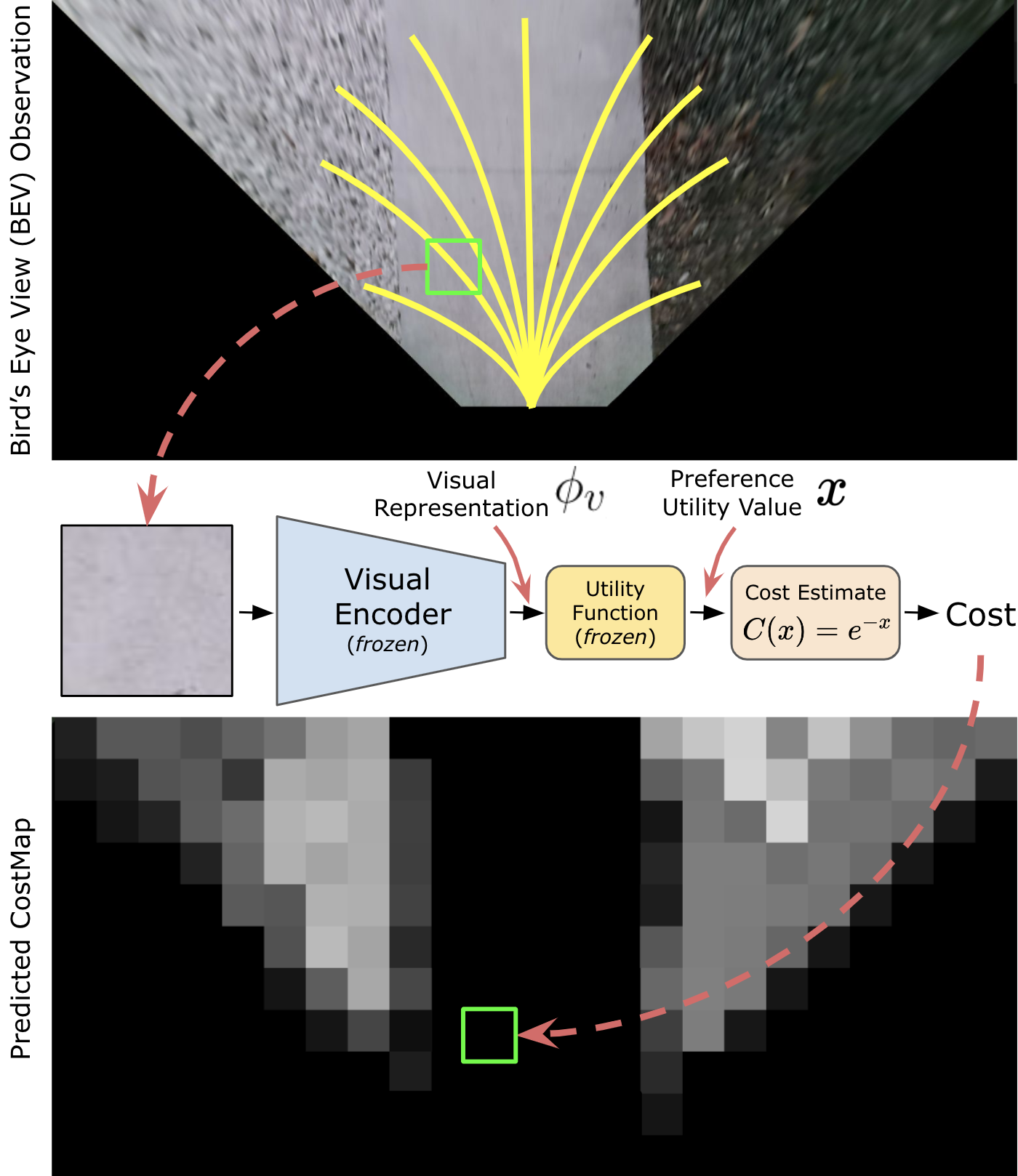}
        \caption{An overview of the cost inference process for local planning at deployment. The constant-curvature arcs (yellow) are overlayed on the BEV image, and the terrain cost $\mathcal{J}_{terrain}(\Gamma)$ is computed on patches extracted along all arcs. White is high cost and black is low cost.}
        \label{fig:deployment}
\end{figure*}

\noindent\textbf{Planning at Deployment:}
\label{sec:planning_at_deployment}
Figure \ref{fig:deployment} provides an overview of the cost inference process for local planning at deployment. To evaluate the terrain cost $\mathcal{J}_{terrain}(\Gamma)$ for the constant-curvature arcs, we overlay the arcs on the bird's eye view image, extract terrain patches at states along the arc, and compute the cost according to Equation \ref{eq:terrain_cost}. We compute the visual representation, utility value, and terrain cost of all images at once as a single batch inference. Since the visual encoder and the utility function are relatively lightweight neural networks with about 0.5 million parameters, we are able to achieve real-time planning rates of 40 Hz using a laptop-grade Nvidia GPU.

\subsection{Non-Contrastive Terrain Representation Learning}
\label{sec:sterling_learning}

It is desired for learned representations of terrains to be such that representations of similar terrain are close together in the embedding space and that representations of different terrains are sufficiently far apart. Although we do not possess privileged information such as semantic labels of terrains for training, the visual and kinodynamic observations experienced by the robot reflect similarities and differences between terrain samples.
For instance, traversing a smooth terrain that a human may refer to as \texttt{cement sidewalk} may lead to relatively smooth motion by the robot's joints, whereas a rough terrain such as what might be referred to as \texttt{marble rocks} may correspond to jerkier motion. \sterling{} leverages this multi-modal experience observed by the robot and computes a correlation objective between visual and inertial-proprio-tactile signals to learn desired terrain representations. Additionally, \sterling{} uses viewpoint invariance as an objective unique to the visual component of the experience to learn viewpoint-invariant terrain representations.

Figure \ref{fig:sterling_framework} provides an overview of the self-supervised representation learning framework adopted in \sterling{}. A parameterized visual encoder (4-layer CNN with 0.25 million parameters) encodes terrain image patch observations $v_1$ and $v_2$ of the same location $s$ into visual representations $\phi_{v_1}$ and $\phi_{v_2}$, respectively, collectively referred to as $\phi_{v_{1,2}}$ for brevity.
Similarly, an inertial-proprio-tactile encoder (4-layer MLP with 0.25 million parameters) encodes frequency domain \textsc{ipt} observations of the robot at that location to an inertial-proprio-tactile representation $\phi_i$. We follow the framework of prior self-supervised representation learning algorithms from the computer vision community such as \vicreg{} \citep{bardes2021vicreg}, and utilize a parameterized projector network (2-layer MLP with 0.25 million parameters) that maps encoded visual and non-visual representations independently to a higher-dimensional feature space $\psi_{v_{1,2}}$ and $\psi_i$ respectively, over which the self-supervision objectives are computed. The \sterling{} objective composed of the multi-modal correlation $\mathcal{L}_{MM}(\psi_{v_{1,2}}, \psi_i)$ and viewpoint-invariance $\mathcal{L}_{VI}(\psi_{v_1}, \psi_{v_2})$ objectives are defined as:

\begin{equation}
\begin{aligned}
\mathcal{L}_{\sterling{}} &= \mathcal{L}_{VI}(\psi_{v_1}, \psi_{v_2}) + \mathcal{L}_{MM}(\psi_{v_{1,2}}, \psi_i) \\
\mathcal{L}_{VI}(\psi_{v_1}, \psi_{v_2}) &= \mathcal{L}_{\vicreg{}}(\psi_{v_1}, \psi_{v_2}) \\
    \mathcal{L}_{MM}(\psi_{v_{1,2}}, \psi_i) &= [\mathcal{L}_{\vicreg{}}(\psi_{v_1}, \psi_i) + \mathcal{L}_{\vicreg{}}(\psi_{v_2}, \psi_i)] / 2 \\
    \label{eq:sterling_loss}
\end{aligned}
\end{equation}

 $\mathcal{L}_{\vicreg{}}$ is the \vicreg{} loss that is composed of variance-invariance-covariance representation learning objectives, as proposed by \cite{bardes2021vicreg}. Given two alternate projected representations $Z$ and $Z'$ of a data sample (in \sterling{}, $Z$ and $Z'$ are projected representations of the visual and non-visual sensor modalities), the \vicreg{} loss is defined as $\mathcal{L}_{\vicreg{}}(Z, Z')= \lambda s(Z, Z') + \mu [v(Z) + v(Z')] + \nu [c(Z) + c(Z')]$. Note that while \cite{bardes2021vicreg} use \vicreg{} to learn representations from visual inputs using artificial image augmentations, in this work, we extend \vicreg{} to multi-modal inputs and use real-world augmentations via multi-viewpoint image patches as described in Section \ref{sec:data_processing}. $\lambda$, $\mu$, and $\nu$ are hyper-parameters and the functions $v$, $s$, and $c$ are the variance, invariance, and covariance terms computed on a mini-batch of projected features. We refer the reader to \cite{bardes2021vicreg} for additional details on the individual terms and also define them here for completeness. The variance term $v$ is a hinge function defined as $v(Z)=\frac{1}{d}\sum\limits_{j=1}^{d}max(0, \gamma-S(z^j, \epsilon))$, where $S$ is the standard deviation, and $d$ is the dimensionality of the projected feature space. $c$ is the covariance term, defined as $c(Z)=\frac{1}{d}\sum\limits_{i\neq j}[C(Z)]^2_{i,j}$, where $C(Z)$ is the covariance matrix of $Z$. $s$ is the invariance term defined as $s(Z, Z^{'})=\frac{1}{n}\sum\limits_{i}||z_i-z^{'}_i||$. 
 
 While the foundational work by \cite{bardes2021vicreg} uses image augmentations to learn visual representations in a self-supervised way, we utilize images from multiple viewpoints and multi-modal inputs such as vision, inertial, proprioceptive, tactile using a novel formulation to learn relevant terrain representations in a self-supervised way. The \sterling{} loss based on \vicreg{} is defined in Section \ref{sec:sterling_learning}. The \vicreg{} loss is defined as $\mathcal{L}_{\vicreg{}}(Z, Z')= \lambda s(Z, Z') + \mu [v(Z) + v(Z')] + \nu [c(Z) + c(Z')]$, where $\lambda$, $\mu$ and $\nu$ are hyperparameters. We use the values 25.0, 25.0, 1.0 for these hyperparameters respectively, as suggested by \cite{bardes2021vicreg}. $s(Z, Z')$ denotes the invariance between the two inputs. In \sterling{}, this invariance term $s(Z, Z')$ is computed across the two image patches from different viewpoints, and also between the visual and non-visual (\textsc{ipt}) projections. $v(Z)$ denotes the variance across the batch dimension, which we compute for the projections of individual patches and the \textsc{ipt} signals. $c(Z)$ denotes the covariance across the feature dimension which encourages distinct, non-correlative features which we again compute for the projections of individual patches and the \textsc{ipt} signals. We apply an $l^2$ norm on the visual and non-visual features to ensure they are on a hypersphere, which helps improve the quality of learned representations. On a mini-batch of data containing paired terrain image patches and \textsc{ipt} observations, we compute the $\mathcal{L}_{\sterling{}}$ loss and update parameters of the two encoder networks and the shared projector network together using Adam optimizer.


\begin{figure}[t]
        \centering
        \includegraphics[width=\columnwidth]{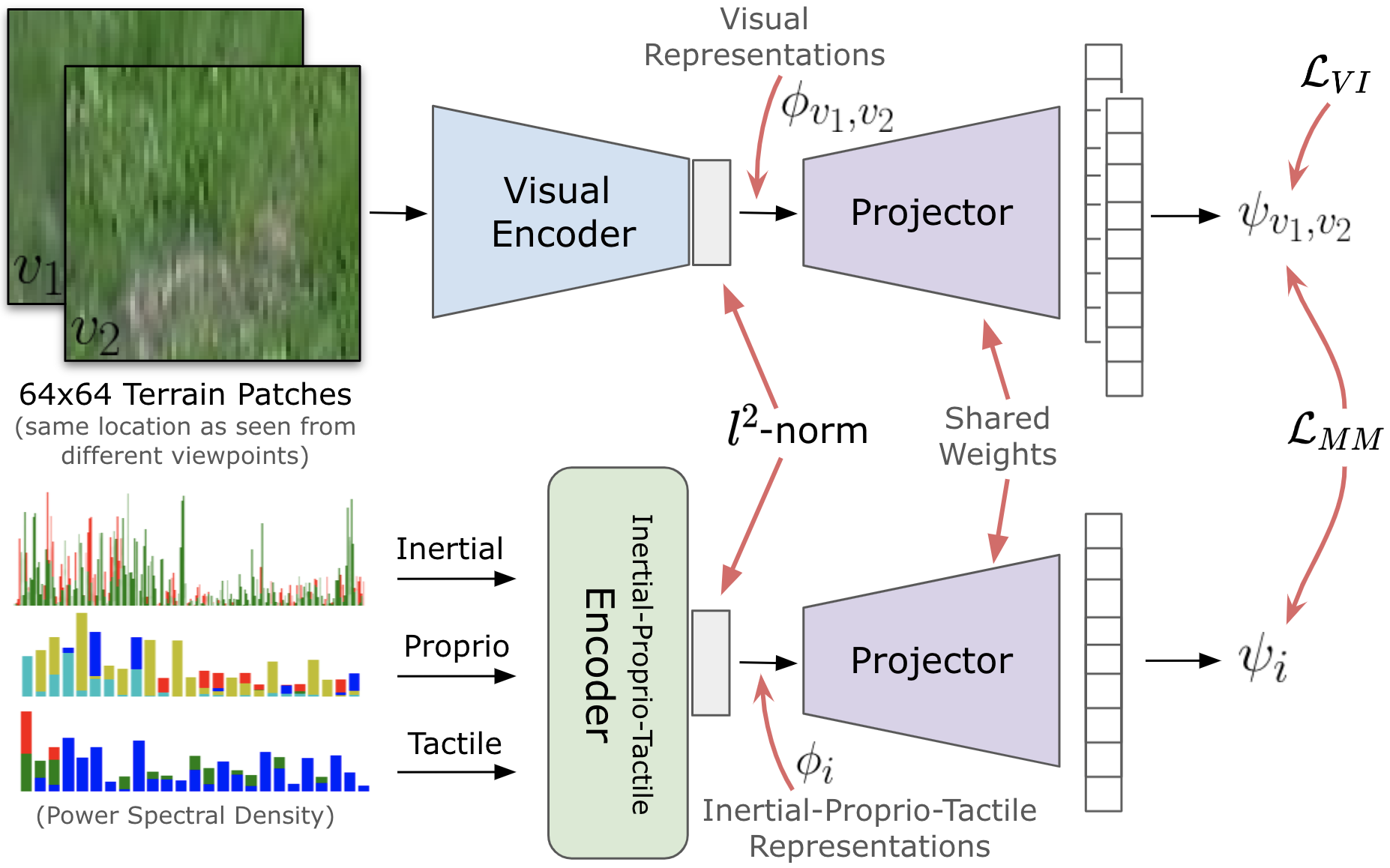}
        \caption{Overview of the training architecture in \sterling{}. Terrain patches $v_1$ and $v_2$ from different viewpoints of the same location are encoded as $\phi_{v_1}$ and $\phi_{v_2}$ respectively, and mapped into embeddings $\psi_{v_1}$ and $\psi_{v_2}$. Similarly, inertial, proprio, tactile signals are encoded as $\phi_i$, and mapped as $\psi_i$. Self-supervision objectives $\mathcal{L}_{VI}$ for viewpoint-invariance and $\mathcal{L}_{MM}$ for multi-modal correlation are computed on the minibatch to perform gradient descent.}
        \label{fig:sterling_framework}
\end{figure}

\subsection{Preference-Aligned Off-Road Navigation}

In this section, I describe the downstream navigation task of preference-aligned visual navigation that we focus on when evaluating \sterling{}.


\noindent\textbf{Preliminaries:} We formulate the task of preference-aligned terrain-aware navigation as a local path-planning problem, where the robot operates within a state space $\mathcal{S}$, action space $\mathcal{A}$, and a deterministic transition function $\mathcal{T}:\mathcal{S} \times \mathcal{A} \longrightarrow \mathcal{S}$ in the environment. The state space consists of $s=[x, y, \theta, \phi_v]$, where $[x,y,\theta]$ denote the robot's position in $SE(2)$ space, and $\phi_v$ denotes the visual features of the terrain at this location. Given a goal location $G$, the preference-aligned navigation task is to reach this goal while adhering to operator preferences over terrains. We assume access to a sampling-based planner, the details of which are provided in Section \ref{sec:sampling_based_planning}.

\noindent\textbf{Learning the preference utility:} Following \cite{zucker2011optimization}, we learn the utility function $u: \Phi_v \rightarrow \mathbb{R}^+$ using human queries. From the predicted terrain features on data samples in our training set, we cluster the terrain representations using k-means with silhouette-score elbow criterion, and sample candidate terrain patches from each cluster, which is presented to the human operator using a GUI. The human operator then provides a full-order ranking of terrain preferences over clusters, which is utilized to learn the utility function $u(.)$, represented by a 2-layer MLP. While recovering absolute cost values from ranked preference orders is an under-constrained problem, we find that this approximation provided by \cite{zucker2011optimization} works well in practice.

\subsection{Experiments}
\label{sec:experimental_results_sterling}

	In this section, I describe the experiments performed to evaluate \sterling{}. Specifically, the experiments presented in this section are tailored to address the following questions:
 
\begin{enumerate}[label=($Q_\arabic*$)]
        \item How effective are \sterling{} features in comparison to baseline approaches at enabling terrain awareness in off-road navigation?
            \item How effective are the proposed \sterling{} objectives in learning discriminative terrain features in comparison to other representation learning objectives?
\end{enumerate}


\noindent We investigate $Q_1$ through physical robot experiments on the task of preference-aligned off-road navigation. We perform quantitative evaluations in six different outdoor environments, and then further perform a large-scale qualitative evaluation by semi-autonomously hiking a 3-mile long off-road trail using preference costs learned using \sterling{} features. To compare various methods, we use the success rate of preference alignment as a metric. If a trajectory followed by any algorithm fails to reach the goal, or at any time traverses over any terrain that is less preferred than any traversed by the operator-demonstrated trajectory, we classify the trial as a failure. We additionally investigate $Q_2$ by comparing \sterling{} against other unsupervised terrain representation learning methods and perform an ablation study on the two \sterling{} objectives.

\noindent \textbf{Baselines:}
To perform quantitative evaluations for $Q_1$, we compare \sterling{} with \ser{} \citep{ser}, \rca{} \citep{yao2022rca}, \ganav{} \citep{ganav}, geometric-only planning \citep{graphnavgithub}, and a fully-supervised baseline. \ser{} and \rca{} perform self-supervised learning from unconstrained robot experience to learn terrain representations and traversability costs respectively, making them relevant baselines for this problem. Since there is no open-source implementation of \rca{}, we replicate it to the best of our abilities. The geometric-only approach ignores terrain costs ($\mathcal{L}_{terrain}$) and plans with geometric cost ($\mathcal{L}_{geom}$) only, making it a relevant ablation on the cost formulation for preference-aware planning. \ganav{}\footnote{\href{https://github.com/rayguan97/GANav-offroad}{https://github.com/rayguan97/GANav-offroad}}  \citep{ganav} is a segmentation-based approach trained on the RUGD \citep{rugd} dataset. We additionally train the fully-supervised baseline in which the terrain cost function is learned end-to-end using supervised learning from linear extrapolation of operator preferences. \ganav{} and the fully-supervised baseline require supervision via terrain labels to learn and hence serve as references for comparison.
We normalize the terrain cost predicted by all methods to be between 0 and 1 for a fair comparison.

\begin{figure*}
    \centering
        \includegraphics[width=\columnwidth]{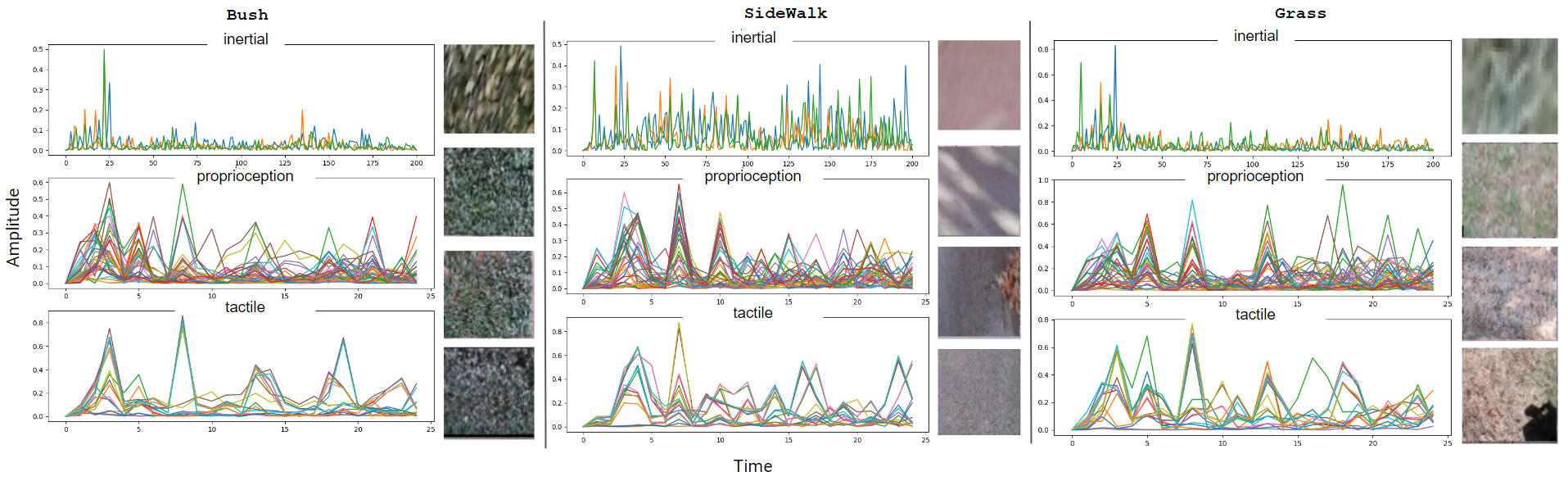}
        \caption{Visual depiction of 1-second of time-series features of inertial-proprioception-tactile data and visual patches of three representative terrains (\texttt{Bush}, \texttt{SideWalk} and \texttt{Grass}).}
        \label{fig:sensor_data}
\end{figure*}

\subsubsection{Experimental Setup and Methodological Details}

In this subsection, we outline the specifics of our experimental setup, detailing hyperparameters, architectural decisions, data, and sensory inputs. These insights ensure the clarity and reproducibility of our experiments.

In all experiments in \sterling{}, including the baselines \rca{} and \ser{}, we use a shallow 4-layer CNN with a kernel size of 3 and stride of 1. Our choice of a shallow 4-layer CNN was driven by the specific need for a lightweight and efficient model that could operate in real time at 40Hz on a laptop GPU, a requirement that was effectively met with this simple architecture of 0.25 M parameters, which we found was sufficient for the problem. While modern architectures like vision transformers / Mobile-ViT could be applied with larger-scale data, the primary concern was real-time performance and compatibility with our robot's hardware. Our experiments and results demonstrate that the selected architecture was sufficient for the purpose, and we do not find evidence that our approach's effectiveness is constrained by this architectural choice.

To train \sterling{}, \ser{}, and \rca{}, we used a total of 117,604 data samples for all terrains combined. Example raw time-series sensor data is shown in Figure \ref{fig:sensor_data}.
Each data sample contains a minimum of 2 visual patches and a maximum of 20 visual patches of the same location from multiple different viewpoints from which we randomly sample 2 patches per location during training. We convert the time-series \textsc{ipt} signals into their corresponding Power Spectral Density \textsc{psd} values. Power spectral density describes the power of a signal across different frequency components. To compute this, we perform a fast-fourier transform over the time-series signal (inertial/proprioceptive/ tactile) and compute the \textsc{psd} defined as $\psd{}(\omega) = \mathbb{E}[|X(\omega)|^2]$ across each frequency component $\omega$. 

On the Spot robot, we use a VectorNav IMU to record the inertial signals (angular velocities in the x and y-axis and linear acceleration in z-axis) at 200Hz, the joint angles and velocities of the legged robot, referred as proprioceptive feedback in this work are recorded at 25 Hz, and the feet contact measurements (contact booleans and estimated feet depth from ground) collectively referred to as tactile feedback in this work are recorded at 25Hz. An Azure kinect camera is mounted on the Spot, used for visual sensing of the terrain. On the wheeled Clearpath Jackal robot, we use a Zed2 camera for visual sensing, and utilize the internal IMU sensor for inertial feedback. Figure \ref{fig:robot_pics} depicts the two robots, sensor mounts, and other sensors used in this work. 

\begin{figure*}
    \centering
        \includegraphics[width=\columnwidth]{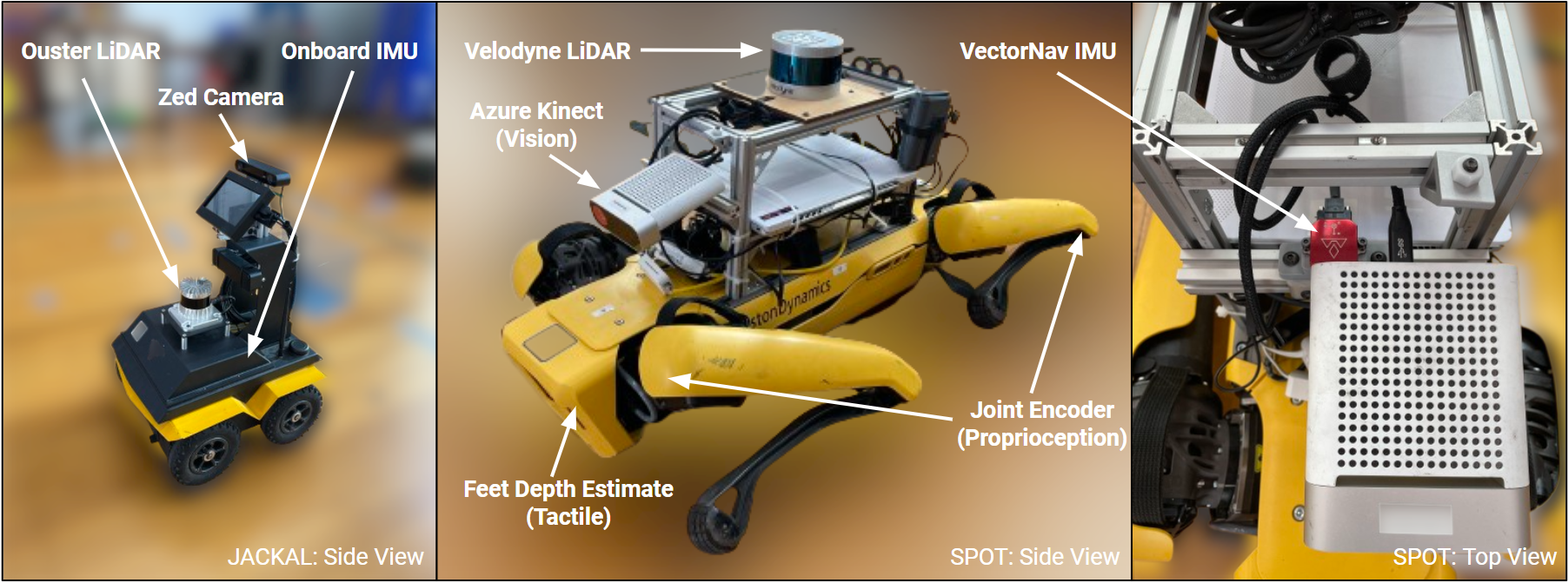}
        \caption{Figure depicting the legged Spot and the wheeled Jackal robot, along with other sensors used, to experimentally evaluate the contributions in this chapter.}
        \label{fig:robot_pics}
\end{figure*}

Note that in all experiments, to prevent overfitting to a specific environment, we pretrain the visual encoder and utility function once and deploy them in all environments. The encoders and utility functions are not being retrained/finetuned per environment, including the large-scale outdoor trail.  

\begin{figure*}[ht]
        \centering
        \includegraphics[width=\columnwidth]{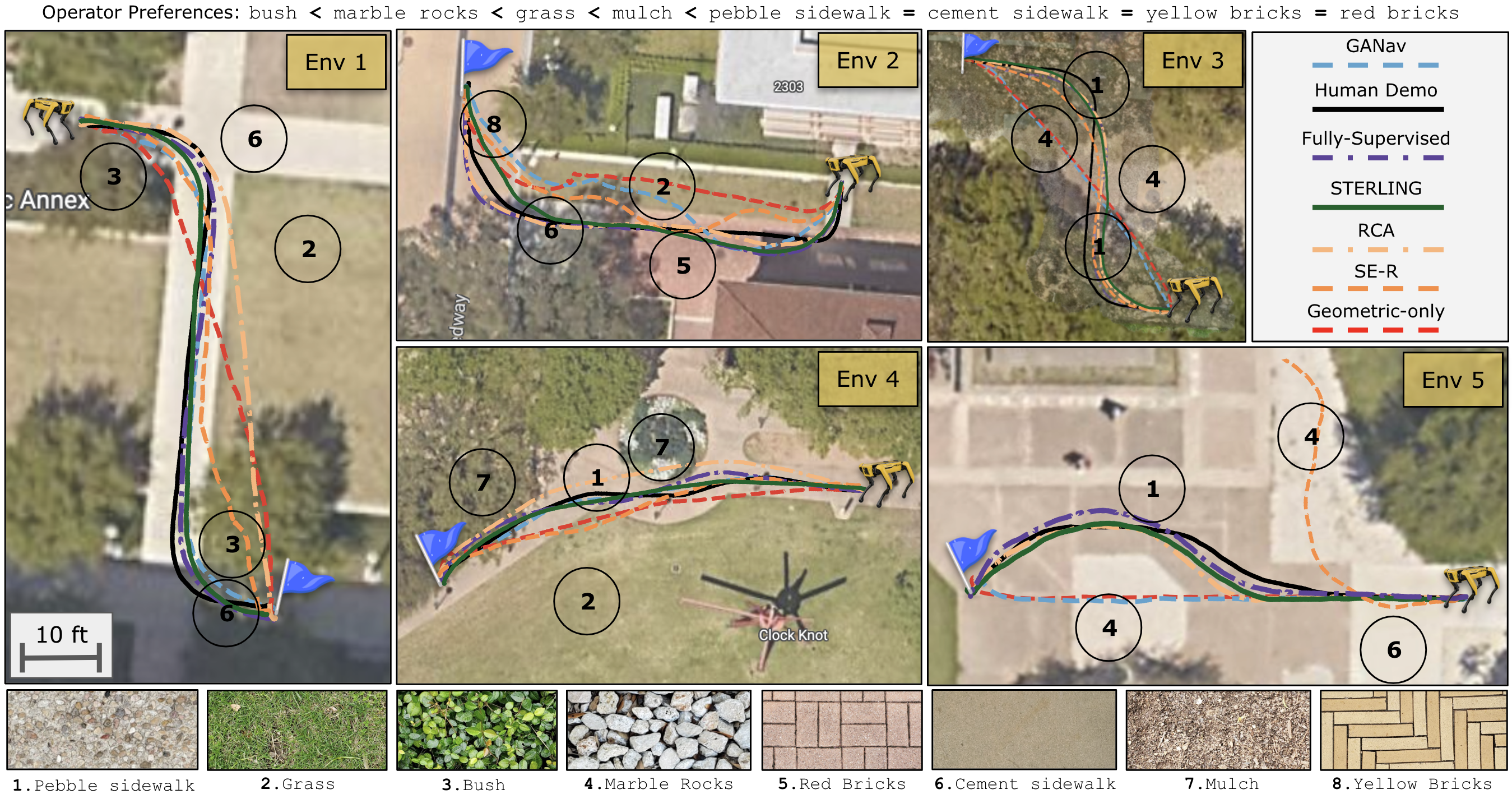}
        \caption{Trajectories traced by different approaches in 5 environments containing 8 different terrains. The operator preferences are shown above. We see that \sterling{} navigates in an operator-preference aligned manner, by preferring \texttt{cement sidewalk, red bricks, pebble sidewalk, and yellow bricks} over \texttt{mulch, grass, marble rocks, and bush}, outperforming other baselines and performing on-par with the Fully-Supervised approach.}
        \label{fig:robot_expts_sterling}
\end{figure*}

\begin{table}[h]
\centering
\caption{Hyperparameter Choices for \sterling{} Experiments}
\label{tab:hyperparameters}
\begin{tabular}{|c|c|}
\hline
\textbf{Hyperparameter} & \textbf{Value/Range} \\
\hline
Learning Rate & $3 \times 10^{-4}$ \\
\hline
Batch Size & 128 \\
\hline
Number of Epochs & 50 \\
\hline
Optimizer & Adam \\
\hline
Weight Decay & $5 \times 10^{-5}$ \\
\hline
Activation Function & ReLU \\
\hline
Kernel Size & 3 \\
\hline
Stride & 1 \\
\hline
\end{tabular}
\end{table}

\subsubsection{Evaluating Terrain-Awareness via Robot Experiments}
\label{sec:alignment_with_operator_preferences}

In this subsection, we report on experiments to investigate the effectiveness of \sterling{} features in enabling terrain awareness during off-road navigation. We quantitatively compare the performance of \sterling{} with baselines \rca{} \citep{yao2022rca}, \ganav{} \citep{ganav}, \ser{} \citep{ser} and the fully-supervised baseline, on the task of preference-aligned navigation. We identify six environments within the university campus, with eight different terrain types, as shown in Figure \ref{fig:robot_expts_sterling}. For this study, we use the same data collected on the robot to train \rca{}, \ser{}, fully-supervised baseline, and \sterling{}, and the operator provides the same rankings for all methods during training. Note that we use the same encoder and utility function across all environments and do not retrain/finetune to each environment to prevent environment-specific overfitting. 

Figure \ref{fig:robot_expts_sterling} shows the operator's terrain preferences for all Environments 1 to 5, and the performance of baseline approaches, including an operator-demonstrated trajectory for reference. In all environments, we see that \sterling{} navigates in a terrain-aware manner while adhering to the operator's preferences. Note that although Fully-Supervised also completes the task successfully, it requires privileged information such as terrain labels during training, whereas \sterling{} does not require such supervision, and can potentially be used on large datasets containing unlabeled, unconstrained robot experiences. \ganav{}, trained on the \textsc{rugd} dataset fails to generalize to unseen real-world conditions. \rca{} uses inertial spectral features to learn terrain traversability costs and hence does not adhere to operator preference. \ser{} does not address viewpoint invariance which is a significant problem in vision-based off-road navigation and hence performs poorly in Environments 1 and 2. We perform additional experiments in an outdoor environment (Environment 6) to study adherence to operator preferences, detailed in Section \ref{sec:adherance-test}.

Table \ref{table:success_rate} shows the success rate of preference alignment for all approaches in all environments, over five different trials. \sterling{} outperforms other self-supervised baselines and performs on par with the fully-supervised approach. In summary, the physical experiments conducted in six environments quantitatively demonstrate the effectiveness of \sterling{} features in enabling terrain awareness during off-road navigation.

\begin{table}[h]
\centering
\caption{Success rates of different algorithms on the task of preference-aligned off-road navigation. While both \sterling{} and Fully-Supervised succeed in all five trials in all environments, \sterling{} achieves visual terrain awareness without utilizing extensive human feedback in the form of terrain label annotations.}
\begin{tabular}{|l|ccccccc|}
\hline
\multicolumn{1}{|c|}{\multirow{2}{*}{\textbf{Approach}}} & \multicolumn{7}{c|}{\textbf{Environment}} \\ \cline{2-8}
\multicolumn{1}{|c|}{} & \multicolumn{1}{c|}{\textbf{1}}  & \multicolumn{1}{c|}{\textbf{2}}  & \multicolumn{1}{c|}{\textbf{3}}   & \multicolumn{1}{c|}{\textbf{4}} & \multicolumn{1}{c|}{\textbf{5}} & \multicolumn{1}{c|}{\textbf{6 (a)}} & \textbf{6 (b)} \\ \hline
Geometric-only & \multicolumn{1}{c|}{0/5} & \multicolumn{1}{c|}{0/5} & \multicolumn{1}{c|}{0/5} & \multicolumn{1}{c|}{0/5} & \multicolumn{1}{c|}{0/5} & \multicolumn{1}{c|}{0/5} & 5/5 \\ \hline
\rca{}\citep{yao2022rca} & \multicolumn{1}{c|}{2/5} & \multicolumn{1}{c|}{4/5} & \multicolumn{1}{c|}{2/5} & \multicolumn{1}{c|}{0/5} & \multicolumn{1}{c|}{1/5} & \multicolumn{1}{c|}{5/5} & 0/5 \\ \hline
\ganav{}\citep{ganav} & \multicolumn{1}{c|}{5/5} & \multicolumn{1}{c|}{0/5} & \multicolumn{1}{c|}{0/5} & \multicolumn{1}{c|}{5/5} & \multicolumn{1}{c|}{0/5} & \multicolumn{1}{c|}{4/5} & 5/5 \\ \hline
\ser{}\citep{ser} & \multicolumn{1}{c|}{1/5} & \multicolumn{1}{c|}{0/5} & \multicolumn{1}{c|}{5/5} & \multicolumn{1}{c|}{1/5} & \multicolumn{1}{c|}{3/5} & \multicolumn{1}{c|}{5/5} & 4/5 \\ \hline
Fully-Supervised & \multicolumn{1}{c|}{5/5} & \multicolumn{1}{c|}{5/5} & \multicolumn{1}{c|}{5/5} & \multicolumn{1}{c|}{5/5} & \multicolumn{1}{c|}{5/5} & \multicolumn{1}{c|}{5/5} & 5/5 \\ \hline \hline
\sterling{} (Ours) & \multicolumn{1}{c|}{5/5} & \multicolumn{1}{c|}{5/5} & \multicolumn{1}{c|}{5/5} & \multicolumn{1}{c|}{5/5} & \multicolumn{1}{c|}{5/5} & \multicolumn{1}{c|}{5/5} & 5/5 \\ \hline
\end{tabular}
\label{table:success_rate}
\end{table}

\begin{figure*}
    \centering
    \includegraphics[width=\columnwidth]{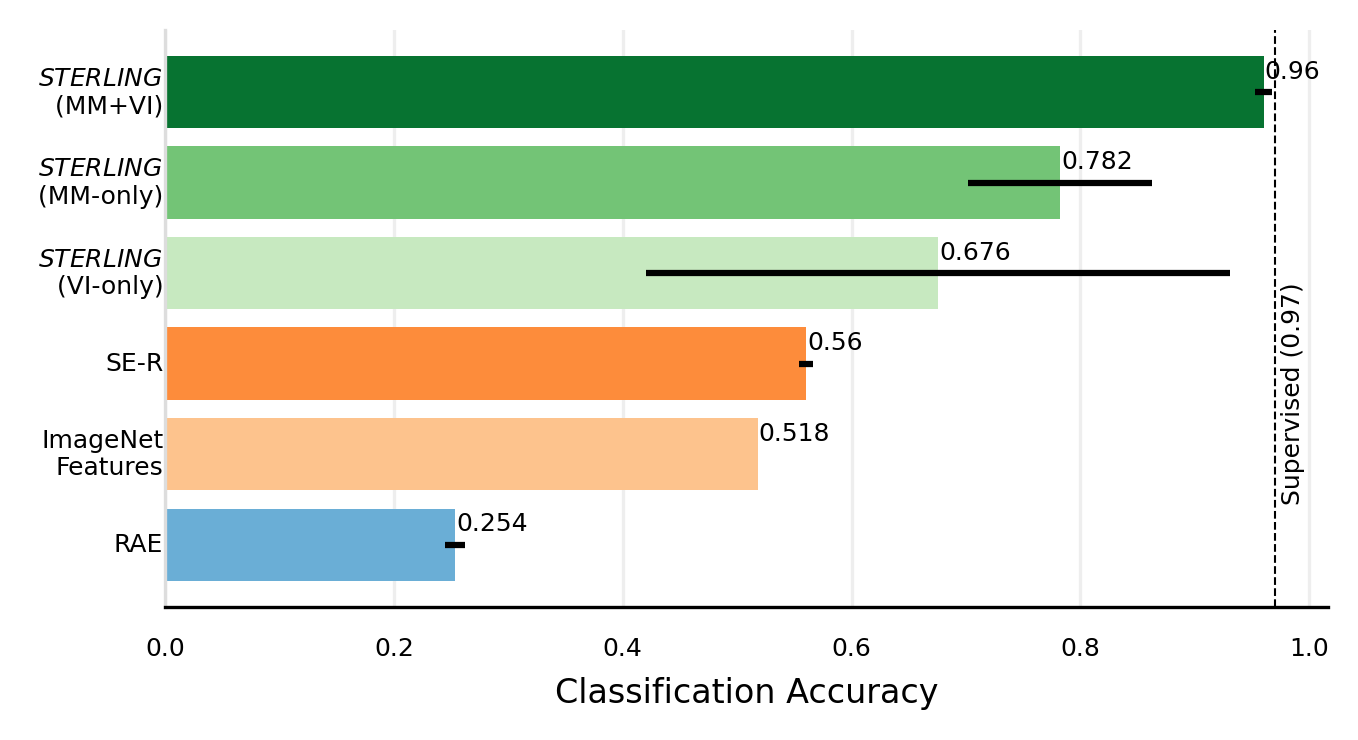}
        \caption{Ablation study depicting classification accuracy (value closer to $1.0$ is better) from terrain representations learned using different approaches and objectives. The combined objective (\vi{} + \mm{}) proposed in \sterling{} achieves the highest accuracy, indicating that the learned representations are sufficiently discriminative of terrains.}
        \label{fig:ablation}
\end{figure*}

\subsubsection{Evaluating Self-Supervision Objectives}
\label{sec:selfsupervisionablation}
In this subsection, we investigate the effectiveness of \sterling{} at learning discriminative terrain features and compare with baseline unsupervised terrain representation learning methods such as Regularized Auto-Encoder (\textsc{rae}) and \ser{} \citep{ser} and large pretrained networks such as a ResNet-50 pretrained on ImageNet. \sterling{} uses multi-modal correlation ($\mathcal{L}_{MM}$) and viewpoint invariance ($\mathcal{L}_{VI}$) objectives for self-supervised representation learning, whereas, \ser{} and  \textsc{rae} use soft-triplet-contrastive loss and pixel-wise reconstruction loss, respectively. Additionally, we also perform an ablation study on the two objectives in \sterling{} to understand their contributions to learning discriminative terrain features. To evaluate different visual representations, we perform unsupervised classification using k-means clustering and compare their relative classification accuracies with manually labeled terrain labels. For this experiment, we train \sterling{}, \ser{}, and \textsc{rae} on our training set and evaluate on a held-out validation set. Figure \ref{fig:ablation} shows the results of this study. We see that \sterling{}-features using both the self-supervision objectives perform the best among all methods. Additionally, we see that using a non-contrastive representation learning approach such as \vicreg{} \citep{bardes2021vicreg} within \sterling{} performs better than contrastive learning methods such as \ser{}, and reconstruction-based methods such as \textsc{rae}. This study shows that the proposed self-supervision objectives in \sterling{} indeed help learn discriminative terrain features.

\subsubsection{Preference Alignment Evaluation}
\label{sec:adherance-test}
In addition to the evaluations of \sterling{}-features with baseline approaches in five environments as shown in Section \ref{sec:experimental_results_sterling}, we utilize Environment 6 to further study adherence to operator preferences. We hypothesize that the discriminative features learned using \sterling{} is sufficient to learn the preference cost for local planning. To test this hypothesis, in Environment 6 containing three terrains as shown in Figure \ref{fig:preference_alignment}, the operator provides two different preferences 6(a) and 6(b). While $\texttt{bush}$ is the least preferred in both cases, in 6(a), $\texttt{sidewalk}$ is more preferred than $\texttt{grass}$ and in 6(b), both $\texttt{grass}$ and $\texttt{sidewalk}$ are equally preferred. We see in Figure \ref{fig:preference_alignment} that using \sterling{} features, the planner is able to sufficiently distinguish the terrains and reach the goal while adhering to operator preferences. Although \ser{} \citep{ser} adheres to operator preference in 6(b), it incorrectly maps \texttt{grass} to \texttt{bush}, assigning a higher cost and taking a longer route to reach the goal. On the other hand, \rca{} \citep{yao2022rca} fails to adhere to operator preferences since it directly assigns traversability costs using inertial features.

\begin{figure*}
        \centering
        \includegraphics[width=\textwidth]{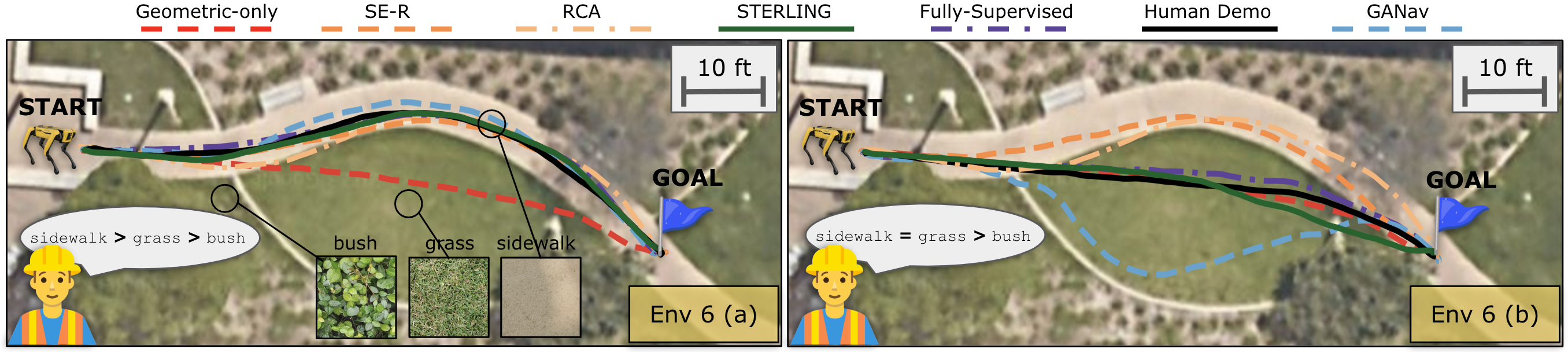}
        \caption{Trajectories traced by different approaches for the task of preference-aligned off-road navigation. Shown here are two different preferences expressed by the operator in the same environment---in 6 (a), \texttt{sidewalk} is more preferred than \texttt{grass} which is more preferred than \texttt{bush}, and in 6 (b), \texttt{grass} and \texttt{sidewalk} are equally preferred and \texttt{bush} is least preferred. We see that without retraining the terrain features, in both cases (a) and (b), \sterling{} optimally navigates to the goal while adhering to operator preferences.}
        \label{fig:preference_alignment}
\end{figure*}

\subsubsection{Large-Scale Qualitative Evaluation}
\label{sec:large_scale}
In this subsection, we perform a qualitative evaluation of \sterling{} by reporting a large-scale study of semi-autonomously hiking a 3-mile-long off-road trail using the Spot robot.

\begin{figure*}
        \centering        \includegraphics[width=\textwidth]{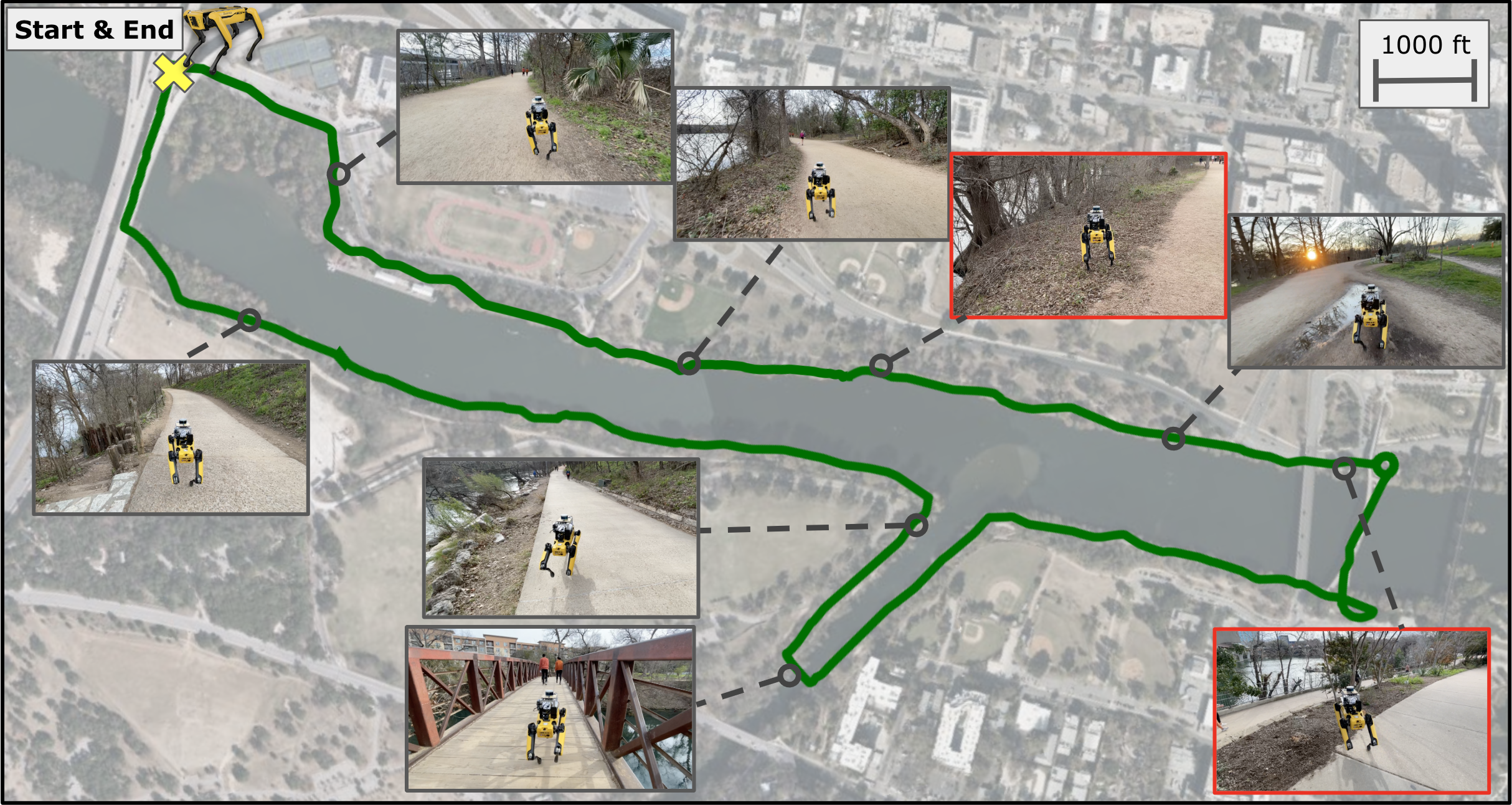}
        \caption{A large-scale qualitative evaluation of \sterling{} on a 3-mile outdoor trail. \sterling{} features successfully complete the trail with only two manual interventions (shown in red).}
        \label{fig:large_scale}
\end{figure*}

We train \sterling{} using unconstrained robot experience collected within the university campus and train the preference utility function using operator-provided preferences: \texttt{marble rocks < grass < dirt = cement}. The task is to navigate the trail without a global map while adhering to operator preferences at all times. Since we do not use a global map, visual terrain awareness is necessary to navigate within the trail and avoid catastrophic events such as falling into the river next to the trail. We set a moving goal of six meters in front of the robot, updated every second. While the robot navigates autonomously, the operator walks behind the robot and takes manual control only to correct the robot's path during forks, or to yield to incoming pedestrians and pets.\footnote{ 
A video of the robot navigating the trail successfully while avoiding less preferred terrains can be accessed here: \href{https://youtu.be/dQb1XzocdtE}{https://youtu.be/dQb1XzocdtE}. The robot needed two manual interventions while traversing along the trail.}
Figure \ref{fig:large_scale} shows the 3-mile trajectory traced by the robot and the two failure cases that required manual intervention. This large-scale qualitative experiment demonstrates the reliability of \sterling{} during real-world off-road deployments.

\subsubsection{Experiments on a Wheeled Mobile Robot}

\sterling{} is intended as a general algorithm to learn relevant terrain representations for off-road navigation. Towards demonstrating the versatility of \sterling{} to being applied to robots of different morphology, we conduct two additional experiments on the Clearpath Jackal, a wheeled mobile robot. 

\noindent \textbf{Learning Representations on Wheeled Robots: } We utilize unconstrained data collected on the Jackal consisting of multi-modal visual and inertial sensor data and learn terrain representations using \sterling{} followed by a utility function of operator preferences. Figure \ref{fig:sterling_jackal} (\sterling{}-Jackal) shows the path traversed by the Jackal in Environment 6, following the human preference $\texttt{sidewalk} > \texttt{grass} > \texttt{bush}$. This experiment demonstrates the applicability of \sterling{} on wheeled robots with inertial sensors, as against legged robots that have access to additional sensors such as joint encoders and tactile information. 

\noindent \textbf{Zero-Shot Cross-Morphology Transfer: } In a noteworthy experiment to evaluate the transferable property of terrain representations across robot morphologies, we utilized the visual encoder trained on data from the legged Spot robot and applied it on the wheeled Jackal robot without additional fine-tuning. Figure \ref{fig:sterling_jackal} (\sterling{}-Spot) showcases the Jackal's trajectory, leveraging \sterling{} representations learned from Spot's data, and adhering to the operator's terrain preference: $\texttt{sidewalk} > \texttt{grass} > \texttt{bush}$. Figure \ref{fig:jackal_costmaps} shows costmaps generated using \sterling{} features, used by the sampling-based planner to navigate in an operator-aligned manner. This demonstrates \sterling{}'s capability to generalize across diverse robotic platforms, emphasizing its adaptability and broad applicability.

Figure \ref{fig:jackal_expts} shows a third-person view of the deployment of \sterling{}-Jackal in Environment 6. In both experiments above, we see the Jackal robot reaches the goal successfully while adhering to human operator preferences, in a terrain-aware manner, highlighting \sterling{}'s adaptability regardless of robot morphology.

\begin{figure*}
        \centering
        \includegraphics[width=\columnwidth]{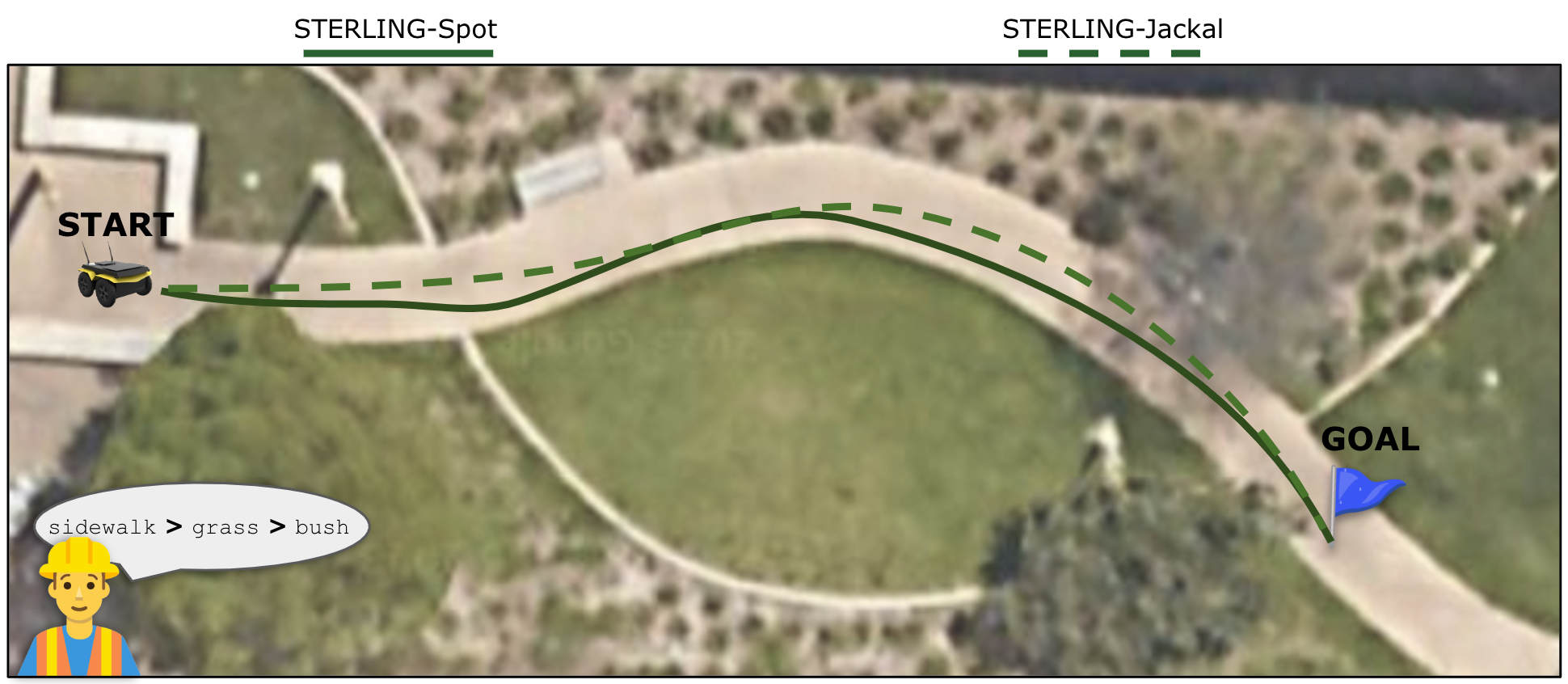}
        \caption{Experimental study of \sterling{} on a  Clearpath Jackal---a wheeled mobile robot in Environment 6. \sterling{}-Spot shows the trajectory traced using \sterling{} trained on data collected on the Spot, deployed zero-shot on the Jackal robot, whereas \sterling{}-Jackal shows the trajectory traced by \sterling{} trained on data collected on the Jackal, deployed also on the Jackal robot. In both experiments, we see the robot reach the goal successfully while adhering to human operator's preferences over terrains ($\texttt{sidewalk} > \texttt{grass} > \texttt{bush}$).}
        \label{fig:sterling_jackal}
\end{figure*}

\begin{figure*}
        \centering
        \includegraphics[width=\columnwidth]{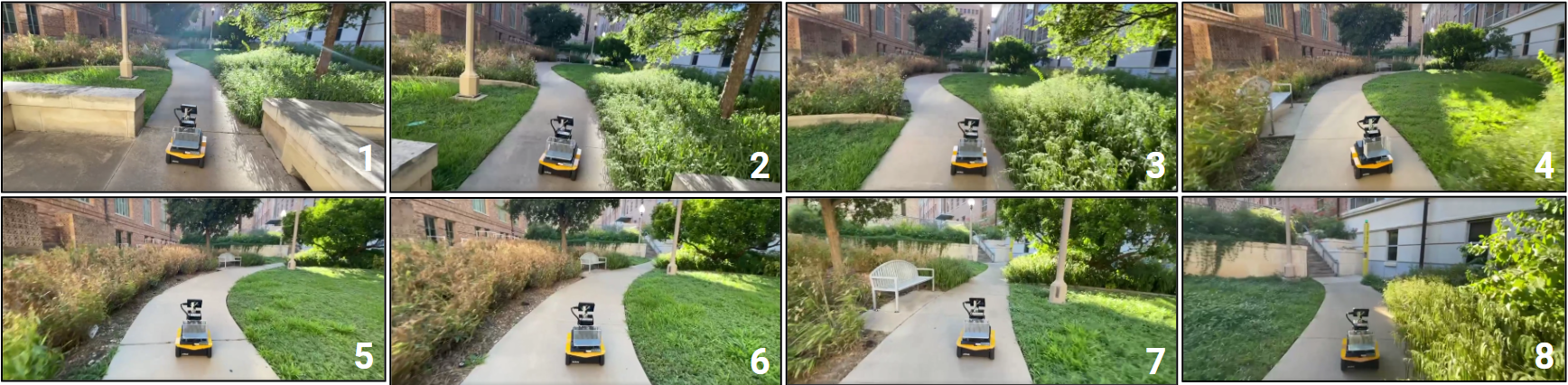}
        \caption{Clearpath Jackal, a wheeled robot navigating using \sterling{} features trained on unconstrained data collected on the Jackal robot (\sterling{}-Jackal). We see here in Environment 6 that the robot reaches the goal while adhering to operator preferences $\texttt{Sidewalk} > \texttt{Grass} > \texttt{Bush}$. This experiment demonstrates the versatility of \sterling{} in being applied to robots of different morphology.}
        \label{fig:jackal_expts}
\end{figure*}
\begin{figure*}
        \centering
    \includegraphics[width=\columnwidth]{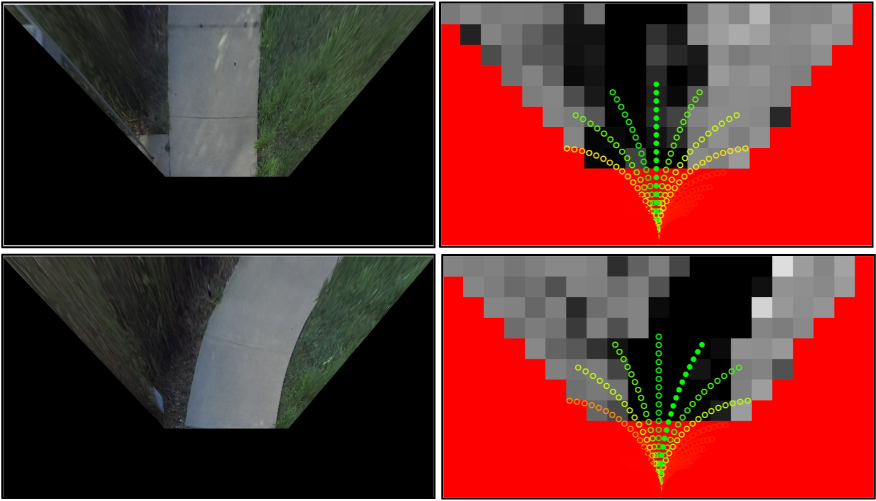}
        \caption{Visualizing the costmaps from the Jackal robot when traversing Env 6., trained using data from the Jackal (\sterling{}-Jackal).}
        \label{fig:jackal_costmaps}
\end{figure*}

\subsubsection{Visualizing the learned terrain representations}
Figure \ref{fig:tsne_sterling} depicts a t-SNE visualization of terrain representations learned using \sterling{}. Individual patches are color-coded by their ground truth semantic terrain label. We see that \sterling{} learns relevant features for terrains, given their unique clustering in this latent space. 

\begin{figure*}
    \centering
        \includegraphics[width=\columnwidth]{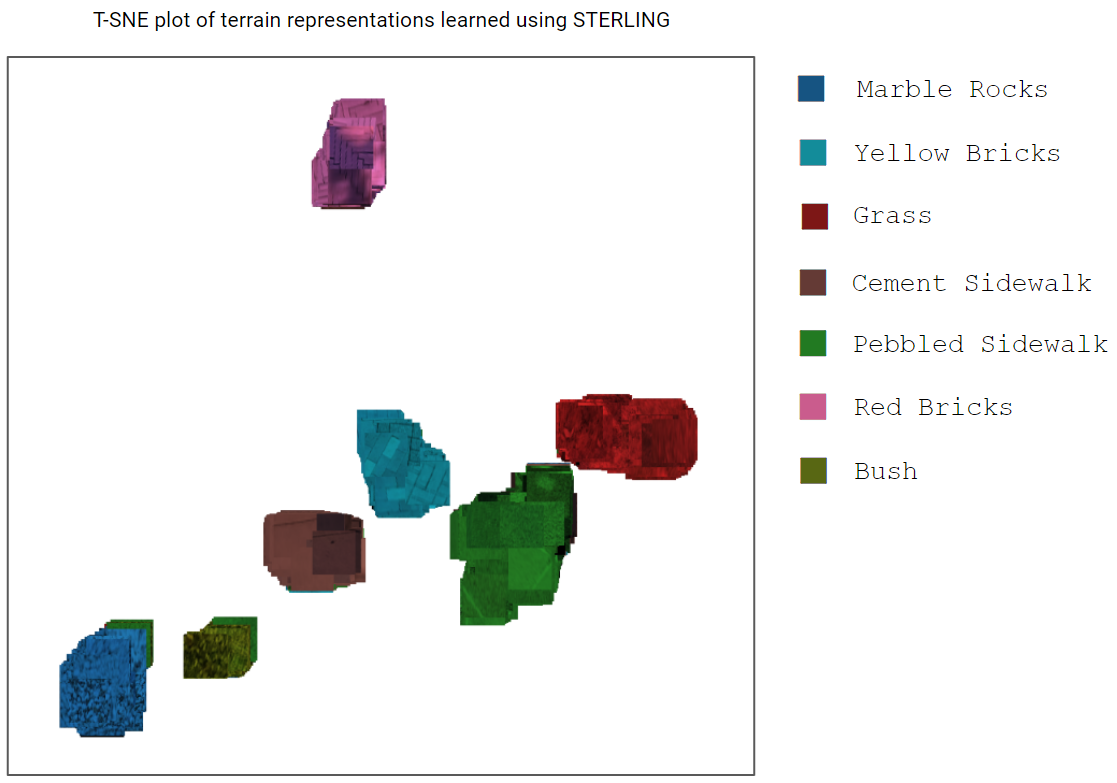}
        \caption{t-SNE visualization of terrain representations learned using STERLING. Each data point represents a terrain example, color-coded by its ground truth label. The clustering of colors showcases the efficacy of \sterling{} in capturing meaningful and distinctive terrain features.}
        \label{fig:tsne_sterling}
\end{figure*}

\subsubsection{Visualizing the costmaps}
Figure \ref{fig:costmaps} shows cost visualizations of baseline approaches - \rca{} \citep{yao2022rca}, \ser{} \citep{ser}, \ganav{} \citep{ganav} and Fully-Supervised in comparison with \sterling{}. Figure \ref{fig:costmaps} shows that \rca{} and \ser{} exhibit issues with visual artifacts due to homography transformations. \ganav{}, trained on the \textsc{rugd} \cite{rugd} dataset, fails to generalize to novel real-world situations. In contrast, costmaps from both the fully-supervised model and \sterling{} efficiently guide planning, as demonstrated by quantitative results in Section \ref{sec:experimental_results_sterling} and results in behaviors that align with operator preferences, prioritizing sidewalks over terrains like rocks or bushes.

\begin{figure*}
    \centering
        \includegraphics[width=\columnwidth]{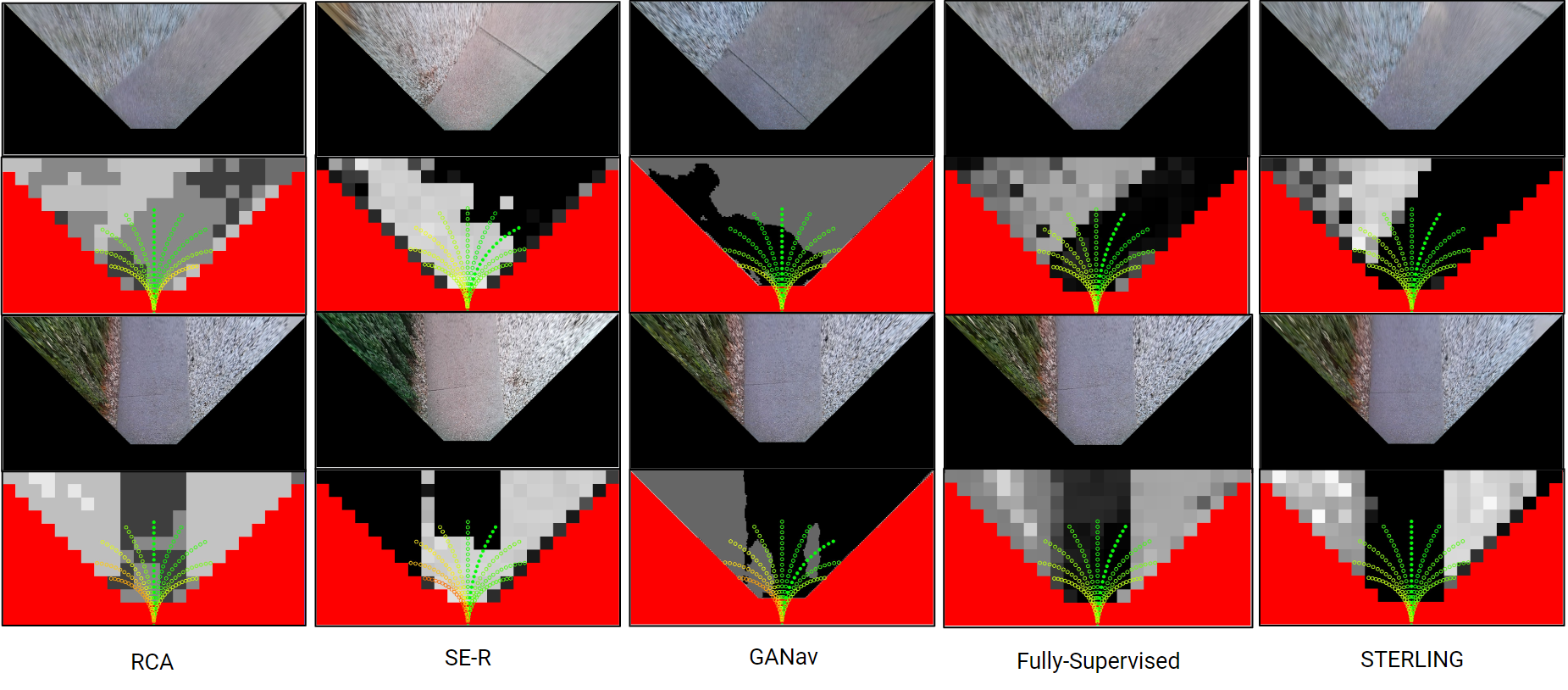}
        \caption{Comparative visualization of the costmaps generated by \sterling{} (this work) and other baseline algorithms (\rca{} \citep{yao2022rca}, \ser{} \citep{ser}, \ganav{} \citep{ganav}, Fully-Supervised) for a given scene. Paired with each costmap is a bird's-eye view image of the corresponding terrain. In the costmaps, white regions indicate high traversal cost, black signifies low cost, and areas in red are ignored or non-observable regions. We see that compared to other approaches, using \sterling{} features results in costmaps that align with operator preference of $\texttt{Sidewalk} > \texttt{Rocks} > \texttt{Bush}$.}
        \label{fig:costmaps}
\end{figure*}

\subsubsection{Generalization to Unseen Terrains}
During autonomous off-road navigation, generalization to novel terrains is paramount. Although difficult to comment on generalizability, we document an instance during the large-scale deployment where \sterling{} navigates around an unseen terrain ``\texttt{water puddle}", as shown in Figure \ref{fig:water_puddle}.

\begin{figure*}
    \centering
        \includegraphics[width=\columnwidth]{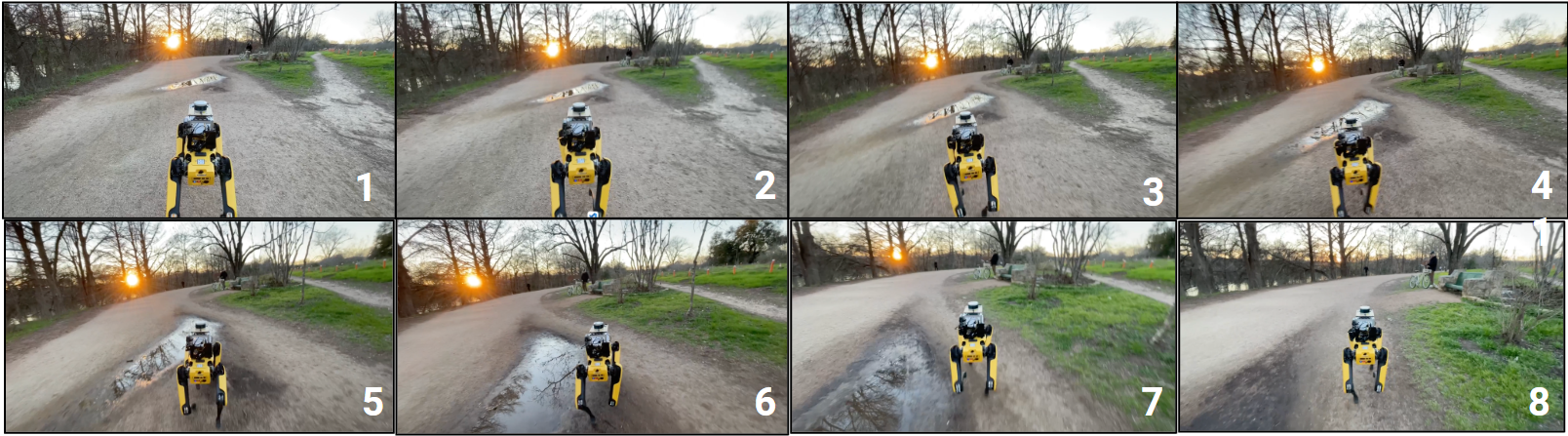}
        \caption{\sterling{} navigating around an unfamiliar terrain, specifically a ``\texttt{water puddle}", during the qualitative 3-mile off-road deployment.}
        \label{fig:water_puddle}
\end{figure*}

\subsection{On the Efficacy of Multi-Modal Data Over Vision Alone}
 While it might seem that visual cues are sufficient for distinguishing terrains, as evidenced in Figure~\ref{fig:robot_expts_sterling}, the reality is more complex. Variations in lighting, shadow, color, texture, and other artifacts may lead to inconsistent representations for the same terrain type and can render visually distinctive terrains deceptively similar. For instance, while six different visual patches of the terrain ``sidewalk" as shown in Figure~\ref{fig:sidewalk_patches} might each exhibit unique visual characteristics because of these variations, they all denote the same terrain category and evoke similar inertial-prioprio-tactile (\textsc{ipt}) response on the robot (feel similar to the robot). Solely relying on vision may lead to overlooking underlying commonalities between terrains, resulting in inconsistent terrain representations. 

Another concrete example is the scenario of fallen leaves. A sidewalk, a grass patch, and a forest trail could all be covered with fallen leaves, making them visually similar. However, underneath those leaves, the actual terrain properties – and the robot's interaction with them – vary significantly. While the leaves might visually mask the terrain differences, the robot would feel different terrain responses when moving over them due to differences in underlying ground properties.

Furthermore, visual similarity is not a conclusive indicator of identical terrains. Consider four images as a case in point, as shown in Figure \ref{fig:terrain_patches}. Though the first two and the last two images might seem visually similar, they represent distinct terrains: the first image depicts ``bush", the second and third denote ``grass", and the fourth is ``sidewalk". These three terrains induce different inertial, proprioceptive, and tactile responses in a robot. Thus, the mere semblance of appearance does not capture the relevant features of a terrain.

\sterling{}'s approach of integrating additional modalities allows for more precise terrain identification by accounting for these subtleties. By considering variations and similarities among terrains across different modalities, we ensure relevant terrain representations for off-road navigation. In all examples shown in Figs. \ref{fig:sidewalk_patches} and \ref{fig:terrain_patches}, \sterling{} correctly associates the samples with the right cluster for each terrain. 

\begin{figure*}
        \centering
        \includegraphics[width=\columnwidth]{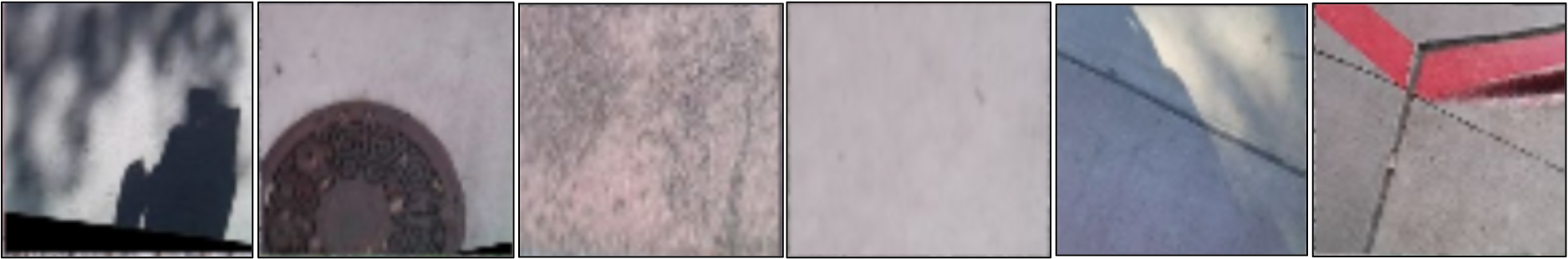}
        \caption{Six distinct instances of \texttt{sidewalk} terrain, showcasing the variability in visual appearance due to factors such as lighting, texture, shadows, and other artifacts. Despite the visual differences, each patch represents the same terrain type.}
        \label{fig:sidewalk_patches}
\end{figure*}

\begin{figure*}
    \centering
        \includegraphics[width=\columnwidth]{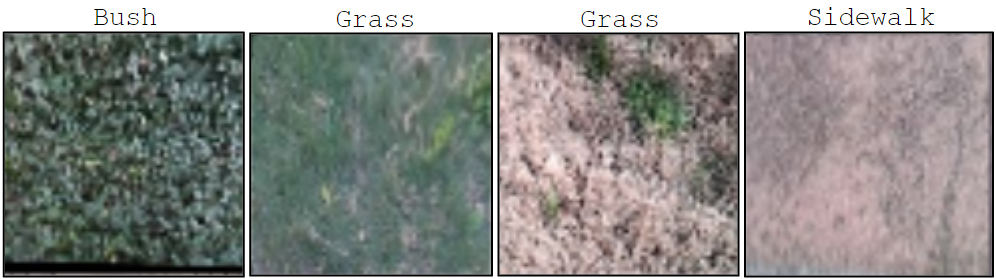}
        \caption{A collection of four terrain patches, illustrating the challenge of terrain representation learning based solely on visual cues. From left to right: bush, grass, grass, sidewalk. Despite visual similarities (and differences), the terrains can elicit different non-visual \textsc{ipt} responses on a robot.}
        \label{fig:terrain_patches}
\end{figure*}

\subsection{Limitations}

In \sterling{}, the data processing step assumes a flat-terrain geometry for all environments, which may not always be true in all conditions, such as stairs \citep{quad_loco} and boulders \citep{xuesurocks}. Incorporating terrain geometry by utilizing depth observations to learn relevant terrain representations is a promising direction for future work. \sterling{} also requires that the robot's hardware is equipped with sensors such as inertial, proprioceptive, or tactile sensors to observe the terrains, in order to learn consistent terrain representations. 
Extending \sterling{} by pretraining with large-scale off-road datasets using modern architectures such as transformers that are known to scale well with large-scale data \citep{kaplan2020scaling} is an exciting direction for future work.
\sterling{} requires a mobile robot to physically traverse over diverse terrains in order to learn representations for preference-aligned navigation, which may be quite unsafe in certain situations. To address this concern, uncertainty-aware safe exploration and curiosity-based exploration focusing on informative and diverse terrains for data collection is a promising direction for future work. 


In the \sterling{} framework, there is an inherent requirement for a human operator to specify their preferences for all observed terrains. However, \sterling{} does not account for situations where the robot encounters a visually novel terrain, leaving the operator's preference for such terrains undetermined. Overcoming this limitation, in the upcoming section, I introduce a novel algorithm \patern{} \citep{patern}, which complements \sterling{} in handling certain visually novel terrains. \patern{} capitalizes on the understanding that, under specific conditions, non-visual observations of terrain can effectively be used to extrapolate operator preferences from known to visually novel terrains. This method thus extends the capabilities of terrain representation learning algorithms like \sterling{}, enabling the extrapolation of known preferences to novel terrains.

\section{Extrapolating Operator Preferences to Visually Novel Terrains}
\label{sec:patern}
In this section, I introduce \patern{} \citep{patern}, a novel approach to extrapolate operator preferences in a self-supervised manner from known terrains to visually novel terrains. I first describe the data preprocessing, followed by the pre-adaptation training steps in Subsections \ref{sec:datapreprocessing_patern} and \ref{sec:pre_adaptation_training}, respectively. Section \ref{sec:pref_extrapolation_training} then introduces the preference extrapolation training scheme. Finally, in Section \ref{sec:expts_patern} I describe the experiments performed to evaluate \patern{} in comparison with relevant baseline algorithms. 

\subsection{Data Pre-Processing}
\label{sec:datapreprocessing_patern}
In tandem with the visual patch extraction process used in the projection operator $\Pi$ as in prior methods \citep{vrlpap, sterling}, for every state $s_t$, we also extract a 2-second history of time-series inertial (angular velocities along the x and y-axes and linear acceleration in the z-axis), proprioceptive (joint angles and velocities), and tactile (feet depth penetration estimates) data. To ensure the resulting input data representation for training is independent of the length and phase of the signals, we compute statistical measures of center and spread as well as the power spectral density, and maintain that as the input. All the visual patches extracted with the projection operator $\Pi$ and the non-visual data for each state $s$ are then tagged with their corresponding terrain name, given that each trajectory uniquely contains a particular terrain type. In addition to processing the recorded data in the pre-adaptation phase, a human operator is queried for a full-order ranking of terrain preference labels.

\subsection{Pre-Adaptation Training}
\label{sec:pre_adaptation_training}
We use a supervised contrastive learning formulation inspired by \cite{vrlpap} to train the baseline functions $f_{vis}$ and $u_{vis}$, represented as neural networks.

\noindent \textbf{Training the Encoders:} Given labeled visual patches and proprioception data, we generate triplets for contrastive learning such that for any anchor, the positive pair is chosen from the same label and the negative pair is sourced from another label. Given such triplets, we use triplet loss \citep{tripletloss} with a margin of $1.0$ to independently train the visual and proprioception encoders through mini-batch gradient descent using the AdamW optimizer. For the visual encoder, we use a 3-layer CNN of $5\times5$ kernels, each followed by ReLU activations. This model, containing approximately 250k parameters, transforms $64\times64$ size RGB image patches into an 8-dimensional vector representation $\phi_{vis}$. Similarly, our inertial encoder is comprised of a 3-layer MLP with ReLU activations, encompassing around 4k parameters, and maps proprioceptive inputs to an 8-dimensional vector $\phi_{pro}$. To mitigate the risk of overfitting, data is partitioned in a 75-25 split for training and validation, respectively.

\noindent \textbf{Training the Utility Functions}: In our setup, the utility function is represented as a two-layer MLP with ReLU non-linearity and output activation that maps an 8-dimensional vector into a singular non-negative real value. Given ranked operator preferences of the terrains, we follow Zucker et al. \cite{zucker2011optimization} and train the visual utility function $u_{vis}$ using a margin-based ranking loss \cite{BMVC2016_119}. Furthermore, to ensure consistent predictions from $u_{vis}$ and $u_{pro}$ for both visual and non-visual observations at identical locations, we update parameters of $u_{pro}$ using the loss $\mathcal{L}_{\text{MSE}}( u_{\text{pro}}) = \frac{1}{N} \sum_{i=1}^{N} \left( sg(u_{\text{vis}}(\phi_{vis})) - u_{\text{pro}}(\phi_{pro}) \right)^2$. Here, $sg(\cdot)$ denotes the stop-gradient operation, and $\phi_{vis}$ and $\phi_{pro}$ are the terrain representations from paired visual and non-visual data, respectively, at the same location.

The functions $f_{vis}$ and $u_{vis}$ prior to adaptation are collectively termed as $\patern{}^-$, signifying their non-adapted state with respect to visually novel terrains. In our implementation, although we use supervised contrastive learning, in instances where explicit terrain labels might be absent, one can resort to self-supervised representation learning techniques, such as \sterling{} \citep{sterling}, to derive $f_{vis}$ and $u_{vis}$. \patern{} can be applied regardless of the specific representation learning approach used.

\subsection{Preference Extrapolation Training}
\label{sec:pref_extrapolation_training}
During deployment, if the robot encounters a visually novel terrain, both visual and inertial-proprioceptive-tactile data is recorded to be used in the adaptation phase in \patern{}, aiding in preference extrapolation and subsequent model adaptation. We refer to this collected data as the \textit{adaptation-set}. We extract paired visual and non-visual observations at identical locations from the \textit{adaptation-set} and use $f_{pro}$ to extract proprioceptive representations $\phi_{pro}$. We cluster samples of $\phi_{pro}$ and perform a nearest-neighbor search against existing terrain clusters from the pre-adaptation dataset that is within a threshold $\mu$. We set this threshold to be the same as the triplet margin value of 1.0 which we find to work well in practice. This procedure seeks a known terrain that ``feels" similar to the novel terrain which then inherits the preference of its closest match.  Following this self-supervised preference extrapolation framework, the adaptation-set is aggregated with the pre-adaptation training set, and the visual encoder $f_{vis}$ is retrained using the procedure described in Section \ref{sec:pre_adaptation_training}. Additionally, the visual utility function $u_{vis}$ is retrained with the extrapolated preference for the novel terrain. The updated functions $f_{vis}$ and $u_{vis}$ are collectively referred to as $\patern{}^+$. Figure \ref{fig:training_setup} illustrates retraining and preference extrapolation as described above.

\begin{figure*}
    \centering
    \includegraphics[width=\columnwidth]{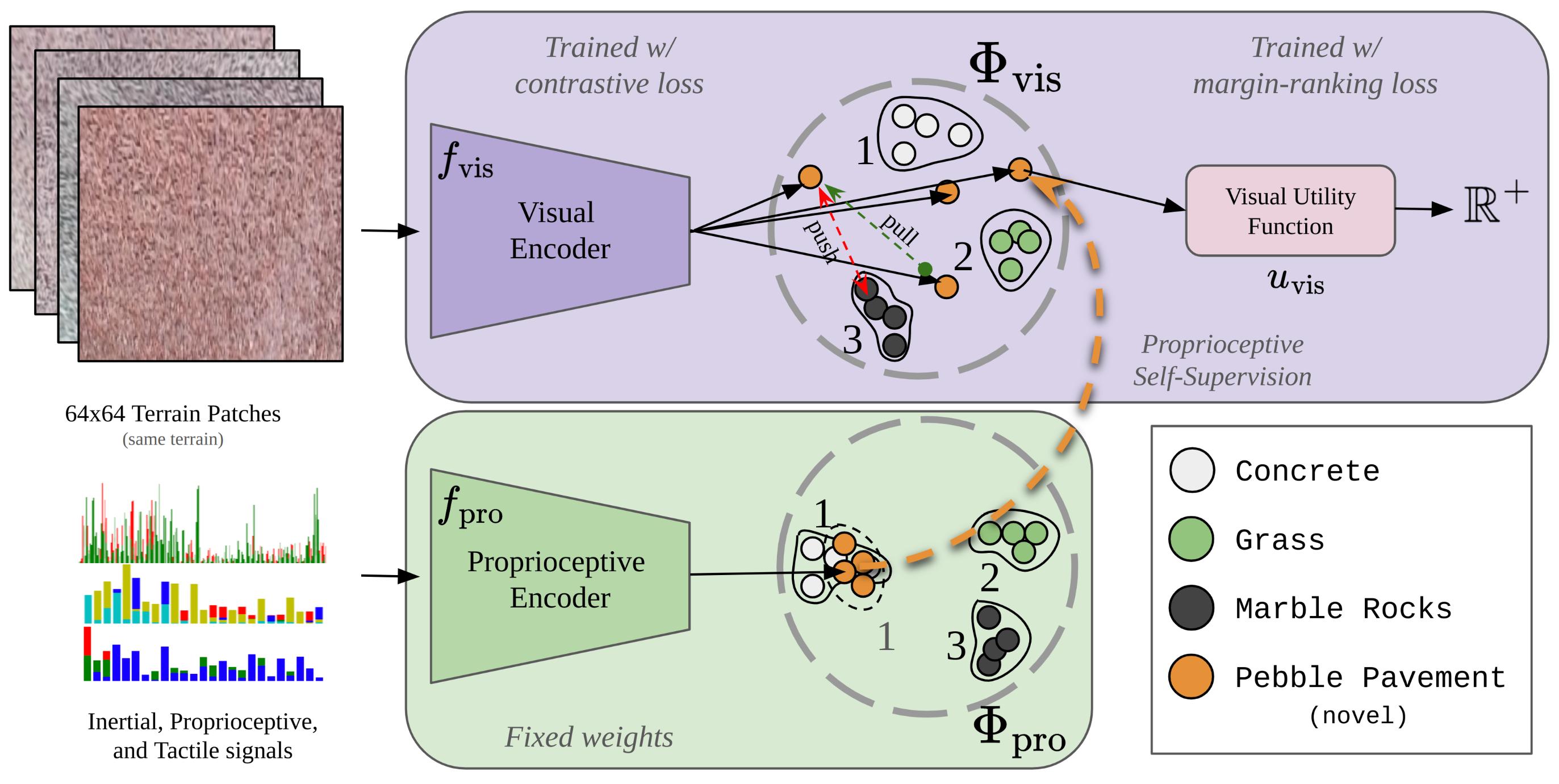}
    \caption{An illustration of the training setup for preference extrapolation proposed in \patern{}. We utilize two encoders to map visual and inertial-proprioceptive-tactile samples to $\Phi_{vis}$ and $\Phi_{pro}$ respectively. For a visually novel terrain, a preference is hypothesized and extrapolated from $\Phi_{pro}$, following which the visual encoder and utility function are retrained. 
    }
    \label{fig:training_setup}
\end{figure*}

\subsection{Experiments}
\label{sec:expts_patern}
In this section, I describe the physical robot experiments conducted to evaluate \patern{} against other state-of-the-art visual off-road navigation algorithms. Specifically, our experiments are designed to explore the following questions: 

\begin{enumerate}[label=($Q_\arabic*$)]
    \item Is \patern{} capable of extrapolating operator preferences accurately to novel terrains?
    \item How effectively does \patern{} navigate under challenging lighting scenarios such as nighttime conditions? 
    \item How well does \patern{} perform in large-scale real-world off-road conditions?
\end{enumerate}

\begin{figure*}[h!]
        \centering
        \includegraphics[width=\columnwidth]{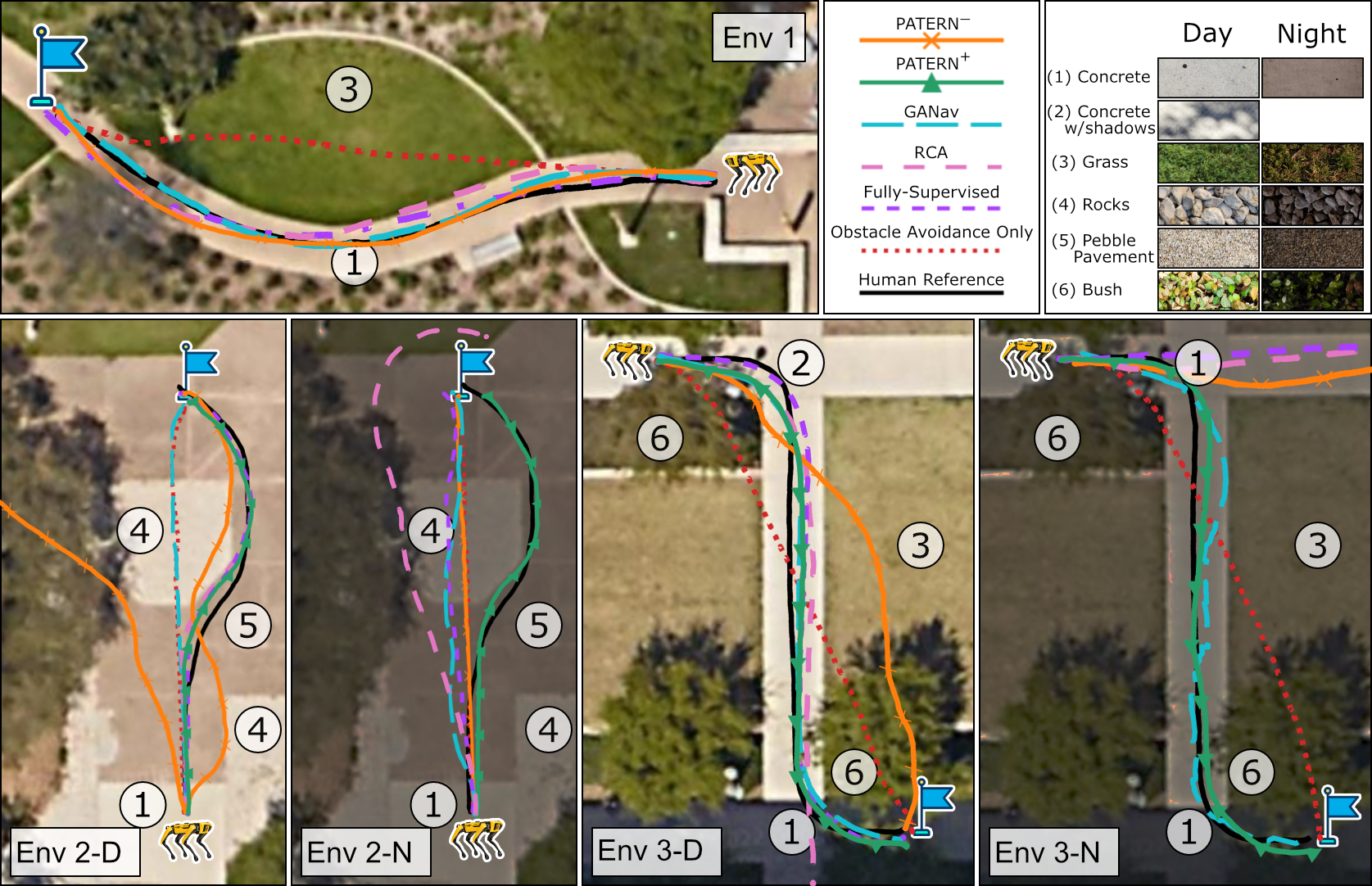}
        \caption{Trajectories traced by \patern{} and baseline approaches across three different environments and varied lighting conditions within the UT Austin campus. Note the drastic changes in the appearance of the terrain between day and night, which pose a significant challenge for visual navigation. In environments where $\patern{}^-$ fails to generalize, $\patern{}^+$ successfully extrapolates and reaches the goal in a preference-aligned manner.}
        \label{fig:robot_expts}
\end{figure*}

\noindent To study $Q_1$ and $Q_2$, we execute a series of experiments consisting of short-scale outdoor navigation tasks. We then conduct a large-scale autonomous robot deployment along a 3-mile outdoor trail to qualitatively evaluate $Q_3$.
We use the legged Boston Dynamics Spot robot for all experiments, equipped with a VectorNav VN100 IMU, front-facing Kinect RGB camera, Velodyne VLP-16 LiDAR for geometric obstacle detection, and a GeForce RTX 3060 mobile GPU. For local planning, we use an open-source sampling-based motion planner called \textsc{graphnav} \citep{graphnavgithub} and augment its sample evaluation function with the inferred preference cost $L_P(\Gamma)$. For real-time planning, we run $f_{vis}$ and $u_{vis}$ on the onboard GPU.

\begin{table}[ht]
\centering
\caption{Mean Hausdorff distance relative to a human reference trajectory.}
\label{table:hausdorff}
\begin{tabular}{|l|ccccc|}
\hline

    \multicolumn{1}{|c|}{\multirow{2}{*}{\textbf{Approach}}}
    & \multicolumn{5}{c|}{\textbf{Environment}}
    \\ \cline{2-6}

    \multicolumn{1}{|c|}{}
    & \multicolumn{1}{c|}{\textbf{1}}
    & \multicolumn{1}{c|}{\textbf{2-D}}
    & \multicolumn{1}{c|}{\textbf{2-N}}
    & \multicolumn{1}{c|}{\textbf{3-D}}
    & \textbf{3-N}
    \\ \hline

    Geometric-Only
    & \multicolumn{1}{c|}{2.87}
    & \multicolumn{1}{c|}{2.34}
    & \multicolumn{1}{c|}{2.34}
    & \multicolumn{1}{c|}{3.44}
    & 3.69
    \\ \hline

    \rca{}\citep{yao2022rca}
    & \multicolumn{1}{c|}{0.84}
    & \multicolumn{1}{c|}{0.91}
    & \multicolumn{1}{c|}{6.061}
    & \multicolumn{1}{c|}{2.57}
    & 7.37
    \\ \hline

    \ganav{}\citep{ganav}
    & \multicolumn{1}{c|}{1.47}
    & \multicolumn{1}{c|}{2.98}
    & \multicolumn{1}{c|}{3.07}
    & \multicolumn{1}{c|}{0.898}
    & 1.42
    \\ \hline

    Fully-Supervised
    & \multicolumn{1}{c|}{0.58}
    & \multicolumn{1}{c|}{\textbf{0.44}}
    & \multicolumn{1}{c|}{2.735}
    & \multicolumn{1}{c|}{\textbf{0.763}}
    & 6.747
    \\ \hline \hline

    $\patern{}^-$
    & \multicolumn{1}{c|}{\textbf{0.54}}
    & \multicolumn{1}{c|}{2.31}
    & \multicolumn{1}{c|}{2.29}
    & \multicolumn{1}{c|}{2.305}
    & 5.76
    \\ \hline

    $\patern{}^+$
    & \multicolumn{1}{c|}{-}
    & \multicolumn{1}{c|}{0.56}
    & \multicolumn{1}{c|}{\textbf{1.097}}
    & \multicolumn{1}{c|}{0.86}
    & \textbf{0.763}
    \\ \hline
\end{tabular}
\end{table}

\begin{table}[ht]
\centering
\caption{Mean trajectory percentage aligned with operator preferences.}
\label{table:aligned_percent}
\begin{tabular}{|l|ccccc|}
\hline
    \multicolumn{1}{|c|}{\multirow{2}{*}{\textbf{Approach}}}
    & \multicolumn{5}{c|}{\textbf{Environment}}
    \\ \cline{2-6}

    \multicolumn{1}{|c|}{}
    & \multicolumn{1}{c|}{\textbf{1}}
    & \multicolumn{1}{c|}{\textbf{2-D}}
    & \multicolumn{1}{c|}{\textbf{2-N}}
    & \multicolumn{1}{c|}{\textbf{3-D}}
    & \textbf{3-N}
    \\ \hline

    Geometric-Only
    & \multicolumn{1}{c|}{44.0\%}
    & \multicolumn{1}{c|}{68.8\%}
    & \multicolumn{1}{c|}{68.8\%}
    & \multicolumn{1}{c|}{43.6\%}
    & 43.6\%
    \\ \hline

    \rca{}\cite{rca}
    & \multicolumn{1}{c|}{100\%}
    & \multicolumn{1}{c|}{97.3\%}
    & \multicolumn{1}{c|}{67.4\%}
    & \multicolumn{1}{c|}{100\%}
    & 99.4\%
    \\ \hline

    \ganav{}\cite{ganav}
    & \multicolumn{1}{c|}{93.9\%}
    & \multicolumn{1}{c|}{71.6\%}
    & \multicolumn{1}{c|}{71.4\%}
    & \multicolumn{1}{c|}{100\%}
    & 100\%
    \\ \hline

    Fully-Supervised
    & \multicolumn{1}{c|}{100\%}
    & \multicolumn{1}{c|}{100\%}
    & \multicolumn{1}{c|}{71.7\%}
    & \multicolumn{1}{c|}{100\%}
    & 93.6\%
    \\ \hline \hline

    $\patern{}^-$
    & \multicolumn{1}{c|}{\textbf{100\%}}
    & \multicolumn{1}{c|}{94.1\%}
    & \multicolumn{1}{c|}{71.6\%}
    & \multicolumn{1}{c|}{81.3\%}
    & 100\%
    \\ \hline

    $\patern{}^+$
    & \multicolumn{1}{c|}{-}
    & \multicolumn{1}{c|}{\textbf{100\%}}
    & \multicolumn{1}{c|}{\textbf{98.2\%}}
    & \multicolumn{1}{c|}{\textbf{100\%}}
    & \textbf{100\%}

\\ \hline
\end{tabular}
\end{table}


\subsubsection{Data Collection}
\label{subsec:dataprocessing}
 To collect labeled data for training $\patern{}^-$, we manually teleoperated the robot across the UT Austin campus, gathering 8 distinct trajectories each across 3 terrains: \texttt{concrete}, \texttt{grass}, and \texttt{marble rocks}. Each trajectory, five minutes long, is exclusive to a single terrain type for ease of labeling and evaluation. The pre-adaptation training data was collected in the daytime under sunny conditions. Our evaluations then centered on two preference extrapolation scenarios: one, extending to new terrains such as \texttt{pebble}-\texttt{pavement}, \texttt{concrete-with-shadows}, and \texttt{bushes}, all experienced under varying daylight conditions ranging from bright sunlight to overcast skies, and two, adapting to nighttime illumination for familiar terrains that appear visually different. In our experiments, we use the preference order $\texttt{concrete} \succ \texttt{grass} \succ \texttt{marble rocks}$.

\subsubsection{Quantitative Short-Scale Experiments}

We evaluate \patern{} in three environments with a variety of terrains within the UT Austin campus. We also test under two different lighting conditions, as shown in Figure \ref{fig:robot_expts}. The primary task for evaluation is preference-aligned visual off-road navigation, in which the robot is tasked with reaching a goal, while adhering to operator preferences over terrains.

To evaluate the effectiveness of \patern{}, we compare it against five state-of-the-art baseline and reference approaches: \textbf{Geometric-Only} \citep{graphnavgithub}, a purely geometric obstacle-avoidant planner; \textbf{RCA} \citep{yao2022rca}, a self-supervised traversability estimation algorithm based on ride-comfort; \textbf{GANav}, a semantic segmentation method \footnote{\href{https://github.com/rayguan97/GANav-offroad}{https://github.com/rayguan97/GANav-offroad}} trained on RUGD dataset~\citep{rugd}; \textbf{Fully-Supervised}, an approach that utilizes a visual terrain cost function comprehensively learned using supervised costs drawn from operator preferences; and lastly, \textbf{Human Reference}, which offers a preference-aligned reference trajectory where the robot is teleoperated by a human expert. To train the \rca{} and Fully-Supervised baselines, in addition to the entirety of the pre-adaptation dataset on the 3 known terrains, we additionally collect 8 trajectories each on the novel terrains during daytime. 

In each environment, we perform five trials of each method to ensure consistency in our evaluation. For each trial, the robot is relocalized in the environment, and the same goal location $G$ is fed to the robot. In the environments where $\patern{}^-$ fails to navigate in a preference-aligned manner, we run five trials of the self-supervised $\patern{}^+$ instance that uses experience gathered in these environments to extrapolate preferences to the novel terrains or novel lighting conditions. Figure \ref{fig:robot_expts} shows the qualitative results of trajectories traced by each method in the outdoor experiments. Only one trajectory is shown for each method unless there is significant variation between trials.
Table \ref{table:hausdorff} shows quantitative results using the mean Hausdorff distance between a human reference trajectory and evaluation trajectories of each method. Table \ref{table:aligned_percent} shows quantitative results for the mean percentage of preference-aligned distance traversed in each trajectory. Note that both the reported metrics may be high if a method does not reach the goal but stays on operator-preferred terrain. 

From the quantitative results, we see that, as expected, the $\patern{}^-$ approach is able to successfully navigate in an operator-preference-aligned manner in Environment 1, which did not contain any novel terrain types. However, $\patern{}^-$ fails to consistently reach the goal and/or navigate in alignment with operator preferences in the remaining environments.
In the daytime experiments, Environment 2 contains a novel terrain (\texttt{pebble pavement}) absent from training data for $\patern{}^-$, while Environment 3 contains both a novel terrain type (\texttt{bush}) and novel visual terrain appearances caused by tree shadows. In the nighttime experiments, all terrains contain novel visual appearances. Following deployments in Environments 2 and 3, \patern{} extrapolates terrain preferences for new visual data using the corresponding proprioceptive data to retrain environment-specific $\patern{}^+$ instances.
In each environment that the $\patern{}^-$ model fails, the self-supervised $\patern{}^+$ model is able to successfully navigate to the respective goal in a preference-aligned manner, without requiring any additional operator feedback during deployment, addressing $Q_1$ and $Q_2$. While the fully-supervised baseline more closely resembles the human reference trajectory compared to $\patern{}^+$ during the day in Environments 2 and 3, unlike the fully-supervised approach, $\patern{}^+$ does not require operator preferences over all terrains and is capable of extrapolating to visually novel terrains.

\subsubsection{Qualitative Large-Scale Experiment}

\begin{figure*}[h]
    \centering
    \includegraphics[width=\columnwidth]{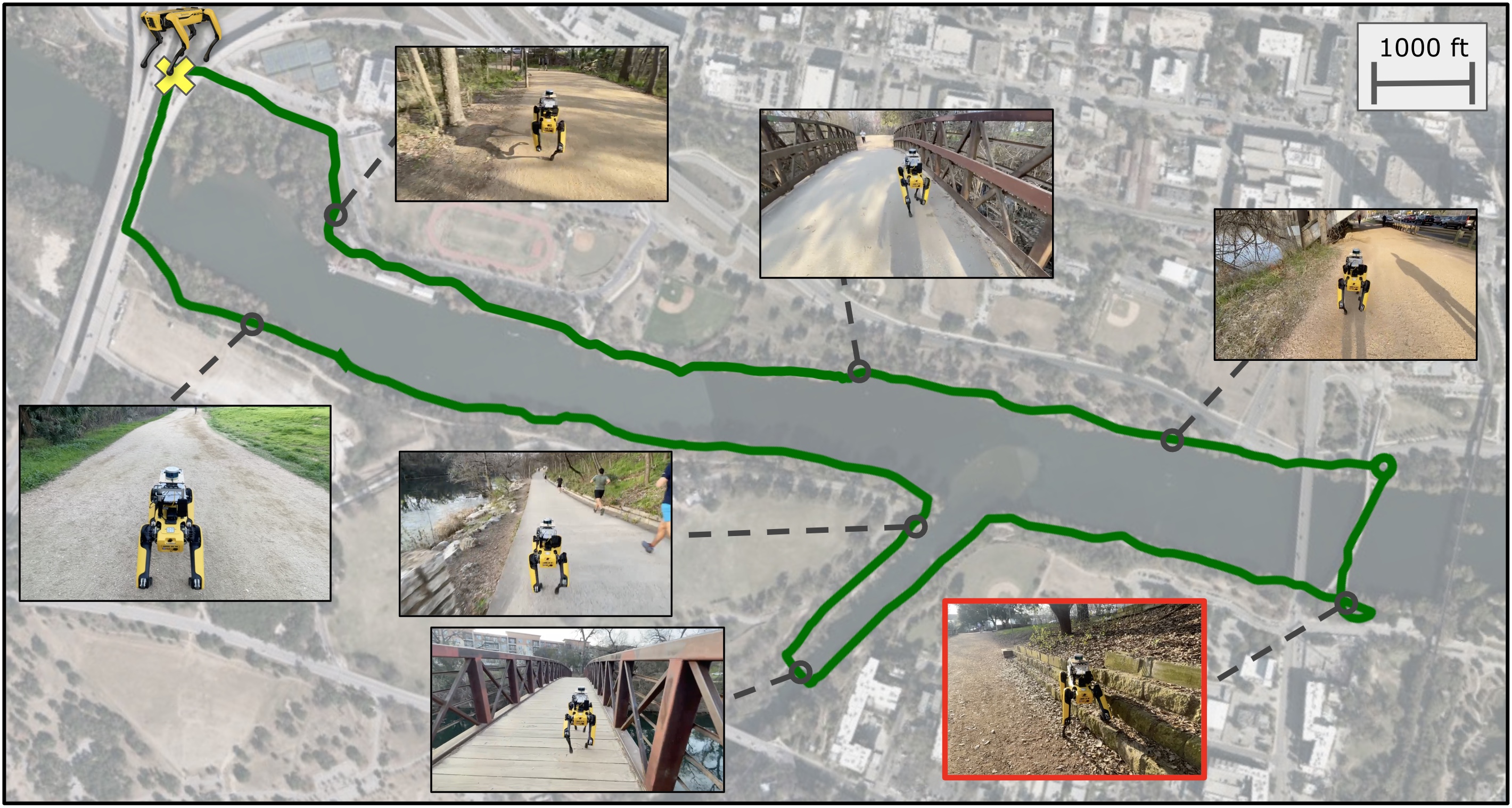}
    \caption{
        Trajectory trace of a large-scale qualitative deployment of $\patern{}^+$ along a 3-mile segment of the Ann and Roy Butler trail located in Austin, Texas. With only five minutes of supplementary data, \patern{} required only one manual intervention to stay on the trail and successfully completes the hike, demonstrating robustness and adaptability to real-world off-road conditions. 
    }
    \label{fig:large_scale}
\end{figure*}

To investigate $Q_3$, we execute a large-scale autonomous deployment of \patern{} along a challenging 3-mile off-road trail.\footnote{Ann and Roy Butler trail, Austin, TX, USA} The robot's objective is to navigate in a terrain-aware manner on the trail by preferring \texttt{dirt}, \texttt{gravel} and \texttt{concrete} over \texttt{bush}, \texttt{mulch}, and \texttt{rocks}.
Failure to navigate in a preference-aligned manner may cause catastrophic effects such as falling into the river next to the trail.
An operator is allowed to temporarily take manual control of the robot only to prevent such catastrophic effects, adjust the robot's heading for forks in the trail, or yield to pedestrians and cyclists.
The $\patern{}^-$ model used for short-scale experiments is augmented with approximately five minutes of combined additional data for \texttt{dirt, bush,} and \texttt{mulch} terrains commonly seen in the trail. Following this preference extrapolation, the $\patern{}^+$ model is able to successfully navigate the 3-mile trail, while only requiring one human intervention. Figure \ref{fig:large_scale} shows the trajectory of the robot and a number of settings along the trail, including the single unexpected terrain-related intervention in the lower right corner, for the hour-long deployment. \footnote{\href{https://youtu.be/j7159pE0u6s}{\patern{} A video footage of the robot being deployed in the trail can be accessed here: https://youtu.be/j7159pE0u6s}}. This large-scale study addresses $Q_3$ by qualitatively demonstrating the effectiveness of \patern{} in scaling to real-world off-road conditions.

\subsection{Limitations}

\patern{} is limited to robot configurations equipped with multi-modal sensing capabilities such as proprioception, depth, inertial, and tactile sensing alongside visual sensors. Additionally, \patern{} uses similarities between novel and known terrains in its learned proprioception representation space to extrapolate preferences. Thus, \patern{} needs to have had experiences with terrains bearing close inertial-proprioceptive-tactile resemblances for successful extrapolation. A noticeable limitation is that if a terrain evoking similar proprioceptive features has not been encountered before, \patern{} might not be able to extrapolate preferences. To extrapolate preferences, \patern{} requires a robot to physically drive over terrains, which may be unsafe or infeasible in certain cases. Extending \patern{} with depth sensors to handle non-flat terrains is a promising direction for future work. 

\section{Summary}
\label{sec:summaryoffroad}
In this chapter, as outlined in Section \ref{contrib2}, I introduce two novel algorithms that enable a mobile robot operating in off-road environments to align its navigation objectives with a human operator's preference over terrain. 

In the first contribution of this chapter, I introduced \textit{Self-supervised TErrain Representation LearnING} (\sterling{}) \citep{sterling}, a novel framework for learning terrain representations from easy-to-collect, unconstrained (e.g., non-expert), and unlabeled robot experience. \sterling{} utilizes non-contrastive representation learning through viewpoint invariance and multi-modal correlation self-supervision objectives to learn relevant terrain representations for visual navigation. We show how features learned through \sterling{} can be utilized to learn operator preferences over terrains and integrated within a planner for preference-aligned navigation. We evaluate \sterling{} against state-of-the-art alternatives on the task of preference-aligned visual navigation on a Spot robot and find that \sterling{} outperforms other methods and performs on par with a fully supervised baseline. We additionally perform a qualitative large-scale experiment by successfully hiking a 3-mile-long trail using \sterling{}, demonstrating its robustness to off-road conditions in the real world.

In the second contribution of this chapter, I presented \textit{Preference extrApolation for Terrain-awarE Robot Navigation} (\patern{}) \citep{paternarxiv}, a novel approach to extrapolate human preferences for novel terrains in visual off-road navigation. A limitation of \sterling{} is that it requires human preference feedback for all observed terrains, which might be unable to acquire in certain conditions. \patern{} extends \sterling{}, addressing this limitation by learning an inertial-proprioceptive-tactile representation space to detect similarities between visually novel terrains and the set of known terrains. Through this self-supervision, \patern{} successfully extrapolates operator preferences for visually novel terrain segments, without requiring additional human feedback. Through extensive physical robot experiments in challenging outdoor environments in varied lighting conditions, we find that \patern{} successfully extrapolates preferences for visually novel terrains and is scalable to real-world off-road conditions. 

\textbf{Comparison with VOILA:} The selection between \voila{}, \sterling{}, and \patern{} depends on the nature of available human feedback and the suitability of online learning to the specific navigation task. \voila{} is an online learning algorithm that learns a reactive navigation policy through reinforcement learning, imitating a video-only navigation demonstration in the same environment. In the case of the unavailability of such constrained video-only demonstrations, or lack of access to the demonstration environment necessary for online learning, \voila{}'s applicability is limited. \sterling{}, in contrast, is a self-supervised representation learning algorithm that learns relevant terrain representations from unconstrained robot experience. These representations allow offline querying of an operator's terrain preferences, informing the development of a costmap that aligns off-road navigation with operator preferences. To enable such preference-aligned navigation, \sterling{} requires access to an operator to query their relative terrain preferences for all observed terrains. \patern{} complements \sterling{} by enabling extrapolation of operator preferences from known to visually novel terrains, thereby reducing the requirement of consistent human feedback for certain terrains. 

The three algorithms \voila{}, \sterling{}, and \patern{} significantly advance the state-of-the-art in robot navigation in static indoor and outdoor environments, effectively addressing the value alignment problem. However, autonomous mobile robots deployed in the physical world frequently face dynamic scenarios, such as navigating around pedestrians, necessitating safe and socially compliant behavior in line with human preferences, as outlined by \cite{francis2023principles}. Towards enabling social compliance of robot navigation in the presence of humans, the forthcoming chapter introduces a comprehensive dataset named \scand{} \citep{scand}, consisting of demonstrations for socially compliant robot navigation. Additionally, it presents a novel hybrid algorithm designed to foster social compliance in autonomous mobile robots, combining the benefits of learning-based and heuristic-based navigation methods.
\chapter{Socially Compliant Robot Navigation}
\label{chap:socially_compliant_nav}

In this chapter, I introduce the third contribution of this dissertation, as outlined in Section \ref{contrib3}, focused on the problem of socially compliant mobile robot navigation in the presence of dynamic agents such as humans. The contributions in this chapter consist of two parts. First, I introduce a large-scale dataset of demonstrations for socially compliant robot navigation behaviors called \scand{} \citep{scand}. We show that navigation policies learned through imitation learning using \scand{} are perceived to be safer and more socially compliant than a classical navigation stack. Second, I introduce a hybrid algorithm for socially compliant robot navigation \citep{raj2023targeted}, combining the benefits of a learning-based and a classical navigation algorithm. The chapter is organized as follows. Section \ref{sec:intro_to_soc_nav_problem} provides an introduction to the problem addressed in this chapter. Section \ref{sec:related_work_soc_nav} reviews relevant related work in datasets and algorithms for socially compliant robot navigation, followed by Sections \ref{sec:data_collection_procedure_scand}, \ref{scand_analysis}, and \ref{sec:hybrid_approach} that introduce the dataset and algorithmic contributions of this chapter, respectively. 

\section{Introduction}
\label{sec:intro_to_soc_nav_problem}
In this section, I introduce the concept of socially compliant robot navigation and the need for a dataset of human-teleoperated demonstrations for socially compliant robot navigation. Additionally, I introduce limitations with using either a classical or learning-based navigation stack for enabling socially compliant navigation and highlight our contribution consisting of a hybrid approach combining classical and learning-based navigation policies for mobile robot navigation.

Social navigation is the capability of an autonomous agent to navigate in a socially compliant manner such that it recognizes and reacts to the objectives of other navigating agents, at least somewhat adjusting its own path in response, while also projecting signals that can help the other agents reciprocate. Enabling mobile robots to navigate in a socially compliant manner has been a subject of great interest recently in the robotics and learning communities \citep{socialforce, xuesusurvey, chen2018socially, collavoideverett, socialnavsurvey}. Towards enabling this capability, demonstration data of socially compliant navigation for mobile robots, such as the ones shown in Figure \ref{fig:demo_speedway}, can be a valuable resource. For instance, such demonstration information can be used to learn socially compliant robot navigation using the paradigm of Learning from Demonstrations (\textsc{l}f\textsc{d}) \citep{lfdpaper, argall2009survey} or understanding human navigation in the presence of autonomous robots \citep{jackrabbot}.

\begin{figure*}[!tb]
    \centering
    \includegraphics[width=0.8\columnwidth]{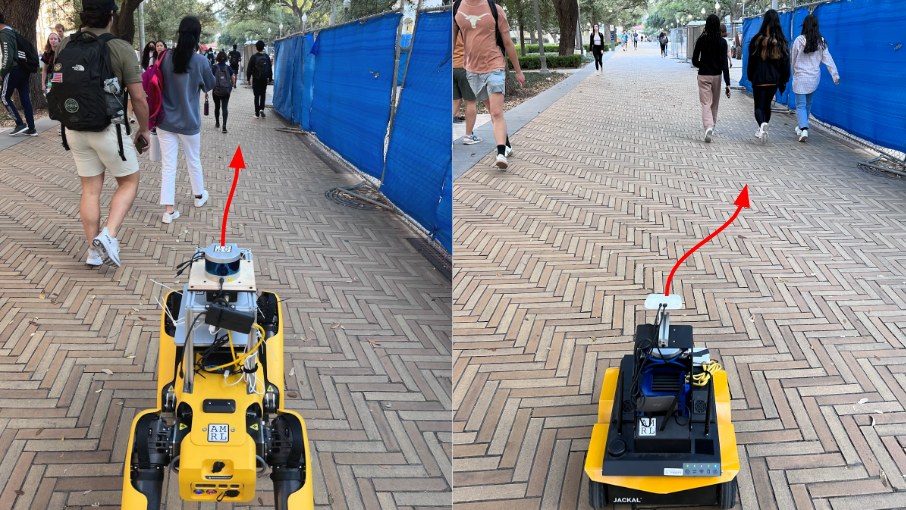}
    \caption{A human demonstrator teleoperates the two robots, following a socially compliant strategy (left- moving with traffic, right- sticking to the right of the road) around human crowds.}
    \label{fig:demo_speedway}
\end{figure*} 

Datasets for social navigation, generally used for learning and benchmarking, include data collected both in the real world \citep{thorDataset2019} and in simulated environments \citep{tsoi2020sean, manso2020socnav1}. While such datasets provide basic trajectories of the robots and humans, they either contain limited interactions in constrained, orchestrated environments or restrict themselves to indoor-only navigation scenarios. When collecting data in such controlled settings \citep{thorDataset2019}, naturally occurring social interactions including---but not limited to---following lane rules of a country, yielding to pedestrians and vehicles, walking with and against a crowd of people, and street crossing are not captured.
Additionally, the robots used for data collection in previous social navigation datasets \citep{thorDataset2019} tend to use a simple controller for point-to-point navigation that does not explicitly exhibit socially aware navigation. 

Recently, imitation learning has emerged as a useful paradigm for designing mobile robot navigation controllers \citep{byronboots, voila, bojarskie2e, offroadoa}. In this paradigm, the desired navigation behavior is first demonstrated by an agent such as a human, the recording of which is then utilized by an imitation learning algorithm to imitate. This intuitive way of teaching a task to a robot is also easy for non-expert humans since it only requires providing demonstrations, instead of defining the rules of the task itself, which may be hard to explicitly define for social navigation. Motivated by recent successes of imitation learning in robot navigation, we posit that one way to enable autonomous agents to navigate socially is through learning from human demonstrations of socially compliant navigation behavior. However, there is a lack of large-scale datasets containing socially compliant navigation demonstrations in the wild that can be utilized for imitation learning.

To fill this gap, in Section \ref{sec:data_collection_procedure_scand}, I introduce a dataset of demonstrations for socially compliant robot navigation in the wild. Our dataset contains 8.7 hours of human-teleoperated, socially compliant, navigation demonstrations, specifically, Velodyne lidar scans, joystick commands, odometry, camera visuals, and 6D inertial (IMU) information collected on two morphologically different mobile robots---a Clearpath Jackal and a Boston Dynamics Spot---within the University of Texas at Austin university campus. Comprising 25 miles in total over 138 trajectories, Socially CompliAnt Navigation Dataset (\textsc{scand}) \citep{scand} is publicly released\footnote{\href{www.cs.utexas.edu/~xiao/SCAND/SCAND.html}{www.cs.utexas.edu/$\sim$xiao/SCAND/SCAND.html}} and also contains labeled tags of naturally occurring social interactions with every trajectory. 
We demonstrate the utility of the dataset for studying questions relevant to social navigation. We show that a navigation policy learned using imitation learning on the \scand{} dataset leads to policies that are perceived by human participants to be more socially compliant and safe than the ROS \movebase{} navigation stack. 

As the second contribution of this chapter, in Section \ref{sec:hybrid_approach}, I leverage the insight that although a mobile robot can learn simple social navigation behaviors end-to-end as shown in Section \ref{scand_imitation_learning}, the majority of navigation scenarios encountered by a mobile robot do not require such a specialized end-to-end learned policy. Additionally, without the safety guarantees present in classical navigation stacks, such a policy may not be reliable to use in all scenarios. Our analysis on \scand{} reveals that classical navigation systems adequately cater to a significant proportion---up to 80\%---of social navigation scenarios in the \scand{} dataset (Figure~\ref{fig:cover} left), underscoring the importance of classical navigation systems in routine social situations. To address the limitations identified, the second key contribution of this chapter proposes a novel hybrid approach. This method alternates between a classical geometric planner and a data-driven planner, effectively balancing the strengths of both through a data-driven approach. Our comparative analysis demonstrates that this hybrid model surpasses both classical and learning-based methods when evaluated independently, especially in terms of social compliance across diverse metrics. In addition to evaluations on the \scand{} dataset, we also perform a practical human study involving two robots operating in distinct campus environments (Figure~\ref{fig:cover} left).

\begin{figure}[t]
     \centering
         \includegraphics[width=1\columnwidth]{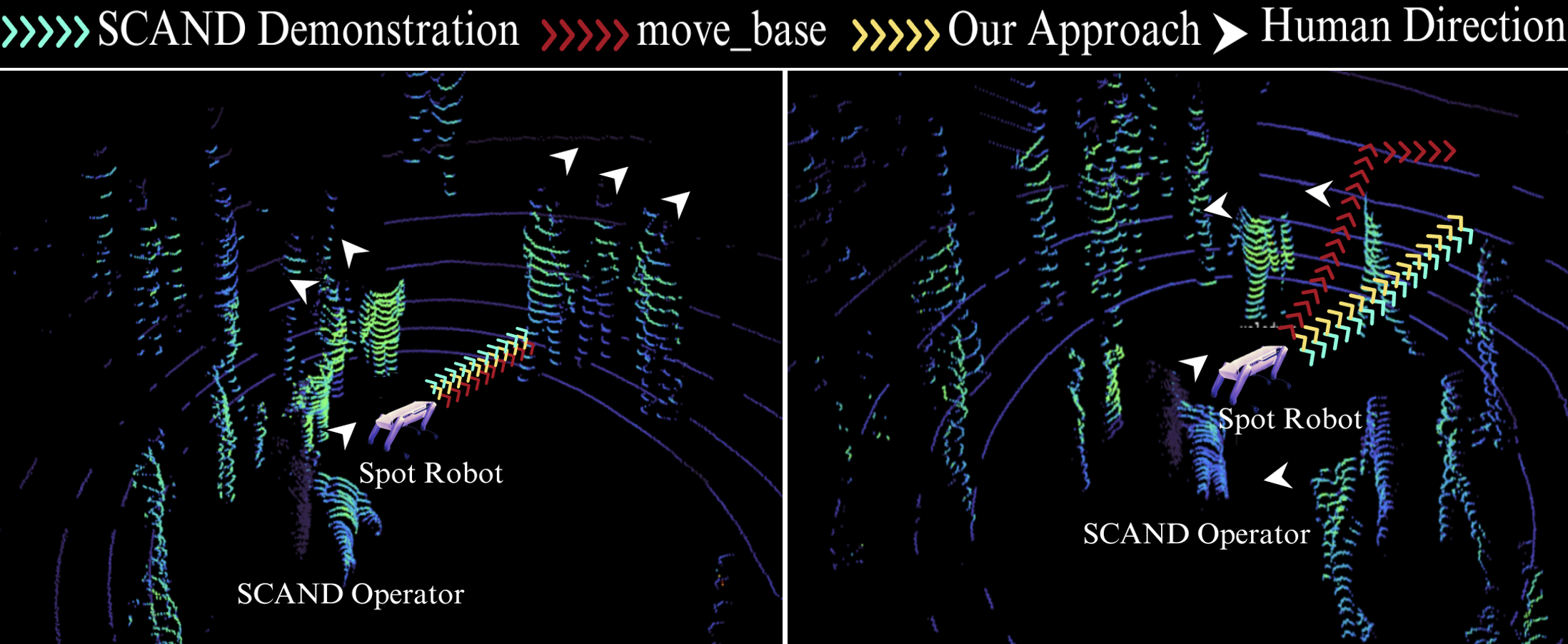}
        \caption{Comparison of the navigation behavior from \textsc{scand} demonstration, \texttt{move\_base}, and our approach at the same time step in two different social scenarios: Classical \texttt{move\_base} (red) aligns with (left) or deviates from (right) the \textsc{scand} demonstration, while our approach is always close to the socially compliant demonstration. }
        \label{fig:cover}
\end{figure}

\section{Related Work}
\label{sec:related_work_soc_nav}
In this section, I review related literature on classical and learning-based approaches for social navigation. I additionally survey relevant datasets for robot navigation and contrast their contributions with this work. For a more general overview of related approaches to social navigation, refer to Section \ref{sec:ml4socnav}.

\subsection{Classical Methods for Social Navigation}
\label{sec:related_classical_soc}
For several decades, roboticists have been developing an array of classical navigation systems to facilitate collision-free robot movement from one location to another. These systems typically employ a global planner, such as A*~\citep{Hart1968} or D*\citep{ferguson2006using}, to outline a high-level path, augmented by a local planner\citep{fox1997dynamic, quinlan1993elastic} for generating detailed motion commands. This combination ensures adherence to the global route while navigating around obstacles. Classical navigation systems generally necessitate a pre-defined cost function~\citep{lu2014layered}, balancing various factors like path length, obstacle clearance, energy efficiency, and increasingly, social compliance. 
To execute these functions, methodologies such as sampling-based~\citep{kavraki1996probabilistic, kuffner2000rrt, fox1997dynamic}, optimization-based~\citep{ratliff2009chomp, quinlan1993elastic}, or potential-field-based~\citep{koren1991potential} techniques are utilized for motion command generation. These approaches are valued for their safety, explainability, and testability attributes, which are either provably or empirically optimal. Such qualities are crucial for real-world applications of physical robots, especially in human-centric environments, thus maintaining the popularity of classical navigation systems in the robotics industry~\citep{Amazonscout, starship, dilligent, tinymile}.
However, adapting these classical approaches to socially complex environments often involves significant engineering efforts, including the manual design of cost functions~\citep{lu2014layered} and the fine-tuning of navigation parameters~\citep{zheng2021ros}. These challenges have spurred interest in exploring learning-based methodologies for improving the social compliance of autonomous mobile robots.

\subsection{Machine Learning for Social Robot Navigation}

Recently, several algorithms have emerged that show the potential of applying machine learning to address challenges in social robot navigation \citep{xuesusurvey, socialnavsurvey, francis2023principles}. Broadly speaking, in the robot navigation literature, learning-based approaches have been shown to be successful in problems such as adaptive planner parameter learning \citep{xiao2021appl}, overcoming viewpoint invariance in demonstrations \citep{voila}, and end-to-end learning for autonomous driving \citep{bojarskie2e, pfeiffer2018reinforced, wang2021agile}. Specifically in applying imitation learning for social navigation, the work by \cite{tai2018socially} is the closest to our work. They provide a simulation framework in gazebo \citep{gazebo} along with a dataset generated using the same where virtual human agents navigate following the social force model \citep{socialforce}. They additionally train a social navigation policy using the Generative Adversarial Imitation Learning algorithm assuming the social force model as the ``expert" demonstrator and show a successful deployment of the learned policy in the real world on a TurtleBot~\citep{turtlebot}. While their work has shown that imitation learning can be applied to address the social navigation problem, they do so assuming the social force model in simulation as the ``expert" demonstration. While simulated environments enable fast and safe data collection for online learning, they lack the naturally occurring social interactions seen in the wild. Also, as I show in Section \ref{sec:demonstrator_classification}, there can be more than one strategy for an agent to navigate socially in a scene, which is not considered in their work. 

Other learning paradigms such as Reinforcement Learning (\textsc{rl}) have also been applied to address the social navigation problem. \cite{collavoideverett} present \textsc{ca}-\textsc{drl}, a multi-agent collision avoidance algorithm learned using \textsc{rl}. While this work shows impressive real-world results, their approach is limited to specific social scenarios and requires simulating these scenarios for the online learning algorithm to learn episodically. \cite{soc_com_robot_nav_paper} use Inverse Reinforcement Learning to learn cost functions for a socially compliant navigation policy. Similar to our work, they utilize human demonstrations of the social navigation task, however, they do so utilizing a small-scale, one-hour-long dataset. In this work, we contribute a large-scale dataset of robot social navigation demonstrations comprising multi-modal real-world data over multiple hours, both indoors and outdoors, on two different robots. Additionally, we train an imitation learning algorithm to show it is possible to learn socially compliant global and local navigation policies using our dataset.

\subsection{Datasets for Social Navigation}
Over the last decade, datasets containing robots navigating in both simulated and real-world environments have been useful for a wide variety of research areas, such as tracking groups of people \citep{thorDataset2019, lau2009track, linder2016multi}, human trajectory prediction \citep{chung2012incremental}, navigation \citep{hirose2019deep}, robot localization \citep{biswas2013longterm, kitti} and collision risk assessment \citep{lo2019robust}.

\noindent\textbf{Simulated Datasets for Social Navigation:}
Social environments in simulation can provide researchers with fast data collection on social navigation \citep{tsoi2020sean, tsoi2021approach, holtz2021socialgym, tai2018socially}. Moreover, such simulated environments can be generated with a specified number of elements: the number and locations of the humans, the structure of the room, the number of objects, and the interactions between people and between objects and people \citep{manso2020socnav1}. While simulated platforms provide these benefits, they are limited in that they lack the natural, real-world interactions that are experienced by humans. Datasets that capture real-world robot navigation data in the wild provide researchers with more naturally occurring scenarios \citep{biswas2013longterm,zhimon2020jist,yz17iros,carlevaris2016university}. Additionally, datasets collected in the wild provide sensory data for these scenarios which can be then used for perceptual tasks related to navigation \citep{de2021deepsocnav}. 

\noindent\textbf{Real-world Datasets for Robot Navigation:} In addition to simulated datasets, several real-world datasets for long-term robot navigation in human environments have also been made available over the last decade. In the CoBots dataset \citep{biswas2013longterm}, two CoBots were deployed indoors autonomously using a topological graph planner and collected more than 130 km worth of laser scans, odometry, and localization data over 1082 deployments. Similarly, the L-CAS \citep{yz17iros}, FLOBOT \citep{zhimon2020jist}, JRDB \citep{jackrabbot} and NCLT \citep{carlevaris2016university} datasets contain LiDAR scans, RGBD visuals, GPS, and IMU data collected independently on different robots, addressing perception-related challenges to long-term robot navigation. In all these different datasets, the robots were deployed in a public environment, such as a restaurant or a university campus, and teleoperated by a human as opposed to being autonomous, but these teleoperated demonstrations are not explicitly socially compliant. The JRDB social navigation dataset by \cite{jackrabbot} is the closest to our work, but it is smaller in scale, containing only 64 minutes worth of data from 54 indoor and outdoor trajectories. While the focus of the JRDB dataset is to solve perception-related challenges such as human tracking and detection in social navigation, the focus of the \textsc{scand} dataset in this work is to address the ``navigation" sub-component of social navigation. The TH{\"O}R dataset by \cite{thorDataset2019} provides motion trajectories of both robots and humans using tracking helmets. However, this is smaller in scale since it contains only one hour's worth of data. Also, the data is collected indoors in an 8.4x18.8m laboratory room with an orchestrated social navigation scenario for the human agents in the scene and a socially unaware, pre-defined path for the robot---adjusting neither its speed nor trajectory to account for surrounding people. Existing real-world datasets for robot navigation are summarized in Table \ref{existing_datasets}.

While previous datasets collected with robots and humans have proven to be useful to study localization, perception, and other navigation-related challenges, they lack demonstration information in the form of motion commands and navigation strategies in different social scenarios that could help us understand socially compliant robot navigation in the presence of other autonomous agents. The \textsc{scand} dataset introduced in this work addresses this gap and provides rich human demonstration information in the form of joystick commands and multi-modal robot sensor data in different, naturally occurring social scenarios. \textsc{scand} also contains labeled tags of twelve different types of social interactions, as described in Table \ref{table1}, that occurred along the paths. Also, since robots of different morphologies and capabilities could navigate differently and induce different social interactions, \textsc{scand} also includes data from two different robots. For example, the legged Spot, capable of climbing stairs could choose to prefer the stairs along its path while navigating whereas the wheeled Jackal might choose a ramp to navigate. The other datasets use only one robot to collect data (the Cobots dataset from \cite{biswas2013longterm} uses two robots but they are morphologically the same). Using two morphologically different robots makes \textsc{scand} useful to investigate social navigation in robots with different morphologies (wheeled vs. legged).

\begin{sidewaystable}[]
\centering
\caption{Comparison of real-world datasets for robot navigation.}
\begin{tabular}{>{\centering\arraybackslash}m{1.2cm}>{\centering\arraybackslash}m{1.6cm}>{\centering\arraybackslash}m{1.3cm}>{\centering\arraybackslash}m{1.3cm}>{\centering\arraybackslash}m{5.2cm}>{\centering\arraybackslash}m{1.6cm}>{\centering\arraybackslash}m{1.6cm}>{\centering\arraybackslash}m{1.2cm}}
\toprule
\textbf{Dataset} & \textbf{\# Traj.} & \textbf{Dist. (Km)} & \textbf{Dur. (min)} & \textbf{Sensors}                                        & \textbf{Nav. method}  & \# \textbf{Robots} & \textbf{Location}             \\
\toprule
 CoBot \cite{biswas2013longterm} & 1082 & 131 & 15600 & 2D Range Scanner, RGB-D Camera, Wheel Odometry & Autonomous & 2 & Indoors + Outdoors \\
 \midrule
    L-CAS \cite{yz17iros} &
  3 &
  N/A &
  49 &
  3D LiDAR &
  Teleoperated &
  1 &
  Indoors \\
  \midrule
  NCLT \cite{carlevaris2016university} &
  27 &
  147.4 &
  2094 &
  3D LiDAR, RGB Camera, IMU, Wheel Odometry, GPS &
  Teleoperated &
  1 &
  Indoors + Outdoors \\
  \midrule
   
  FLOBOT \cite{zhimon2020jist} &
  6 &
  N/A &
  27.5 &
    3D LiDAR, RGB-D camera, Stereo Camera, 2D LiDAR, OEM incremental measuring wheel encoder, IMU &
    
    Autonomous &
  1 &
  Indoors \\
  \midrule

JRDB \cite{jackrabbot} &
  54&
  N/A&
  64&
  3D LiDAR, 2D LiDAR, Omnidirectional Stereo Suite, RGB camera, RGB-D stereo camera, 6D IMU &
    Teleoperated &
  1 &
  Indoors + Outdoors \\
  \midrule

TH{\"O}R \cite{thorDataset2019} &
  600 &
  N/A &
  60 &
  3D LiDAR, Motion capture system, Eye-tracking Glasses &
  Autonomous &
  1 &
  Indoors \\
  \midrule
  
  SCAND &
  138 &
  40 &
  522 &
  3D LiDAR, RGB-D Camera, Monocular Camera, Stereo Camera, Wheel Odometry, Visual Odometry &
  Teleoperated &
  2 &
  Indoors + Outdoors \\
\bottomrule

\end{tabular}
\label{existing_datasets}
\end{sidewaystable}

\section{Data Collection Procedure}
\label{sec:data_collection_procedure_scand}
In this section, I outline the data collection procedure used in \textsc{scand} and describe the sensor suite present on both robots. I then describe the labeled annotations of social interactions provided with every trajectory.\footnote{The research described in this section was done in collaboration with Anirudh Nair, Xuesu Xiao, Garrett Warnell, Alexander Toshev, Soeren Pirk, Justin Hart, Joydeep Biswas, and Peter Stone.}

\subsection{Collecting Data}
To collect multi-modal, socially compliant demonstration data for robot navigation, four human demonstrators navigate the robot by teleoperation using a joystick. We collected data within the UT Austin University campus, with the demographics of the humans in the scene comprised mostly of students, faculty, and other campus denizens.
For each of the 138 trajectories in \textsc{scand}, the human demonstrator walks behind the robot at all times, maintaining on average a distance of 2 meters. The human demonstrator does not explicitly interact with the crowd in the scene. Unlike other datasets for social navigation \citep{thorDataset2019}, we do not restrict data collection to a controlled, indoor environment or orchestrate a social scenario for data collection. Instead, similar to the JRDB dataset \citep{jackrabbot}, we perform data collection in the wild in both indoor and outdoor environments.  The two robots are driven around the university campus on frequently used sidewalks, roads, and lawns, and inside buildings, all with people on the scene during peak hours of high foot traffic. Data collection extends to outdoor areas adjacent to the university's football stadium on game days, capturing diverse interactions with dense, public crowds near the venue. The Spot is driven at linear and angular velocities in the range of $[0, 1.6]$ $m/s$ and $[-1.5, 1.5]$ $rad/s$, respectively, and the Jackal is driven in the range of $[0, 2.0]$ $m/s$ and $[-1.5, 1.5]$ $rad/s$, respectively. Note that these velocities are within the range of many people's normal walking speed.  

Figure \ref{fig:sensor_suite} shows the sensors present on the Clearpath Jackal and the Boston Dynamics Spot robots. Both robots have in common a VLP-16 Velodyne laser puck publishing at a frequency of 10 Hz, a 6D inertial (IMU) sensor at 16 Hz, and a front-facing Azure Kinect RGB camera at 20 Hz. In addition to these common sensors, the Jackal has a front-facing stereo camera (20 Hz) and wheel odometry (30 Hz), while the Spot has five monocular cameras on its body (publishing at 5 Hz), placed as shown in Figure \ref{fig:sensor_suite}. We utilize the Boston Dynamics APK to record the visual odometry of its body frame and the joint angles of the legs on the robot. \textsc{scand} also contains transforms between the frames of each of the sensors relative to the robot's body for both robots. We utilize AMRL's software stack \citep{graphnavgithub} for data collection from different sensors which we record in the rosbag format \citep{ros}.

Although we provide visual information of the scene in the form of surround-view monocular images on the Spot, RGB images from the front-facing Kinect camera, and 3D Velodyne laser scans on both robots, since the focus of this work is specifically on navigation, we do not provide any labeled annotations for human detection or tracking. We refer the reader to the JRDB dataset \citep{jackrabbot} which contains detailed, high-quality annotations for solving perception-related tasks. Instead, \textsc{scand} contains joystick commands of linear and angular velocities executed by the demonstrator while teleoperating the robot socially, along with rich, multi-modal sensory information of the environment including labeled annotations of 12 different social interactions in every trajectory. Figure \ref{fig:intersting_social_interactions} shows five example scenarios and their associated tags.

\begin{figure*}[!tb]
    \centering
    \includegraphics[width=\columnwidth]{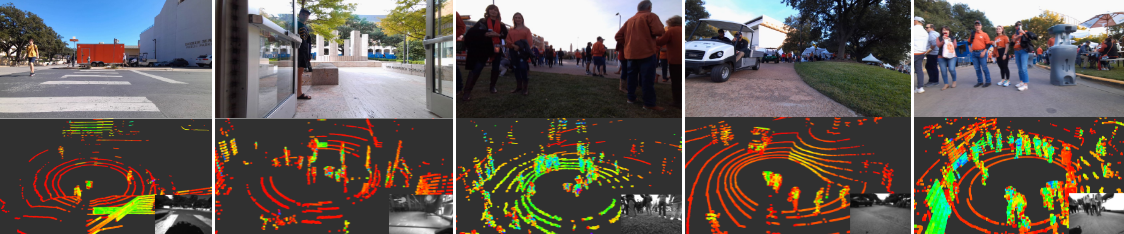}
    \caption{Five example scenarios from \textsc{scand} showing the RGB image and below it the accompanying Lidar with the monocular image from the side camera on the Spot. From left to right, the scenarios have the tags ``Street Crossing", ``Narrow Doorway, ``Navigating Through Large Crowds", ``Vehicle Interaction", and ``Crossing Stationary Queue."}
    \label{fig:intersting_social_interactions}
\end{figure*} 

\begin{table}[!h]
\centering
\caption{Descriptions of labeled tags contained in \textsc{scand}}
\begin{tabular}{>{\centering\arraybackslash}m{0.25\textwidth}>{\centering\arraybackslash}m{0.35\textwidth}>{\centering\arraybackslash}m{0.25\textwidth}}
\toprule
 \textbf{Tag} & \textbf{Description} &
 \textbf{\# Tags} \\
 \toprule
 Against Traffic & Navigating against oncoming traffic & 22 \\ 
 \midrule
 With Traffic & Navigating with oncoming traffic & 74 \\
 \midrule
 Street Crossing & Crossing across a street & 34 \\
 \midrule
 Overtaking & Overtaking a person or groups of people & 14 \\
 \midrule
  Sidewalk & Navigating on a sidewalk & 57 \\
  \midrule
  Passing Conversational Groups & Navigating past a group of 2 or more people that are talking amongst themselves & 38 \\
  \midrule
  Blind Corner & Navigating past a corner where the robot cannot see the other side & 6 \\
  \midrule
  Narrow Doorway & Navigating through a doorway where the robot waits for a human to open the door & 15 \\
  \midrule
  Crossing Stationary Queue & Walking across a line of people & 6 \\
  \midrule
  Stairs & Walking up and/or down stairs & 22 \\
  \midrule
  Vehicle Interaction & Navigating around a vehicle & 21 \\
  \midrule
 Navigating Through Large Crowds & Navigating among large unstructured crowds & 27 \\ 
\bottomrule
\end{tabular}

\label{table1} 
\end{table}

\subsection{Labeled Annotations of Social Interactions}

We manually annotate each trajectory in \textsc{scand} with labels describing social interactions that occurred along the path. The labels are in the form of a list of textual captions of social interactions taking place in a trajectory, chosen from a set of twelve predefined labels of social interactions observed in \textsc{scand}. For the full list of labels, refer to Table ~\ref{table1}. We intend the labels to be useful for future studies of specific scenarios that occur during social navigation in the real world. 

\begin{figure*}[!tb]
    \centering
    \includegraphics[scale=0.22]{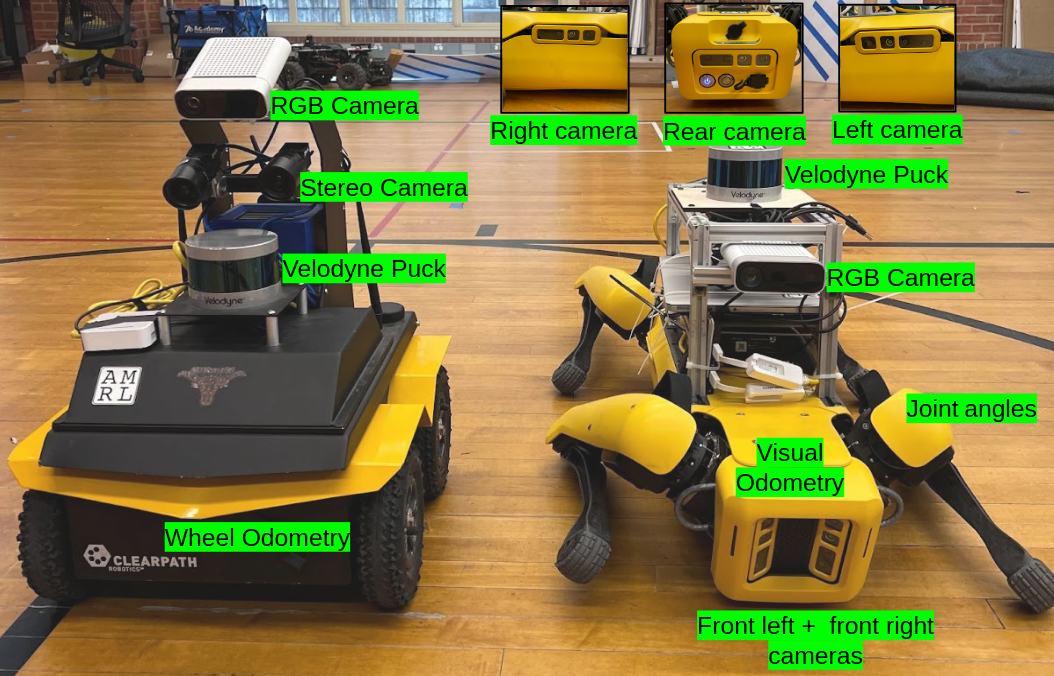}
    \caption{Sensors present on the wheeled Jackal and the legged Spot robots. Along with this multi-modal sensor information, \textsc{scand} also contains joystick commands issued during the navigation demonstration.}
    \label{fig:sensor_suite}
\end{figure*}

\section{SCAND Dataset Analysis}
\label{scand_analysis}




In this section, I present several analyses utilizing \scand{} to explore socially compliant robot navigation. First, to analyze the applicability of tools from imitation learning to learn socially compliant navigation policies, in Section \ref{scand_imitation_learning}, we explore behavior cloning navigation policies using \scand{}. Second, in Section \ref{sec:demonstrator_classification}, we evaluate if there is more than one way to navigate in a socially compliant manner. This evaluation is performed by learning a neural network-based classifier trained for the task of demonstrator classification, given a ten-second sequence of sensor observations and joystick commands as input. Finally, in Section \ref{sec:analyze_soc_compliance}, we perform a case study on \scand{} by analyzing existing classical navigation approaches to understand the social compliance they offer, in comparison to the human demonstrations present in \scand{}. The insights from the analysis performed in this section are utilized in Section \ref{sec:hybrid_approach} to develop a novel hybrid approach to social robot navigation.

\subsection{Imitation Learning for Global and Local Planning}
\label{scand_imitation_learning}
In this section, using \scand{}, we explore the question ``Can we learn a local and global planner for social navigation using behavior cloning on \scand{}?". We hypothesize that the answer is yes, and through physical human-robot experiments in an indoor lab environment observe that the policy learned using \scand{} leads to safe and socially compliant navigation behaviors.

\noindent\textbf{Approach and Implementation:} We apply the \textsc{bc} imitation learning algorithm \citep{behaviorcloning} on \textsc{scand} to jointly train end-to-end a socially-aware global and local planner for robot navigation. The objective of the global planner agent is to predict the socially compliant global plan (the future trajectory driven by the human demonstrator, within a horizon of ten meters distance from the robot). The local planner agent's objective is to predict the forward and the angular velocities demonstrated in \textsc{scand} in a socially compliant manner. We jointly train the local and the global planner agents using a common representation space of observations, as shown in Figure \ref{fig:planning_local_global}. However, unlike the demonstrator classifier network with a single classifier head, here we use two different heads (three-layer fully connected networks) for the global and the local planner agents.
As inputs to the \textsc{bc} agent, we provide processed sensor observations from \textsc{scand} of two seconds in length to account for temporal variations in the scene; this includes BEV lidar scans (subsampled to 2 Hz and represented as grayscale BEV image as shown in Figure \ref{fig:planning_local_global}), positions of the previous lidar frames relative to the first lidar frame and inertial information at each of the lidar frames. Additionally, we also provide the global path and desired velocities produced by \texttt{move\textunderscore base} \citep{rosmovebase} using the location of the robot ten meters in the future from its current position as prior information to the network. We posit that feeding this prior information from \texttt{move\textunderscore base} as inputs to the \textsc{bc} agent would enable improved performance. The global planner head predicts 200 points in the path driven by the demonstrator, and the local planner predicts 20 timesteps of joystick commands ($v$, $\omega$) issued by the demonstrator since the current frame. We sum the mean-squared error loss objectives for both agents and update their parameters together. Note that we do not utilize any representation learning algorithm to pretrain the encoders that process the sensor observations, but doing so may potentially improve results. However, since the focus of this analysis is to show the potential of \textsc{scand} in enabling existing imitation learning algorithms to learn socially compliant navigation policies, representation learning is left to future work. 

\begin{figure}[!tb]
    \centering
    \includegraphics[width=\columnwidth]{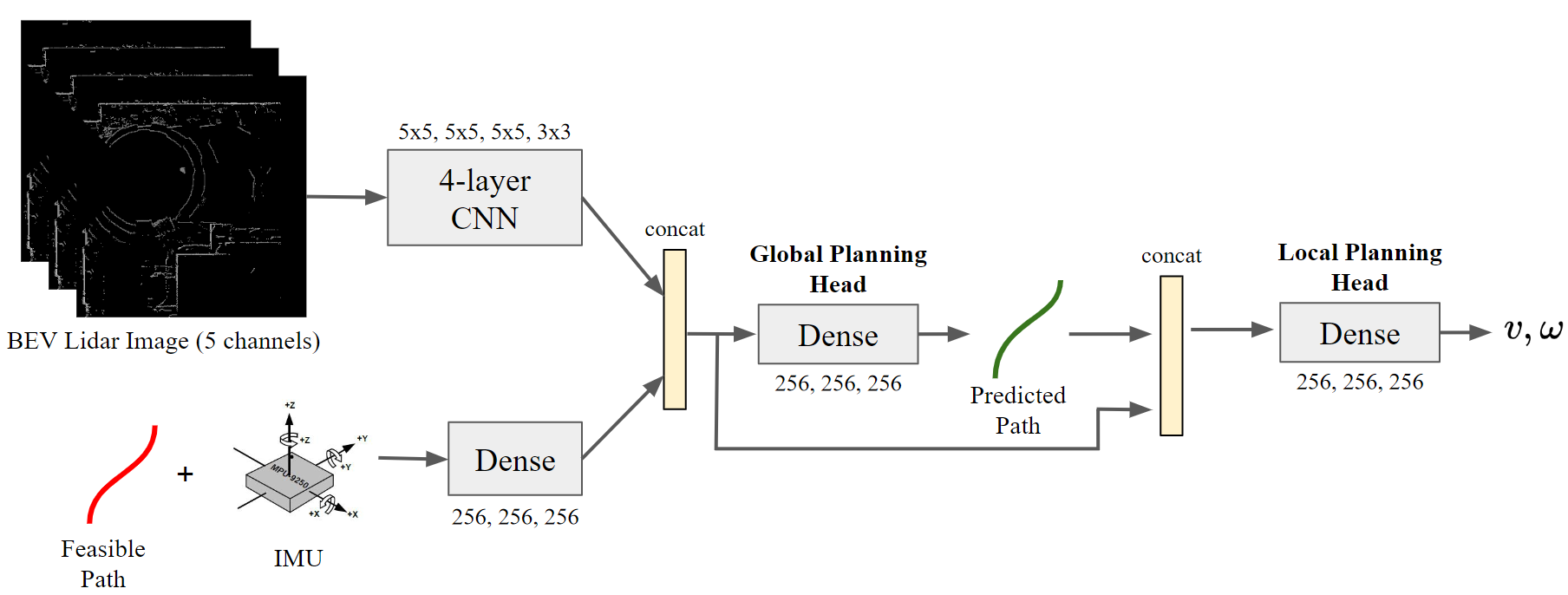}
    \caption{Schematic of the neural network architecture for imitation learning using \scand{}. The network takes as input the past 10 bird's eye view rasterized lidar scans, a kinematically feasible path computed using $A^*$, and time-series IMU observations, employing separate heads for global path planning and local motion planning, trained end-to-end using behavior cloning. The global planning head outputs a predicted path, while the local planning head generates action commands such as linear and angular velocities, trained to match the demonstrations in \scand{}.
    }
    \label{fig:planning_local_global}
\end{figure} 

\begin{figure}[!tb]
    \centering
    \includegraphics[width=0.65\columnwidth]{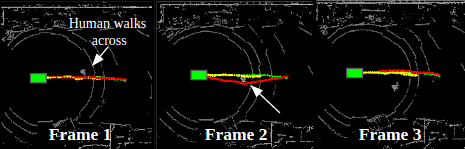}
    \caption{An example sequence of three BEV lidar frames of a human walking across the robot's (green box) path. The green path shows the demonstrated trajectory, the red path shows the \texttt{move\textunderscore base} global path, and the yellow path shows the predicted trajectory by the learned \textsc{bc} agent. In frame 2, the \texttt{move\textunderscore base} path moves in the direction of the human's future state, whereas the learned path closely follows the desired socially compliant path.}
    \label{fig:scand_paths}
\end{figure} 

\noindent\textbf{Results and Conclusion:} To evaluate the social navigation behavior of the global planner, we compute the Hausdorff distance metric on a held-out test set, between the global path predicted by the learned global planner agent and the actual path driven by the demonstrator in the future. The average Hausdorff distance between the \texttt{move\textunderscore base} global path and the demonstrated path in a held-out test set is 1.25. However, after training the \textsc{bc} global planner agent on \textsc{scand}, the average Hausdorff distance between the predicted trajectory and the demonstrated trajectory is improved at 0.26. Figure \ref{fig:scand_paths} shows a scenario involving the robot, and a human walking across the robot's path. We see that in this scenario, the predicted path closely matches that of the socially compliant demonstrated path, whereas \texttt{move\textunderscore base} turns in the direction of the human's future state, creating an undesired interaction.

To validate the learned local planner agent, we conduct real-world experiments using the Spot robot with fourteen human participants in an indoor location. We design two scenarios---static and dynamic---to evaluate the social compliance and safety of the learned local planner and the \texttt{move\textunderscore base} planner, as shown in Figure \ref{fig:scand_human_trial_setup}. In the static scenario, the robot starts five meters ahead of a stationary human in the robot's path, and tries to navigate to a goal position five meters behind the human. In the dynamic scenario, the robot and the human start facing each other 10 meters apart and try to reach the start position of the other. In the dynamic scenario, the participants were asked to navigate in a socially compliant manner to their goal position and in both scenarios, the participants were asked to observe the navigation behavior of the robot. After each scenario, for both the algorithms, a questionnaire was presented to the participant with the two following questions:
\begin{enumerate}
    \item \textit{On a scale of 1 to 5, how ``socially compliant" do you think the robot was? (think of social compliance as how considerate the robot was of your presence)}
    \item \textit{On a scale of 1 to 5, how ``safe" did you feel around the robot?}
\end{enumerate}
We randomized the order in which the two algorithms (\texttt{move\textunderscore base} and \textsc{bc} policy) were played to the participants. Figure \ref{fig:human_trial_metric} shows the responses of the human participants. On average, more humans felt the imitation learning agent trained on \textsc{scand} was more socially compliant (\textsc{scand} mean=$4.39$, sd=$0.99$; \texttt{move\textunderscore base} mean=$2.86$, sd=$0.82$) and safer (\textsc{scand} mean=$4.71$, sd=$0.70$; \texttt{move\textunderscore base} mean=$2.89$, sd=$1.18$) than the \texttt{move\textunderscore base} agent. 
Results for both questions demonstrate statistical significance, as determined by a One-Way Analysis of Variance (ANOVA) (\textit{Safe} $F_{1,55}=47.87, p < 0.001$; \textit{Socially Compliant} $F_{1,55}=38.67, p < 0.001$), aligning with expectations given that the \movebase{} agent was not designed for social compliance.
Refer to the online video\footnote{\href{https://youtu.be/QgBfMjWpQIw}{https://youtu.be/QgBfMjWpQIw}} for scenarios showing the behavior of both the algorithms in the static and dynamic trials. 
The results of this study support our hypothesis that imitation learning using demonstrations provided in \textsc{scand} produces socially compliant navigation policies. To facilitate reproducibility, we have made available the train-test splits of the trajectories collected with the Spot robot in a 75\%-25\% ratio, respectively. These splits, along with the complete dataset, are accessible on Dataverse\footnote{\href{https://dataverse.tdl.org/dataset.xhtml?persistentId=doi:10.18738/T8/0PRYRH}{https://dataverse.tdl.org/dataset.xhtml?persistentId=doi:10.18738/T8/0PRYRH}}.

\noindent\textbf{Limitations:} In this section, I demonstrate that behavior cloning a reactive navigation policy using \scand{} by optimizing the mean-squared error loss objective produces socially compliant and safe navigation behaviors on the robot. Nevertheless, this approach has its constraints. It predicts only a single instance of navigation behavior in a reactive way for each scene, based on the presumption of an unimodal behavior distribution, when in fact, as I show in the following Section \ref{sec:demonstrator_classification}, there can be more than one way to navigate in a socially compliant manner. To extend the capabilities of the imitative navigation policy to handle complex social navigation scenarios that may be multi-modal, alternate approaches that are specifically designed to capture the inherent multi-modality in demonstrations, such as Gaussian Mixture Models \citep{jaquier2020learninggmm}, Inverse Reinforcement Learning \citep{abbeel2004apprenticeship, wulfmeier2015maximum}, or Diffusion Policies \citep{chi2023diffusion} are necessary. Towards this end, I present directions for short-term future work, summarized in Section \ref{sec:multimodal_imitation_learning}.

\begin{figure}[!th]
    \centering
    \includegraphics[width=\columnwidth]{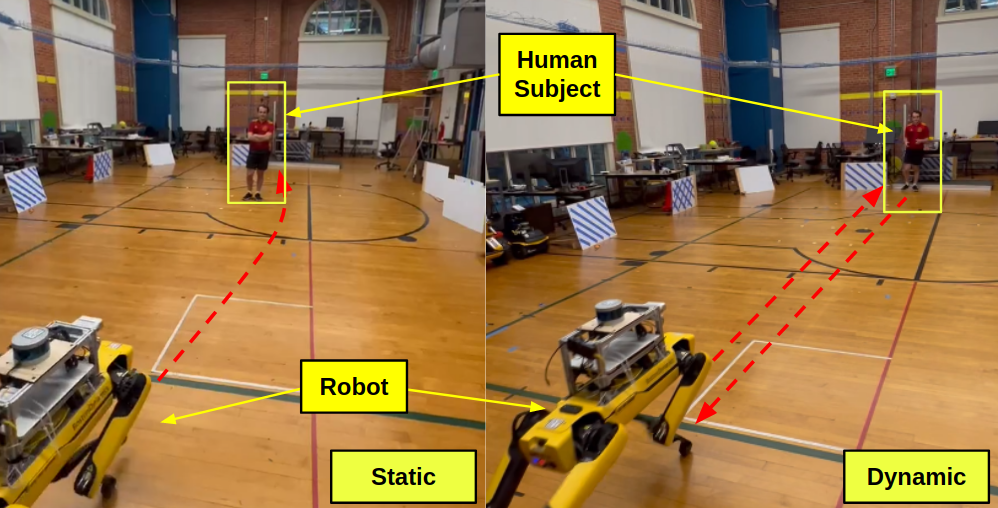}
    \caption{Evaluating the local planner agent trained using Behavior Cloning on \textsc{scand}. The scenario on the left shows a stationary human in the robot's path and the scenario on the right shows a human walking to the location of the robot. The robot is evaluated on social compliance and safety as it navigates to its goal position.}
    \label{fig:scand_human_trial_setup}
\end{figure} 

\begin{figure}[!th]
    \centering
    \includegraphics[width=0.5\columnwidth]{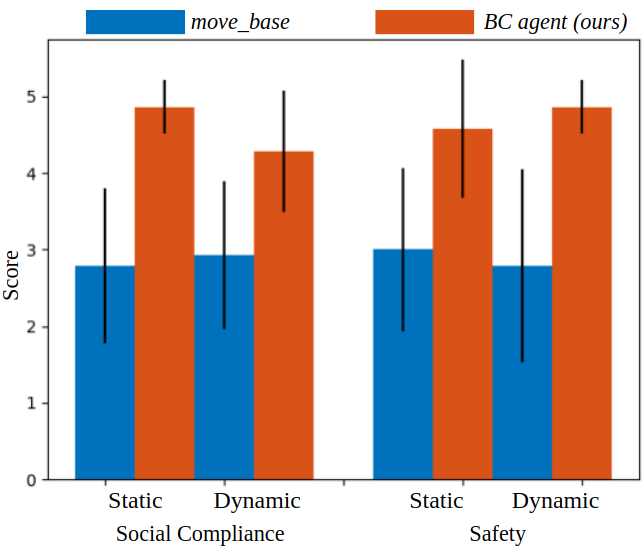}
    \caption{Mean and standard deviation of scores assigned by the fourteen human participants in the evaluation study for the learned local planner. }
    \label{fig:human_trial_metric}
\end{figure} 

\subsection{Demonstrator Classification}
\label{sec:demonstrator_classification}

 In this section, I present an analysis performed using \scand{} to evaluate if there is more than one way to navigate in a socially compliant manner. By training a neural-network-based classifier on the task of demonstrator classification, we find that there may be more than one strategy to navigate in a socially compliant manner in a given scenario. 

\noindent\textbf{Approach and Implementation:} To analyze if there is more than one way to navigate in a socially compliant manner, we choose sixteen trajectories driven by two demonstrators navigating along the same route (Speedway road within the university campus) and train a neural network for the task of demonstrator classification (training on twelve trajectories and validating on four trajectories). The input to our classifier is a ten-second-long sequence of sensor observations. This sequence consists of processed sensor observations provided in \textsc{scand} such as lidar scans (subsampled to 1 Hz and represented as grayscale bird's eye view (BEV) image), positions of the robot relative to the first lidar frame, future trajectory driven by the human consisting of 200 points in the most recent lidar frame, inertial and joystick values executed by the demonstrator at each of the lidar frames. The neural network architecture consists of a four-layer convolutional encoder to process the grayscale BEV lidar images and a three-layer fully connected network to process the other sensor observations. The representations output by these layers is fed into a three-layer fully connected network classifier head. We use the binary cross-entropy loss to train the classifier network. Figure \ref{fig:classifier_arch} shows the demonstrator classifier's inputs and neural network architecture. 

\begin{figure}[!tb]
    \centering
    \includegraphics[width=\columnwidth]{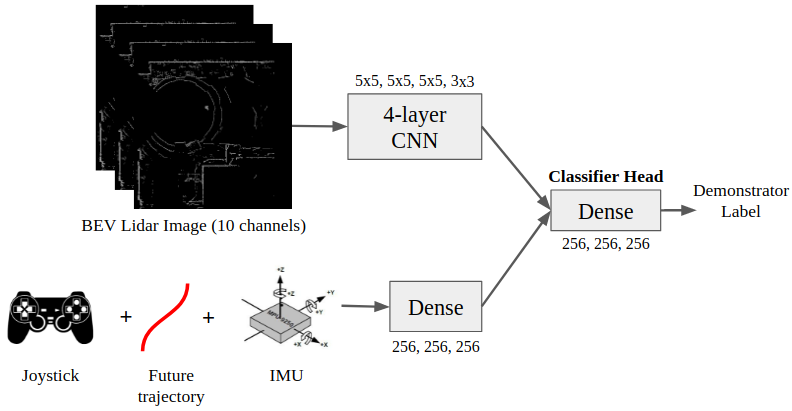}
    \caption{Network architecture and inputs for the demonstrator classifier. The classifier takes as its input ten-second-long sensor observations and demonstrator-issued joystick commands to predict the demonstrator's label. }
    \label{fig:classifier_arch}
\end{figure} 

\noindent\textbf{Results and Conclusion}
We find that the classifier is about 74\% accurate at classifying the expert on the held-out test set. Given that random guessing would lead to a success rate of 50\%, and that many ten-second trajectories do not indicate any differentiating social interactions, this number is indicative of successful prediction. 
The ability of the classifier to identify the demonstrator from their navigation style with an accuracy of 74.48\% using a ten-second sequence of observations, combined with the fact that the demonstrations in \textsc{scand} are socially compliant shows that there exists more than one strategy for socially compliant navigation in a given scenario. Enabling algorithms to take into consideration this manifold of socially compliant robot navigation behaviors naturally observed in human demonstrations is a promising direction for future work.


\subsection{Analyzing Social Compliance of Classical Navigation Algorithms}
\label{sec:analyze_soc_compliance}
In this section, I present an analysis of the social compliance of classical heuristic-based navigation algorithms on the scenarios in \scand{}. The findings demonstrate that classical navigation systems can safely and efficiently produce socially compliant navigation behaviors in a substantial proportion (up to 80\%) of navigation scenarios in \scand{}. The insights gleaned from this evaluation, coupled with earlier analyses from Sections \ref{scand_imitation_learning} and \ref{sec:demonstrator_classification} on \scand{} informs the development of an innovative hybrid approach that integrates classical and learning-based techniques for improving socially compliant robot navigation, as detailed in Section \ref{sec:hybrid_approach}.

\subsubsection{Defining Social Compliance on SCAND}

During social robot navigation in \textsc{scand}, at each time step $t$, we denote a navigation scenario $\mathcal{S}_t$ as observed by the onboard robot perception, which includes a sequence of 3D LiDAR scans $L$, RGB-D images $I$, odometry $O$, and IMU readings $U$, and a navigation goal $G$, i.e., $\mathcal{S}^{D}_t=\{L^{D}_k, I^{D}_k, O^{D}_k, U^{D}_k, G^{D}_t\}_{k=t-N+1}^t$, where $N$ denotes the history length included in the scenario at $t$ and $D$ denotes that the data is from the \textsc{scand} demonstrations. 

We further define the navigation behavior at $t$ as $\mathcal{B}_t$, which can take the form of either a global or local plan ($P_t$ or $A_t$). A demonstrated global plan in \textsc{scand}, computed as the human-driven trajectory starting from time $t$, takes the form of a sequence of waypoints $P^{D}_t = \{(x^{D}_i, y^{D}_i)\}_{i=t}^{t+M-1}$. A demonstrated local plan is represented as a sequence of joystick action commands $A^{D}_t = \{(v^{D}_i, \omega^{D}_i)\}_{i=t}^{t+K-1}$, where $v$ and $\omega$ is the linear and angular velocity respectively. $M$ and $K$ denote the length of the navigation behavior on the global and local plan level. 

Producing the navigation behavior $\mathcal{B}_t$ (i.e., $P_t$ or $A_t$) based on $\mathcal{S}_t$ as input, a navigation system is defined as a function $\mathcal{F}$ (i.e., $\mathcal{F}_g$ or $\mathcal{F}_l$ on the global or local level): $\mathcal{B}_t=\mathcal{F}(\mathcal{S}_t)$. 
We use the difference between $\mathcal{B}_t$ and $\mathcal{B}^{D}_t$, i.e., $d = \left \lVert \mathcal{B}_t-\mathcal{B}^{D}_t\right \rVert = ||\mathcal{F}(\mathcal{S}^{D}_t)-\mathcal{B}^{D}_t||$, to quantify the social compliance of the navigation system $\mathcal{F}$ on the demonstrated navigation scenarios in \textsc{scand}. 
In particular, we use the Hausdorff distance between $P_t$ and $P^{D}_t$ and L2-norm between $A_t$ and $A^{D}_t$ to evaluate global and local planning systems respectively. 

Given different social scenarios $\mathcal{S}^{D}_t$, a socially compliant navigation planner is expected to generate navigation behaviors that are similar to the expert demonstration $\mathcal{B}^{D}_t$ in \textsc{scand}. In the context of this case study, we assume that the expert demonstrations $\mathcal{B}^{D}_t$ in \textsc{scand} is the ``ground truth'' socially compliant behavior when facing $\mathcal{S}^{D}_t$. However, this assumption is not without its limitations, as highlighted in Section \ref{sec:demonstrator_classification}. It is acknowledged that there may be more than one way to navigate in a socially compliant manner in any given scenario. This variability prompts the execution of a human study to evaluate the social compliance of navigation strategies, irrespective of their alignment with \textsc{scand} demonstrations (referenced in Section~\ref{sec:hybrid_experiments}). Under this premise, we define social compliance as follows.



\textbf{\textit{Definition 1.}} In a given navigation scenario, denoted as $\mathcal{S}^{D}_t$, a navigation behavior $\mathcal{B}_t$ is classified as \textbf{socially compliant} when the deviation $d = \lVert \mathcal{B}_t - \mathcal{B}^{D}_t \rVert = \lVert \mathcal{F}(\mathcal{S}^{D}_t) - \mathcal{B}^{D}_t \rVert$ is less than a predefined small threshold $\epsilon$.

Further, let $\alpha \in [0,1]$ be the fraction of the total \textsc{scand} time steps $T$, in which $\mathcal{B}_t$ is socially compliant. We will use $\alpha$ to indicate how socially compliant a navigation system $\mathcal{F}$ is. 

\subsubsection{Classical Navigation Systems}
We study four classical navigation systems, each with accessible open-source implementations, on the \textsc{scand} social navigation scenarios $\mathcal{S}_t^D$ and compare their navigation behavior $\mathcal{B}_t = \mathcal{F}(\mathcal{S}_t^D)$ against the \textsc{scand} demonstrations $\mathcal{B}_t^D$. 

\noindent\textbf{1.\texttt{move\_base}:}
The Robot Operating System (\textsc{ros}) \texttt{move\_base}~\citep{rosmovebase} global planner utilizes a static costmap representation of the environment and Dijkstra's algorithm to generate an optimal path from the robot's current pose to the goal. The resulting path is smoothed and interpolated so that the local planner can follow. 
The \texttt{move\_base} default local planner, the Dynamic Window Approach (\textsc{dwa}) by \cite{fox1997dynamic}, operates reactively, considering the robot state, sensor information, kinematic constraints, and global path. It generates real-time linear and angular velocity commands by evaluating various trajectories within the robot's dynamic window, ensuring progress toward the goal while avoiding obstacles. 

\noindent\textbf{2. \texttt{move\_base} with social layer:}
The static costmap of \texttt{move\_base} can be augmented with a social layer with added emphasis on social factors~\cite{social_layer}, while both the global and local planners function similarly to the standard \texttt{move\_base}. The social layer employs LiDAR scans to detect people and adjusts the costmap by introducing Gaussian distributions around them, thereby incorporating their presence into the planning process.

\noindent\textbf{3. Human-Aware Planner:} \cite{kollmitz15ecmr} introduced the Human-Aware Planner, designed to facilitate polite, obedient, and comfortable robot navigation that prioritizes human presence. This approach incorporates social constraints into path planning to respect personal space and avoid close proximity to people. It employs time-dependent, deterministic planning to account for the spatial dynamics between the robot and humans over time. The planner integrates a social cost model with a layered cost map to efficiently merge social constraints with those of the static environment. Its optimization criteria include social comfort, path length, execution time, and environmental constraints.

\noindent\textbf{4. CoHAN:} 
\cite{singamaneni2021human} developed CoHAN, a human-aware navigation planner, to navigate complex and crowded indoor environments. This planner extends the Human-Aware Timed Elastic Band (\textsc{hateb}) method, as detailed by \cite{singamaneni2020hateb}, to handle dense crowds and enhance the legibility and acceptability of navigation. CoHAN integrates with the ROS navigation stack, incorporating human safety and visibility considerations into both the global and local costmap layers.

\subsubsection{Case Study Results}

All four studied classical navigation systems are kept in their default parameterizations and configurations and we observe in general all of them can produce socially compliant navigation behaviors. To assess their social compliance, at every time step $t$, we set the navigation goal $10$m ahead of the robot on the human demonstrated path. We employ Hausdorff distance as the error metric $d = \lVert \mathcal{B}_t-\mathcal{B}^{D}_t \rVert$ to compare global plan $P_t=\{x_i, y_i\}_{i=1}^{200}$ at each \textsc{scand} navigation scenario $\mathcal{S}^{D}_t$ against \textsc{scand} demonstration $P_t^D$. Figure~\ref{fig:HdistanceComparison} shows how different Hausdorff distances look like visually. We choose Hausdorff distance because most global planners in existing navigation systems only plan 2D trajectories, without robot orientation and precise temporal information. As depicted in Figure~\ref{fig:case_study} middle, with $\epsilon = 1.0$, the vanilla \texttt{move\_base} with default costmap yields the highest compliance with respect to the human demonstrations in more than $\alpha = 80\%$ of \textsc{scand} navigation scenarios; \texttt{move\_base} with social layer achieves social compliance in the lowest percentage, $\alpha = 60\%$; The Human-Aware Planner and CoHAN fall in between, producing socially compliant behaviors in approximately $\alpha=75\%$ of \textsc{scand} navigation scenarios. When increasing $\epsilon$ from $1.0$ to $3.0$, all planners are able to achieve social compliance in a larger percentage of \textsc{scand} scenarios. In the case of local planning (Figure~\ref{fig:case_study} left), where L-2 norm is utilized as $d$ to compare local plans $A_t = (v_t,\omega_t)$ against $A_t^D$, 
vanilla \texttt{move\_base} achieves $\alpha = 60\%$, followed by Human-Aware Planner, CoHAN and \texttt{move\_base} with social layer with marginal differences. The jump of the Cumulative Distribution Function (CDF) curves at around 1.6 is due to the fact that the maximal linear velocity $v$ in \textsc{scand} is roughly 1.6m/s. Note that the difference in global plans will directly affect the social compliance of the local plans. 

\begin{figure}[!t]
    \centering
    \includegraphics[width=\columnwidth]{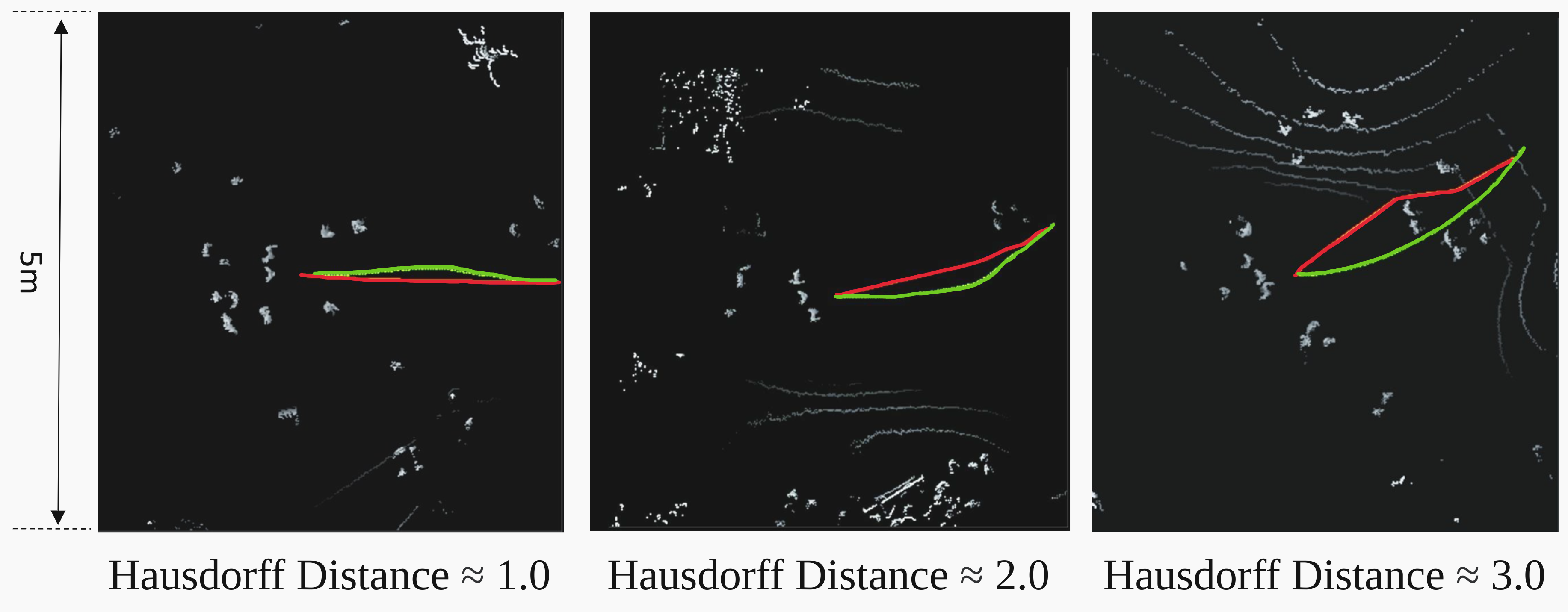}
    \caption{Different Hausdorff distances between the green and red global plan. White dots denote nearby humans and obstacles.}
    \label{fig:HdistanceComparison}
\end{figure}

\begin{figure*}[!t]
    \centering
    \includegraphics[width=\columnwidth]{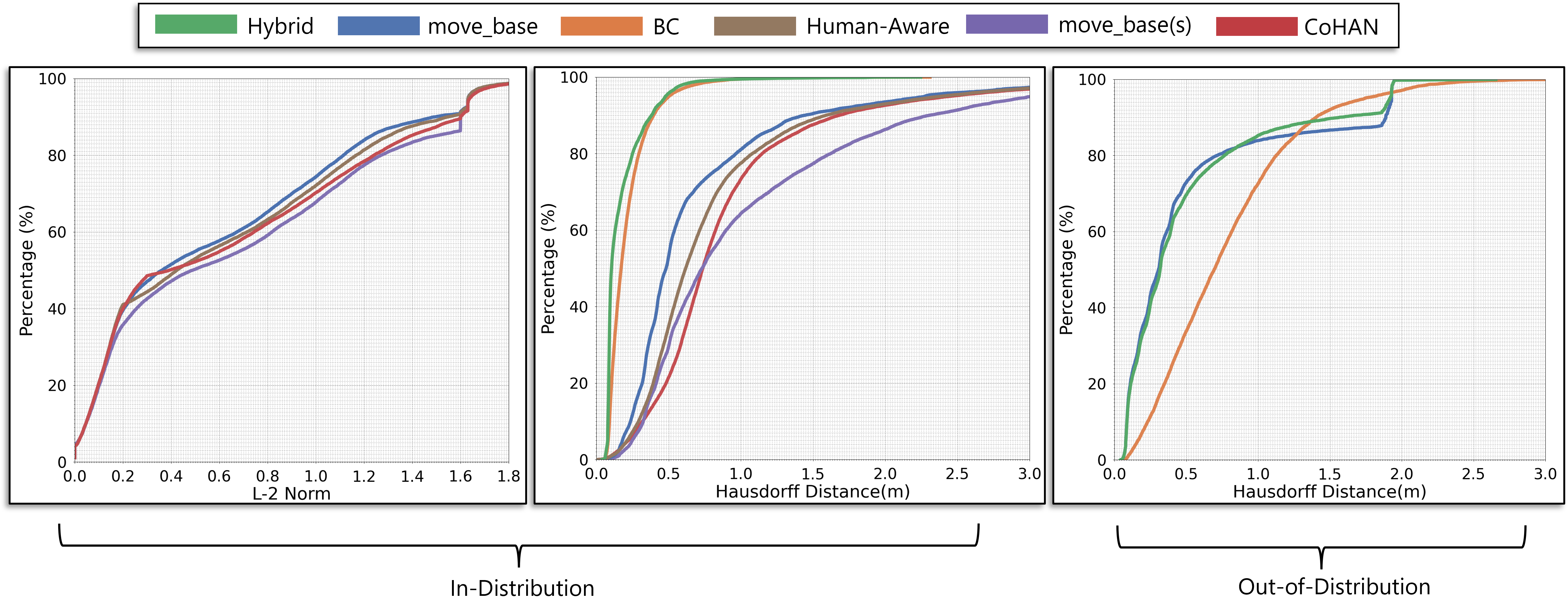}    
    \caption{Case Study Results: Cumulative Distribution Function (CDF) curves of different navigation planners compared against human demonstrations (Hausdorff distance for global planners and L-2 norm for local planners) with in-distribution (\textsc{scand}) and out-of-distribution data.}
    \label{fig:case_study}
\end{figure*}

\subsection{Anticipated Use Cases of SCAND}
\label{sec:anticipated_scand_usecases}

 Although \textsc{scand} includes a wide variety of social navigation scenarios, there may be novel interactions that are less frequent. To improve the generalizability of a learning-based approach to unseen situations, exploring representation learning for social navigation with \textsc{scand} is a promising future direction. \textsc{scand} was collected in a single city (Austin, Texas, USA) and might incorporate regional biases such as staying to the right of the road, or overtaking pedestrians from the left. This potential bias raises a need for algorithms, evaluations, and metrics for social navigation that are flexible enough to work in the presence of different local norms.

Evaluating social navigation policies is an active area of research in the navigation community \citep{soreneval, tsoi2020sean, socmomentum, biswas2021socnavbench}. While benchmarking social navigation policies is out of scope for this work, existing simulation-based navigation benchmarks such as SocNavBench by \cite{biswas2021socnavbench} that use human-only trajectories could be augmented and improved using human-robot interaction trajectories in \textsc{scand}. Similarly, another interesting future research direction is to explore Real-to-Sim transfer \citep{sim2real, garat, sgat, rgat} with \textsc{scand} and improve parameterized simulated social navigation environments to generate more realistic social interactions between virtual agents, directly benefiting data hungry approaches such as reinforcement learning.


Other directions for future work that could directly benefit from \textsc{scand} include trajectory prediction, trajectory classification, and inverse reinforcement learning for large-scale cost function learning. Previously, work on trajectory prediction and classification has used human-only \citep{humantrajpred} or robot-only \citep{robotmotionpred} trajectories, but with access to \textsc{scand}, exploring human-robot trajectories is an interesting direction for future work. 
The work by \cite{wulfmeier2017large} utilized static scenarios to learn a cost function for autonomous robot navigation using Maximum Entropy Deep Inverse Reinforcement Learning (\textsc{medirl}). Applying \textsc{medirl} on \textsc{scand} to learn cost functions that incorporate social compliance is also an interesting direction for future work.  

\section{A Hybrid Approach to Social Robot Navigation}
\label{sec:hybrid_approach}

Our social compliance case study shows that classical planners perform well in a majority of social scenarios in \textsc{scand}. However, as shown in Figure~\ref{fig:case_study}, the classical planners deviate significantly from the human demonstrations in a small percentage of the dataset. On the other hand, learning-based approaches have been shown to enable emergent navigation behaviors \citep{bojarskie2e}, yet, in certain cases, tend to overfit and do not generalize well when facing out-of-distribution scenarios. 

This observation motivates us to rethink how to take advantage of both classical and learning-based approaches. In this section, we first compare a widely used learning approach, Behavior Cloning (BC), with the best classical planner in our study, \texttt{move\_base}, and show that building a completely new learning-based social navigation planner may not work as well as the classical planner. We then introduce our new hybrid approach \citep{raj2023targeted} to leverage the best of both worlds\footnote{This work was a collaborative effort with coauthors from UT Austin and GMU and was accepted for publication at ICRA 2024 \citep{raj2023targeted}.}.


\subsection{Comparing BC with \texttt{move\_base}}
A commonly known issue for learning-based approaches is the lack of generalizability to out-of-distribution data. Therefore, in addition to training and testing on the original \textsc{scand}, we follow \textsc{scand}'s procedure and collect extra demonstrations in manually curated social scenarios (in contrast to in the wild), including intersection encounter, frontal approach, and people following. We also study the social compliance of both \texttt{move\_base} and BC on this out-of-distribution test set. 

For the original \textsc{scand}, as shown in Figure~\ref{fig:case_study} middle, the CDF curves show that BC (orange) performs better than \texttt{move\_base} (blue) on the in-distribution \textsc{scand} test set throughout the entire $\epsilon$ range, indicating that learning-based approaches can efficiently capture social compliance in in-distribution scenarios. For the out-of-distribution dataset collected to test generalizability, as shown in Figure~\ref{fig:case_study} right, while \texttt{move\_base} achieves similar performance compared to the in-distribution test data in the original \textsc{scand}, BC's performance significantly deteriorates, seriously suffering from the commonly known distribution-shift problem of learning-based approaches. Although BC's overall performance drops significantly, we can observe a trend that BC's performance overtakes \texttt{move\_base}'s performance at more challenging social scenarios, indicating that learning-based approaches have the potential to solve what classical approaches cannot solve. 


\subsection{Leveraging the Best of both Worlds}

Our case study results suggest that (1) classical navigation systems can produce socially compliant navigation behaviors in a majority of social scenarios and (2) learning-based approaches (in our case, BC) have the potential to solve challenging social scenarios where classical approaches fail to be socially compliant.
Therefore, we propose a new framework where the classical navigation system acts as a backbone, which is complemented by a learning-based approach for handling difficult social navigation scenarios. To be specific, we instantiate our hybrid navigation planner $\mathcal{F}(\cdot)$ based on a classical navigation planner $\mathcal{C}(\cdot)$, a learning-based planner $\mathcal{L}_\theta(\cdot)$ with learnable parameters $\theta$, and a gating function $\mathcal{G}_\phi(\cdot)$ with learnable parameters $\phi$ that selects between the output from the classical and learning-based planners: 
\begin{equation}
    \mathcal{B}_t = \mathcal{F}(\mathcal{S}_t) = \mathcal{G}_{\phi}(\mathcal{C}(\mathcal{S}_t), \mathcal{L}_\theta(\mathcal{S}_t), \mathcal{S}_t). 
\end{equation}

The parameters $\phi$ and $\theta$ can be learned using supervised learning on the navigation scenario and behavior tuples $\{\mathcal{S}_t^D, \mathcal{B}_t^D\}_{t=1}^T$ in \textsc{scand}: 
\begin{equation}
\begin{split}
    &\argmin_{\phi, \theta}  \sum_{t=1}^T d(\mathcal{S}_t^D), \\ 
    &d(\mathcal{S}_t^D) = \lVert \mathcal{B}_t - \mathcal{B}_t^D \rVert, \\
    &\mathcal{B}_t = \mathcal{G}_{\phi}(\mathcal{C}(\mathcal{S}_t^D), \mathcal{L}_\theta(\mathcal{S}_t^D), \mathcal{S}_t^D).
    \label{eqn::argmin}
\end{split}
\end{equation}

From among the many ways to learn $\mathcal{G}_\phi(\cdot)$ and $\mathcal{L}_\theta(\cdot)$ (either jointly or separately), in this work, we present a simple implementation that first learns a classifier $\mathcal{M}_\phi(\mathcal{S}_t^D)$ based on the difference $d$ between $\mathcal{B}_t^D$ and $\mathcal{C}(\mathcal{S}_t^D)$ to choose between $\mathcal{C}(\mathcal{S}_t^D)$ and $\mathcal{L}_\theta(\mathcal{S}_t^D)$: 
\begin{equation}
    \mathcal{B}_t = 
\begin{cases}
    \mathcal{C}(\mathcal{S}_t^D),& \text{if } \mathcal{M}_\phi(\mathcal{S}_t^D) = 1,\\
    \mathcal{L}_\theta(\mathcal{S}_t^D),& \text{if } \mathcal{M}_\phi(\mathcal{S}_t^D) = 0.
\end{cases}
\end{equation}
$\mathcal{C}(\cdot)$ can already produce socially compliant behaviors when $d \leq \epsilon$, while $\mathcal{L}_\theta(\cdot)$ only learns to address navigation scenarios where $d > \epsilon$ ($\epsilon$ is a manually defined threshold).

To be specific, by comparing $\mathcal{C}(\mathcal{S}_t^D)$ against $\mathcal{B}_t^D$, i.e., $d = ||\mathcal{C}(\mathcal{S}_t^D) - \mathcal{B}_t^D||$, we separate the original \textsc{scand} $\mathcal{D}$ into a socially compliant $\mathcal{D}^C$ and a socially non-compliant $\mathcal{D}^N$ subset with respect to $\mathcal{C}(\cdot)$, and form a supervised dataset $\{\mathcal{S}^D_t, c_t\}_{t=1}^T$, in which $c_t = 1$ if $\mathcal{S}^D_t \in \mathcal{D}^C$ ($d \leq \epsilon$) and $c_t = 0$ if $\mathcal{S}^D_t \in \mathcal{D}^N$ ($d > \epsilon$). Then, $\mathcal{M}_\phi(\cdot)$ is learned via supervised learning with a cross-entropy loss to classify whether $\mathcal{C}(\mathcal{S}_t^D)$ is socially compliant or not:  
\begin{equation}    
    \phi^* = \argmax_\phi \sum_{t=1}^T \log \frac{\exp\big(\mathcal{M}_\phi(\mathcal{S}^D_t)[c_t]\big)}{\exp{\big(\mathcal{M}_\phi(\mathcal{S}^D_t)[0]\big)}+\exp{\big(\mathcal{M}_\phi(\mathcal{S}^D_t)[1]\big)}}.
\end{equation}

The learning-based planner $\mathcal{L}_\theta$ is then learned to minimize the difference between its outputs and demonstrations in $\mathcal{D}^N$: 
\begin{equation}    
    \theta^* = \argmin_{\theta} \sum_{(\mathcal{S}_t^D, \mathcal{B}_t^D) \in \mathcal{D}^N} ||\mathcal{L}_\theta(\mathcal{S}_t^D) - \mathcal{B}_t^D||. 
    \label{eqn:bc}
\end{equation}

During deployment, $\mathcal{F}(\cdot)$ first uses $\mathcal{M}_{\phi^*}(\cdot)$ to classify if $\mathcal{C}(\mathcal{S}_t)$ is socially compliant or not, and then executes $\mathcal{C}(\mathcal{S}_t)$ if compliant or $\mathcal{L}_{\theta^*}(\mathcal{S}_t)$ if not.

\subsection{In- and Out-of-Distribution Experiment Results}
We instantiate $\mathcal{C}(\cdot)$ as a global planner using the \texttt{move\_base} planner and generate the corresponding $\mathcal{D}^N$ to train a BC planner. We apply our hybrid planner on both the original \textsc{scand} and the out-of-distribution test set. As shown in Figure~\ref{fig:case_study} middle, our hybrid approach (green) imitates a larger percentage of \textsc{scand} social scenarios with smaller Hausdorff distance in contrast to \texttt{move\_base} (blue) and BC (orange). 
For the out-of-distribution test set, as shown in Figure~\ref{fig:case_study} right, our hybrid approach does not suffer from the significant performance degradation experienced by BC and achieves similar performance as \texttt{move\_base} with small Hausdorff distance. At a large distance, our hybrid approach is able to improve upon \texttt{move\_base} and approach BC. 
Our results in Figure~\ref{fig:case_study} verify our hypothesis that the hybrid approach can take advantage of the best of both worlds facing both in-distribution and out-of-distribution social navigation scenarios. 

\subsection{Physical Experiments}
\label{sec:hybrid_experiments}


\begin{figure}[t]
    \centering
    \includegraphics[width=\columnwidth]{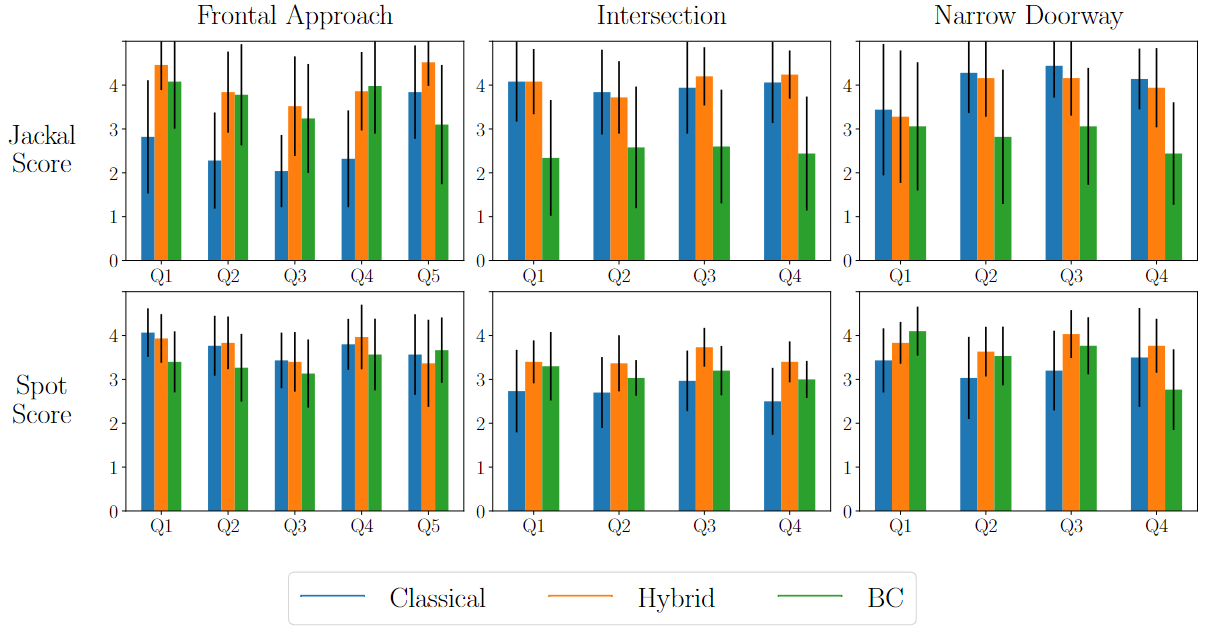}
    \caption{Average scores per question on the human evaluation study for the three navigation approaches: classical, behavior cloning, and the proposed hybrid approach.}
    \label{fig::physical}
\end{figure}

\begin{table}[t]
    \centering
    \caption{Human Study Average Scores Per Method and Scenario: Participants generally prefer the robot with the hybrid approach to the pure classical or the pure BC approach.}
    \begin{tabular}{cccc}
        \toprule
        Jackal & Frontal & Intersection & Doorway \\ 
        \midrule
        Classical & $2.66\pm0.64$ & $3.98\pm0.10$ & $\mathbf{4.08\pm0.38}$\\

        Hybrid & $\mathbf{4.04\pm0.39}$ & $\mathbf{4.06\pm0.20}$ & $3.89\pm0.36$\\

        BC & $3.63\pm0.40$ & $2.49\pm0.11$ & $2.84\pm0.25$\\

        \midrule
        Spot & Frontal & Intersection & Doorway \\ 
        \midrule
        
        Classical & $\mathbf{3.73\pm0.22}$ & $2.72\pm0.17$ & $3.29\pm0.19$ \\
        
        Hybrid & $3.70\pm0.26$ & $\mathbf{3.48\pm0.15}$ & $\mathbf{3.82\pm0.14}$ \\
        
        BC & $3.41\pm0.19$ & $3.13\pm0.12$ & $3.54\pm0.49$ \\
        \bottomrule
    \end{tabular}
    \label{tab::physical}
\end{table}

We conduct a human study in a series of physical experiments to assess the social compliance of our proposed hybrid approach, in comparison to an existing classical planner, i.e., \texttt{move\_base}, and an end-to-end learning-based method, i.e., BC trained on \textsc{scand}. The experiments are conducted using a wheeled Clearpath Jackal and a legged Boston Dynamics Spot to show the generalizability of our proposed hybrid approach to robots with different morphologies on two university campuses, George Mason University (GMU) and The University of Texas at Austin (UTA), respectively. 
We test the robots' social compliance within three distinct social scenarios, i.e., Frontal Approach, Intersection, and Narrow Doorway. 
We keep the same setup of our hybrid approach among all three scenarios. 
The three methods are randomly shuffled and repeated five times, and human participants are requested to respond to a questionnaire with 4-5 questions using Likert scales \citep{soreneval, xiao2022learning} following each run. Each scenario is tested on ten different individuals (fifteen interactions per individual).  

\subsubsection{Social Compliance Questionnaire}
For the Frontal Approach, the five assessments posed to the human participants are as follows\footnote{$^*$ denotes negatively formulated questions, for which we reverse-code the ratings to make them comparable to the positively formulated ones.}.
\begin{enumerate}
    \item \emph{The robot moved to avoid me.}
    \item \emph{The robot obstructed my path$^*$.}
    \item \emph{The robot maintained a safe and comfortable distance at all times.}
    \item \emph{The robot nearly collided with me$^*$.}
    \item \emph{It was clear what the robot wanted to do.}
\end{enumerate}

For Intersection, the four assessments were as follows. 
\begin{enumerate}
    \item \emph{The robot let me cross the intersection by maintaining a safe and comfortable distance.}
    \item \emph{The robot changed course to let me pass.}
    \item \emph{The robot paid attention to what I was doing.}
    \item \emph{The robot slowed down and stopped to let me pass.}
\end{enumerate}

For Narrow Doorway, the four assessments asked were as follows.
\begin{enumerate}
    \item \emph{The robot got in my way$^*$.}
    \item \emph{The robot moved to avoid me.}
    \item \emph{The robot made room for me to enter or exit.}
    \item  \emph{It was clear what the robot wanted to do.}
\end{enumerate}
The quantitative results of our experiments are shown in Figure~\ref{fig::physical}, where we plot the per-question average along with error bars for the three methods in each of the scenarios.

\subsubsection{Jackal Experiments at GMU}
In the experiments conducted at George Mason University (GMU) using the wheeled Jackal robot,\footnote{The research described in this section was conducted in collaboration with Xuesu Xiao, Amir Hossain Raj, and Amirreza Payandeh from George Mason University, and Zichao Hu, Rohan Chandra, Luisa Mao, Joydeep Biswas, and Peter stone from The University of Texas at Austin.} our hybrid approach demonstrated notably distinct social behaviors during frontal approach, ensuring a safe distance when passing directly opposite a pedestrian. In contrast, for the other two scenarios, our hybrid method predominantly relied on the classical planner, leading to behaviors that were similar to those of the classical approach. Conversely, Behavior Cloning (BC) displayed variability across most trials: it either failed to achieve the designated objective, as reflected by the low BC scores for the Jackal Frontal Approach Q5 and Jackal Narrow Doorway Q4 (illustrated in the top left and top right of Figure~\ref{fig::physical}, respectively), or necessitated manual intervention to prevent potential collisions with the human participant or the environment.
Across the different scenarios, we observe that our approach remains consistent by maintaining the highest average in most of the questions. For both the GMU and UT Austin experiments, we run a one-way ANOVA test on the data from each question with three groups, and the test confirms the statistical significance of the comparison at a 95\% confidence level.
We show the Jackal Frontal Approach experiment in Figure~\ref{fig::robot_exp} left as an example: the classical approach follows a trajectory which passes very close to the human; BC avoids the human but it cannot recover back to the correct trajectory and gets too close to the wall before we manually intervene; the hybrid approach reacts early by maintaining a safe distance to the human and successfully reaches the goal. 
\begin{figure}
    \centering
    \includegraphics[width=\columnwidth]
    {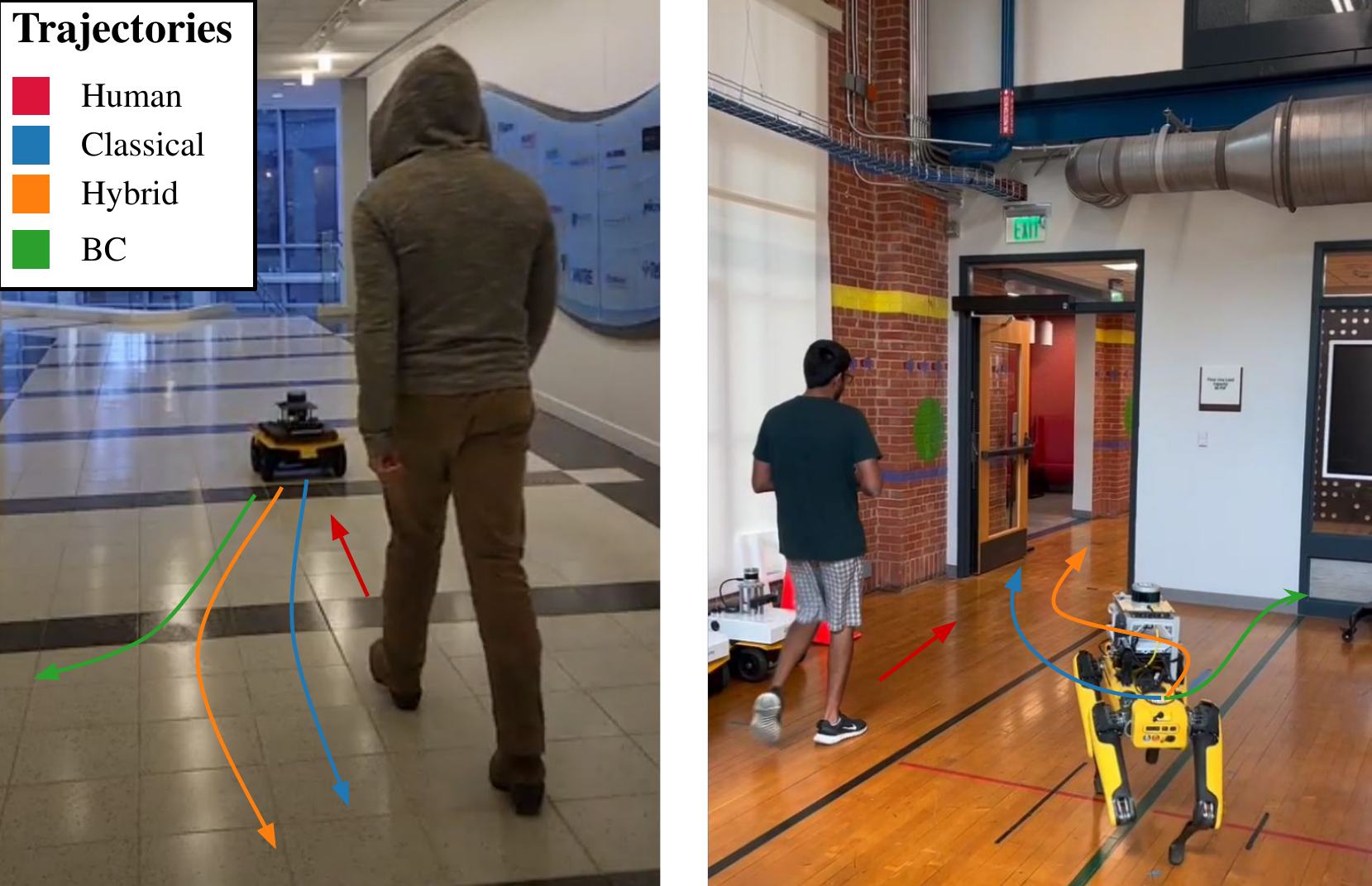}
    \caption{GMU Jackal Frontal Approach (Left) and UT Spot Narrow Doorway (Right) Robot Experiments.}
    \label{fig::robot_exp}
\end{figure}

\subsubsection{Spot Experiments at UT Austin}
We also conduct experiments with a legged Spot at The University of Texas at Austin. In general, we observe that our hybrid approach still performs the most consistently across all three scenarios. However, the classical planner's performance is slightly worse compared to the GMU Jackal experiments. It does not perform well on Spot Intersection (Figure~\ref{fig::physical} lower middle) and Spot Narrow Doorway (Figure~\ref{fig::physical} lower right, except Q4 since unlike BC, the classical approach can always reach the goal). We hypothesize that this discrepancy is caused by the different motion morphology of the Spot: legged robots are holonomic, and like humans, it is possible for them to side-step during social interaction. Not being able to do so due to the limitation of \texttt{move\_base} may cause its movement to be perceived as unnatural. BC performs slightly better for Spot in Intersection (Figure~\ref{fig::physical} lower middle) and Narrow Doorway (Figure~\ref{fig::physical} lower right). 
We show the Spot Narrow Doorway experiment in Figure~\ref{fig::robot_exp} right as an example: the classical approach first follows the shortest path until it gets close to the human and avoids the human; BC avoids the human but gets lost thereafter; the hybrid approach can slow down and avoid the human in the beginning and successfully pass the narrow doorway in the end. 

\section{Conclusion}

In this chapter, as outlined in Section \ref{sec:first_contribs}, I focus on the problem of socially compliant robot navigation in the presence of dynamic objects such as humans in the vicinity of the robot. This chapter introduced two significant contributions aimed at enhancing the social adaptability of robot navigation in such dynamic settings.

The first is a large-scale dataset of demonstrations for socially compliant robot navigation, called Socially CompliAnt Navigation Dataset (\scand{}) \citep{scand, scanddataverse}, collected by human operators teleoperating two robots in a socially compliant manner in indoor and outdoor environments within the UT Austin campus. We show through physical robot experiments in an indoor environment that applying imitation learning using behavior cloning on \scand{} leads to navigation behaviors that are perceived by human participants to be more socially compliant and safer than the heuristic-based \movebase{} navigation stack. I additionally provide several analyses on the \scand{} dataset to study social robot navigation.

The second contribution is a novel hybrid approach to improve the social compliance of robot navigation, by combining the benefits of a learning-based and heuristic-based classical navigation stacks. We show through physical robot experiments that the navigation behavior is perceived to be safe and socially compliant by several human participants in indoor environments across two different university campuses. 

The previous chapters in this dissertation introduced the \voila{}, \sterling{}, and \patern{} algorithms that utilize different forms of human feedback to learn navigation behaviors that are in alignment with operator intentions and preferences. However, they assume a static environment and do not consider dynamic objects such as humans in the scene, which are a common occurrence in the physical world where autonomous robots are increasingly being deployed. The two contributions presented in this chapter significantly advance the state-of-the-art in mobile robot navigation in dynamic environments, and together with the contributions from previous chapters, enable robot navigation that is in alignment with human intentions and preferences in both static and dynamic environments. 
\chapter{Conclusion and Future Work}
\label{chap:conclusion_and_future_work}

In this chapter, I summarize the contributions of this dissertation and additionally outline the potential avenues for future research that the contributions of this thesis open up. 


In this dissertation, I introduced several contributions, as was initially outlined in Section \ref{sec:first_contribs} that address the problem of value misalignment in autonomous mobile robots. As embodied AI agents become increasingly intelligent and autonomous due to advancements in machine learning and accelerated computing hardware, it is paramount to ensure that their internal navigation objectives and behaviors are in alignment with human intentions and preferences. This thesis therefore seeks to answer the following question:

\vspace{0.35cm}
\noindent\fbox{%
    \parbox{\textwidth}{%
       How can machine learning methods be applied to the task of autonomous navigation in unstructured environments such that the learned navigation behaviors of mobile robots align with the intentions and preferences of humans?
    }%
}

\vspace{0.75cm}
To answer this question, this thesis utilizes the learning from demonstrations and learning from preferences framework to learn navigation objectives and policies that align with the intentions and preferences of humans that are reflected in the demonstrations and preference queries. I first introduced \voila{}, an online visual imitation learning algorithm that learns from video-only demonstrations from physically different agents such as humans, overcoming the issue of egocentric viewpoint mismatch that poses a major challenge for existing visual imitation learning algorithms. I then introduced \sterling{}, a self-supervised terrain representation learning algorithm that facilitates operator terrain preference-aligned navigation in outdoor environments. Additionally, I introduced \patern{} which complements \sterling{} by using a self-supervised learning procedure to extrapolate operator preferences to visually novel terrains. All three algorithms \voila{}, \sterling{}, and \patern{} enable a mobile robot to navigate in static indoor and outdoor environments in alignment with a human operator's intentions and preferences. Finally, in the context of navigation in dynamic situations such as in the presence of humans, I introduce \scand{}, a large-scale dataset of demonstrations that facilitates imitation learning to learn socially compliant navigation policies and a novel hybrid approach for socially compliant robot navigation. I showed that learning a navigation policy through imitation learning on \scand{} leads to behaviors that are perceived to be safe and socially compliant, and additionally combining the learned policy with a heuristic-based navigation policy leads to improved performance. In all contributions, we show through physical robot experiments that the approaches enable navigation behaviors that are in alignment with the human's intentions and preferences. 

\section{Summary of Contributions}
\label{sec:summary_contribs}
As first presented in Section \ref{sec:first_contribs}, this dissertation provides the following contributions:

\noindent\textbf{1. Visual Imitation Learning for Robot Navigation:}

In Chapter \ref{chap:imitation_learning_for_robot_nav}, I introduced Visual Observation only Imitation Learning for Autonomous navigation (\voila{}) \citep{voila},\footnote{\voila{} was published in the conference proceedings of ICRA 2022, held in Philadephia, USA.} an imitation learning algorithm that learns from video-only demonstrations from a physically different expert, such as a human, overcoming egocentric viewpoint mismatch, a significant limitation of existing visual imitation learning algorithms. \voila{} addresses this problem by introducing a novel reward function that is invariant to viewpoint mismatch. The novel reward function in \voila{} uses existing local keypoint detection algorithms such as \textsc{superpoint} \citep{superpoint} that are themselves designed to be robust to viewpoint mismatches. By introducing a demonstration-dependent novel reward function that is optimized using an off-policy reinforcement learning algorithm on the robot, \voila{} addresses the imitation learning problem using reinforcement learning. On both simulated and physical robot experiments, we find that \voila{} learns to successfully imitate the expert demonstrations from physically different experts and produces navigation behaviors that are in alignment with the human operator's intentions, as demonstrated in the demonstrations.

\noindent\textbf{2. Preference-Aligned Off-Road Navigation:}

In Chapter \ref{chap:pref_learning_offroad_nav}, I introduced two contributions, focusing on the problem of aligning a mobile robot's off-road navigation behaviors with a human operator's preferences over terrains.

The first contribution is an algorithm named Self-supervised TErrain Representation LearnING (\sterling{}) \citep{sterling, sterlingicra} \footnote{\sterling{} was published in the conference proceedings of CoRL 2023, held in Atlanta, USA.}, that focuses on self-supervised learning of terrain preferences followed by query-based operator preference learning to enable preference-aligned off-road navigation. Prior approaches require large-scale human annotated datasets that may be difficult to curate or require diverse and constrained human demonstrations in the field that may not always be feasible to collect. \sterling{} overcomes these challenges by learning relevant representations of terrains in a non-contrastive, self-supervised manner by leveraging unconstrained trajectories of multi-modal sensor information recorded on the robot. Through physical robot experiments on the Spot and Jackal robots, we observed that in comparison to several state-of-the-art baseline methods, \sterling{} enables operator preference-aligned off-road navigation. 

The second contribution is an algorithm named Preference extrApolation for Terrain-awarE Robot Navigation (\patern{}) \citep{paternarxiv, paternicra} \footnote{\patern{} is accepted for publication at the ICRA 2024 conference.}, that focuses on extrapolating operator preferences from known to visuall novel terrains. A mobile robot operating autonomously in off-road terrains might encounter terrains that are outside the known distribution of terrains seen during training. In such scenarios where the robot is faced with unfamiliar terrain, the standard approach is to query a human operator for their preferences in any form \citep{kahn2021land, vrlpap, sterling} and update the robot's navigation behavior offline to handle the novel terrain. However, this may be infeasible in many scenarios where requesting feedback in real-time from the operator may be delayed or impossible. \patern{} addresses this problem by extrapolating operator preferences from known to visually novel terrains. In \patern{}, we posit that in certain scenarios, where the novel terrain is visually dissimilar but similar to a known terrain in its non-visual observation space such as inertial, proprioceptive, or tactile signals, it may be possible to extrapolate operator preferences from known to visually novel terrains. \patern{} leverages this intuition of utilizing multi-modal terrain observations and uses a self-supervised procedure to update the terrain preferences. Through physical robot experiments in diverse environments within the UT Austin university campus and on the 3-mile Ann and Roy Butler trail in Austin, TX, we observed that \patern{} successfully extrapolates operator preferences on novel terrains and leads to behaviors that are compliant with an operator's preferences of terrains. 

\noindent\textbf{3. Socially Compliant Robot Navigation:}

In Chapter \ref{chap:socially_compliant_nav}, I focus on the problem of socially compliant robot navigation. An autonomous mobile robot operating in the physical world might often encounter dynamic scenarios such as human-occupied environments during which it needs to act in a socially compliant manner to ensure the safety of both the human and the robot, and to improve the acceptance of service robots in human-occupied environments. Towards this end, in this chapter, I introduced two contributions---a large-scale dataset of demonstrations for socially compliant robot navigation, and a hybrid approach for socially compliant robot navigation.

The first contribution of this chapter is a large-scale dataset of demonstrations for socially compliant robot navigation. Previous datasets for social navigation \citep{jackrabbot} focus on the perception problem of robot navigation, such as visual detection of pedestrians and behavior prediction of moving humans. Addressing the lack of datasets to address the navigation component of socially compliant robot navigation, we introduce the Socially CompliAnt Navigation Dataset (\scand{}),  \footnote{\scand{} was published in the proceedings of RA-L 2022 and was presented at IROS 2022, held in Kyoto, Japan.} a large-scale dataset of socially compliant robot navigation. \scand{} consists of 138 trajectories, 25 miles of human-teleoperated demonstrations of a robot navigating in a socially compliant manner. We additionally found that using \scand{} to learn a navigation policy using behavior cloning \citep{behaviorcloning} leads to navigation behaviors that are perceived to be safer and socially compliant than a classical navigation stack \citep{rosmovebase}, as evidenced by a human participant study in an indoor environment. 

The second contribution of this chapter is a novel hybrid algorithm to enable socially compliant navigation behaviors on mobile robots. \footnote{The hybrid approach is accepted for publication at the ICRA 2024 conference.} Through analysis of \scand{}, we observed that a classical heuristic-based navigation stack, preferred in many robots for real-world scenarios due to its inherent safety guarantees, achieves social compliance in 80\% of scenarios in \scand{}. Additionally, while a learning-based planner learned using the \scand{} dataset can handle special scenarios that require social compliance, it lacks the safety guarantees of classical methods, which prevents its wide adoption practically. To bridge this gap, we propose a novel hybrid approach that combines the benefits of learning-based and heuristic-based approaches to robot navigation. This approach uses a learned classifier that switches between a learning-based and heuristic-based navigation stack as appropriate, trained using the \scand{} dataset. On physical robot experiments involving human participants conducted on two university campuses, we found that our hybrid approach is preferred by humans over classical or learning-based navigation stacks. 

Figure \ref{fig:flowchart_thesis} includes a flowchart summarizing the thesis question, problems, and their respective solutions proposed as contributions in individual chapters in this dissertation. 


\section{Future Work}

The work presented in this dissertation presents several avenues for future work. This section is organized as follows. Section \ref{sec:future_work_summary} summarizes and expands on the ideas already presented for future work in previous chapters. In Section \ref{sec:foundationmodels_future_work}, I discuss how recently successful large multi-modal foundation models trained with internet-scale data may be utilized for robot navigation, enabling alignment with operator preferences and intentions.

\subsection{Summary of General Ideas for Future Work}

In this section, I elaborate on ideas for future work that were briefly presented in earlier chapters of this dissertation. 
\label{sec:future_work_summary}

\subsubsection{Planner Parameter Learning Using VOILA}

The \voila{} approach introduced in Chapter \ref{chap:imitation_learning_for_robot_nav} demonstrates successful visual imitation learning capabilities for navigation, using a parameterized neural network as the policy. However, as discussed in Chapter \ref{chap:socially_compliant_nav}, end-to-end learned navigation policies represented as neural networks lack safety guarantees, a characteristic feature of classic heuristic-based navigation stacks. A potential avenue for future research is the integration of a parameterized classical navigation stack, such as \movebase{}, in the role of the parameterized policy, as was previously explored in \textsc{appld} \citep{xiao2020appld}. Adopting this strategy could potentially enable visual imitation learning of behaviors from physically different agents while retaining the safety guarantees inherent in classical navigation methodologies.

\subsubsection{Real-to-Sim and Sim-to-Real for Social Navigation}

Leveraging simulators to learn socially compliant robot navigation using reinforcement learning is a promising alternative to imitation learning. Simulators offer the flexibility of generating synthetic data at more than real-time rates and allow orchestrating several multi-agent scenarios that may be hard to capture or observe in the real world. Additionally, simulators offer access to ground-truth labels of simulated assets, such as locations of humans in the scene, which may not be readily available from data collected in the real world. Moreover, simulators can serve as a standard benchmarking tool to evaluate several social navigation algorithms, a major challenge in the social navigation community \citep{francis2023principles}.

However, existing simulators lack dynamic motion models that accurately model the navigation behavior of humans in the scene, thereby constraining their utility for online learning and benchmarking. Existing simulators typically employ either the social force model \citep{socialforce} or replay recorded trajectories of humans without closed-loop control \citep{biswas2021socnavbench}, which may not be an efficient way to evaluate all reactive social navigation algorithms. The physical world consists of a vast amount of social navigaton scenarios that are not modeled currently in existing social navigation simulators. A promising idea is to utilize the \scand{} dataset and other sources of data, such as YouTube videos, to learn better human motion models and diverse social scenarios. Below, I outline two research directions, categorized as short-term and long-term challenges.

\noindent\textbf{Short-Term Future Work:} The \scand{} dataset introduced in Chapter \ref{chap:socially_compliant_nav} offers numerous observations of human interactions, captured through lidar scans and camera data. These can serve as a foundation for developing human motion models akin to those proposed by \cite{hirose2023sacson}. Although scand lacks direct labels of human positions in the scenes, one could employ large-scale pretrained vision-language models like Segment Anything \citep{kirillov2023segany} to infer human positions from visual data. This positional information may then be used to train a human motion prediction model, enhancing the fidelity of human movement simulation for more effective online learning and benchmarking of social navigation algorithms.

\noindent\textbf{Long-Term Future Work:} Although \scand{} offers a substantial amount of data on social interactions involving a mobile robot and humans in various environments, it may not encompass all forms of social interactions, especially those that are rarer or less represented in \scand{}. A valuable direction for future research could be to explore the use of large internet-scale data sources, like YouTube. This platform likely hosts a more diverse range of scenarios depicting social compliance. The primary challenge here lies in identifying and extracting specific instances of socially compliant navigation among autonomous entities such as humans from YouTube videos. Additionally, developing methodologies for effectively leveraging these passive demonstrations to enhance social navigation skills or to recreate the scenario in simulation remains an area to explore. In this context, the approach by \cite{Fan2022minedojo} in learning diverse open-ended agent abilities from Minecraft gameplay videos on YouTube offers a noteworthy precedent.

\subsubsection{Imitating Multi-Modal Social Navigation Demonstrations}
\label{sec:multimodal_imitation_learning}

In this section, I describe two short-term directions for future work, including promising initial experimental results, on two ideas: learning a cost map using Maximum Entropy Deep Inverse Reinforcement Learning (\medirl{}) \citep{wulfmeier2017large, zhu2020off} and using Diffusion Policies \citep{chi2023diffusion} to learn a multi-modal motion planning policy for social navigation. 

\noindent\textbf{Deep Maximum Entropy IRL for Social Cost Function Learning:}
\label{sec:maxentsection}
 In this section, I motivate learning a costmap using the Maximum Entropy Inverse Reinforcement Learning approach (\medirl{}), as was initially proposed by \cite{meirl} and extended using a non-linear fully convolutional network as the cost function by \cite{wulfmeier2017large}. In comparison to policy learning using the mean squared error behavior cloning loss, learning a costmap using the maximum entropy IRL \citep{ziebart2010modelingthesis} takes a distribution matching approach, treating the demonstrations as a distribution, effectively handling multi-modality of the social navigation problem. Initial experiments on held-out data on the \scand{} dataset show promising results in learning a costmap for social navigation, that may be extended to incorporate the time-dependence of the demonstrations to fully address the social navigation problem.

\begin{figure*}
    \centering
    \includegraphics[width=\columnwidth]
    {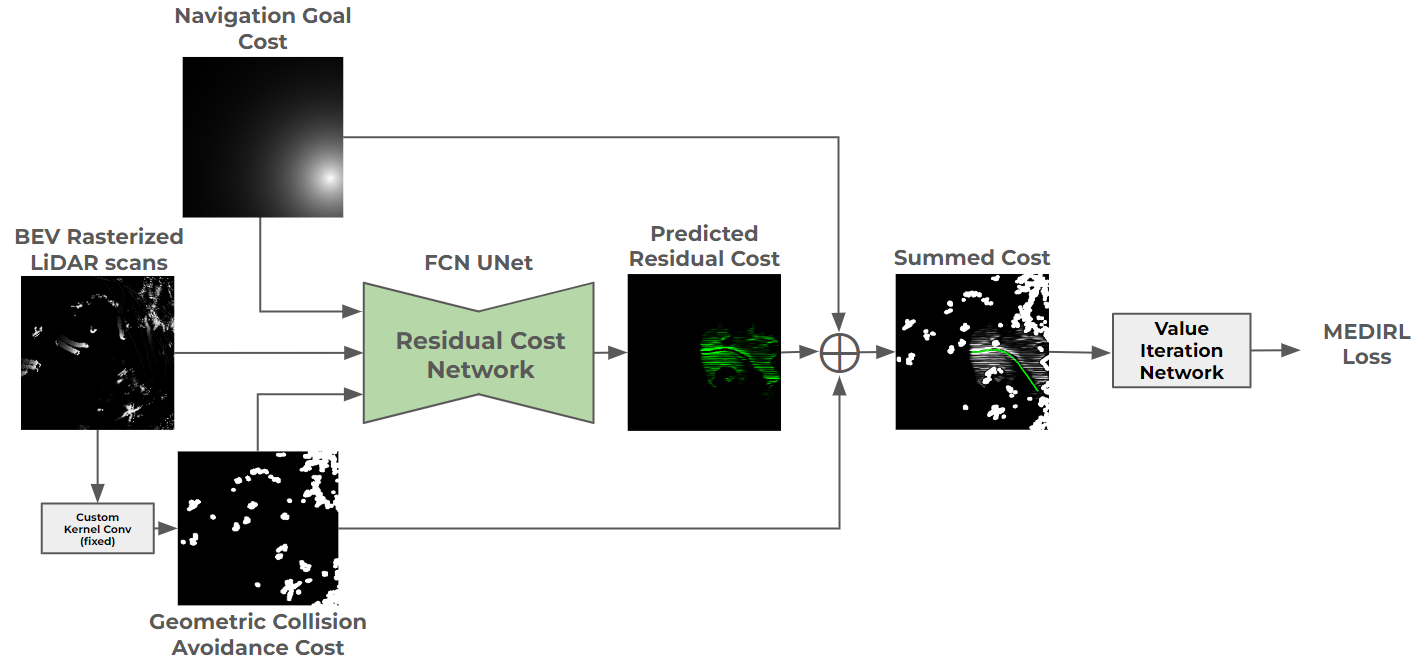}
    \caption{Network architecture and inputs for the residual cost function learning network, trained using \medirl{}. The residual cost function network takes as input the bird's eye view (BEV) rasterized lidar scans with temporally aggregated observations, the most recent lidar scan is processed with a static kernel to create the obstacle inflation representing the geometric collision avoidance cost, and the goal cost is the inverse of the exponential distance in euclidean space to the local goal location ten meters away. The residual cost function network is parameterized using a fully convolutional network similar to \cite{wulfmeier2015maximum} and uses a Tanh output-activation function to predict residual cost values, which when added to the obstacle avoidance cost and the goal-reaching costs, produces the summed costmap. To compute the \medirl{} loss to update the weights of the residual cost function network using gradient descent, a value iteration network is utilized to independently perform the forward and backward passes as initially proposed by \citep{zhu2020off}.}
    \label{fig:medirl_arch}
\end{figure*}


Figure \ref{fig:medirl_socnav} presents the residual social cost predicted by the cost function network on a held-out validation sample, as detailed in Figure \ref{fig:medirl_arch}, learned using the \medirl{} approach. Notably, in a scenario where groups of people are situated ahead of the robot, the residual cost function network assigns a higher cost to the areas surrounding the group. This cost assignment effectively deters the robot from traversing through the crowd, guiding it instead along a path consistent with the human demonstrations observed in the \scand{} dataset. This example demonstrates the potential of learning residual costs using \medirl{}, highlighting it as a promising direction for future research.

\begin{figure*}
    \centering
    \includegraphics[width=\columnwidth]
    {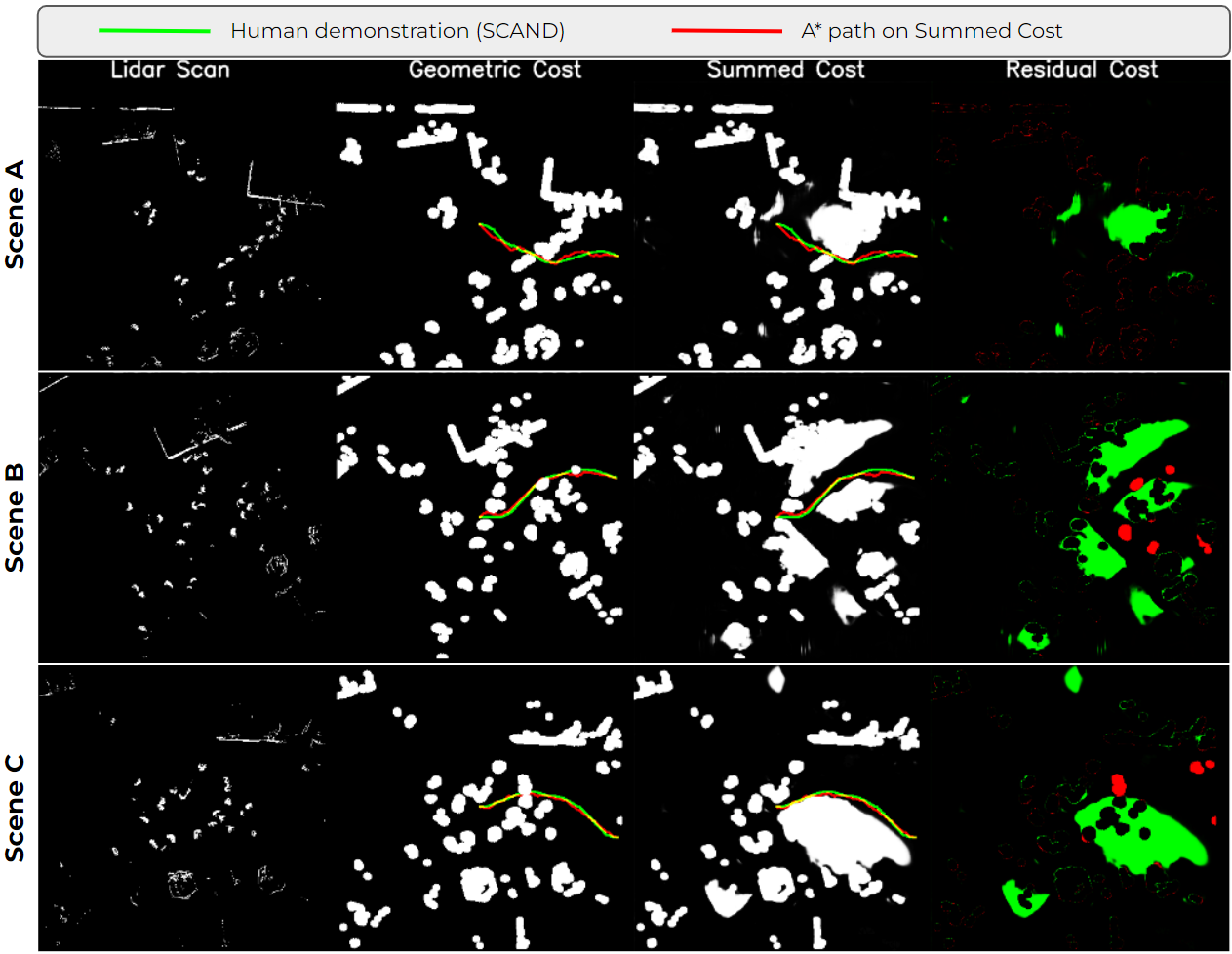}
    \caption{Experiments on a held-out validation set from \scand{} depicting three scenes, and the output of the residual cost function network, trained using Maximum Entropy Deep Inverse Reinforcement Learning \citep{wulfmeier2015maximum, zhu2020off}. Shown in each row are the lidar scan, geometric cost with an inflation radius of 0.25 meters, the summed cost, and the residual cost output by the network, respectively. Positive cost units in the residual costmap are represented in green and negative values are represented in red. The human-demonstrated path (in green) and planned path (in red) using $A^*$ on the summed cost are overlayed on both the geometric cost and the summed cost. We see that the residual costmap predicted by the network leads to paths that are close to the socially compliant path as demonstrated by the human in \scand{}. }
    \label{fig:medirl_socnav}
\end{figure*}


A key limitation of this IRL approach described above is the presumption that the residual social navigation cost $\mathcal{L}_{social}$ (social) remains static. We employ a conventional planner, such as $A^*$, that operates on static costmaps to plan the robot's future positions in the cumulative costmap $\mathcal{L}_{nav}$ for a specific environment. However, this static framework may not adequately represent the dynamic nature of social navigation, particularly in scenarios where humans are moving within the scene, necessitating dynamic, time-dependent costs for $\mathcal{L}_{social}$. An interesting future work to explore in the short term is to extend the \medirl{} approach to learn dynamic costmaps, similar to the work by \cite{lee2022spatiotemporal}.

\noindent\textbf{Diffusion Policy for Social Navigation:}
\label{sec:diffusionpolicysection}
In this section, I investigate preliminary short-term future work of leveraging diffusion as a promising generative modeling approach to learn multi-modal local motion planning policies for social robot navigation. In the domain of language-guided image generation, diffusion has been transformative with successful methods such as \textsc{dall-e} \citep{dalle} generating superior-quality images than its predecessors. The recent diffusion policy paper by \cite{chi2023diffusion} showed the successful application of diffusion to learn a local motion planning policy for robot manipulation tasks. Diffusion is a promising alternative to Gaussian mixture models \citep{jaquier2020learninggmm} in capturing the multi-modality of demonstrations, overcoming mode-collapse, and capturing diverse modes of data, as evidenced in prior work \citep{chi2023diffusion}. 

To evaluate simple behavior cloning, behavior cloning with Gaussian mixture models, and behavior cloning with diffusion policy, we consider a social navigation scene involving a T-junction interaction, where the robot and the human stand facing each other at an open intersection, and walk towards a goal 10 meters ahead, their paths intersecting at a common point, as shown in Figure \ref{fig:simulated_t_junc}. We specifically picked this scenario since there is more than one way to navigate in a socially compliant manner. For instance, the robot could either yield for the human to pass by pausing momentarily, or not yield and expect the human to yield, thereby walking towards its goal without yielding. Using the simple behavior cloning approach that does not model the multi-modality of the target demonstration distribution, we expect the approach to learn the mean of the two modes, resulting in a suboptimal behavior. Using a Gaussian Mixture model on the other hand is known to suffer from mode-collapse and getting stuck in local minima, whereas diffusion has been known to learn consistent behaviors, effectively capturing multi-modality in the demonstrations. 

\begin{figure*}[!ht]
    \centering
\includegraphics[width=\columnwidth]
    {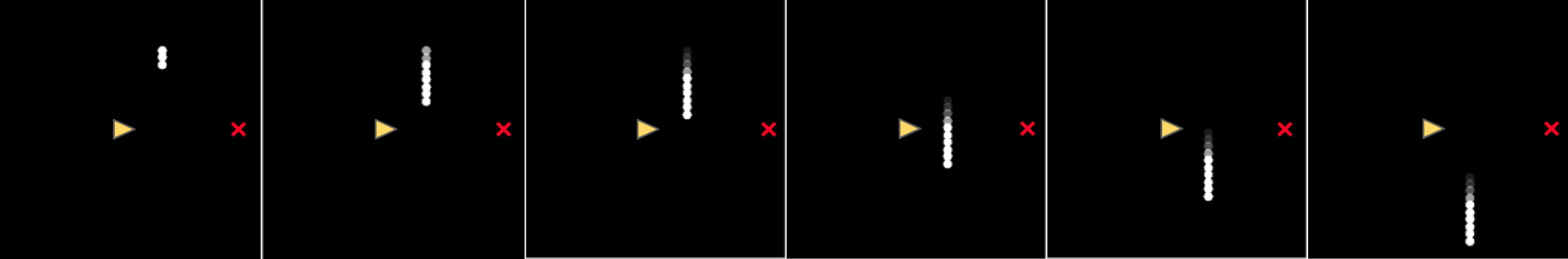}
    \caption{Depiction of the T-junction interaction scenario involving a robot and a human in the scene. The robot is shown in yellow and is standing facing its goal location 10 meters in front, marked in red. The human starts to the left of the robot and traverses in a straight line.}
    \label{fig:simulated_t_junc}
\end{figure*}

Figure \ref{fig:diffusion_expts} shows the results of the three algorithms, trained on the training set in \scand{}, deployed here on the T-junction scenario involving a dynamic human walking across the robot. In this scene, the human does not yield to the robot, and one of the optimal, socially compliant behaviors to prevent collisions or obstructing the human's motion is to pause and let the human pass. In Figure \ref{fig:diffusion_expts}, we see that the diffusion policy leads to such behaviors where the path predicted by the learned motion planner constricts the robot's motion in the future, waits for the human to pass, and then proceeds to move towards the goal. This preliminary experiment on the T-junction scenario shows that social scenarios that are relatively hard to solve with simple behavior cloning may be solved now through improved methods, such as diffusion \citep{chi2023diffusion}. An interesting direction for future work is to include other social navigation datasets, such as MuSoHu by \cite{nguyen2023toward}, and explore input data augmentations during training of the diffusion policy to handle diverse scenes. 

\begin{figure*}
    \centering
\includegraphics[width=\columnwidth]
    {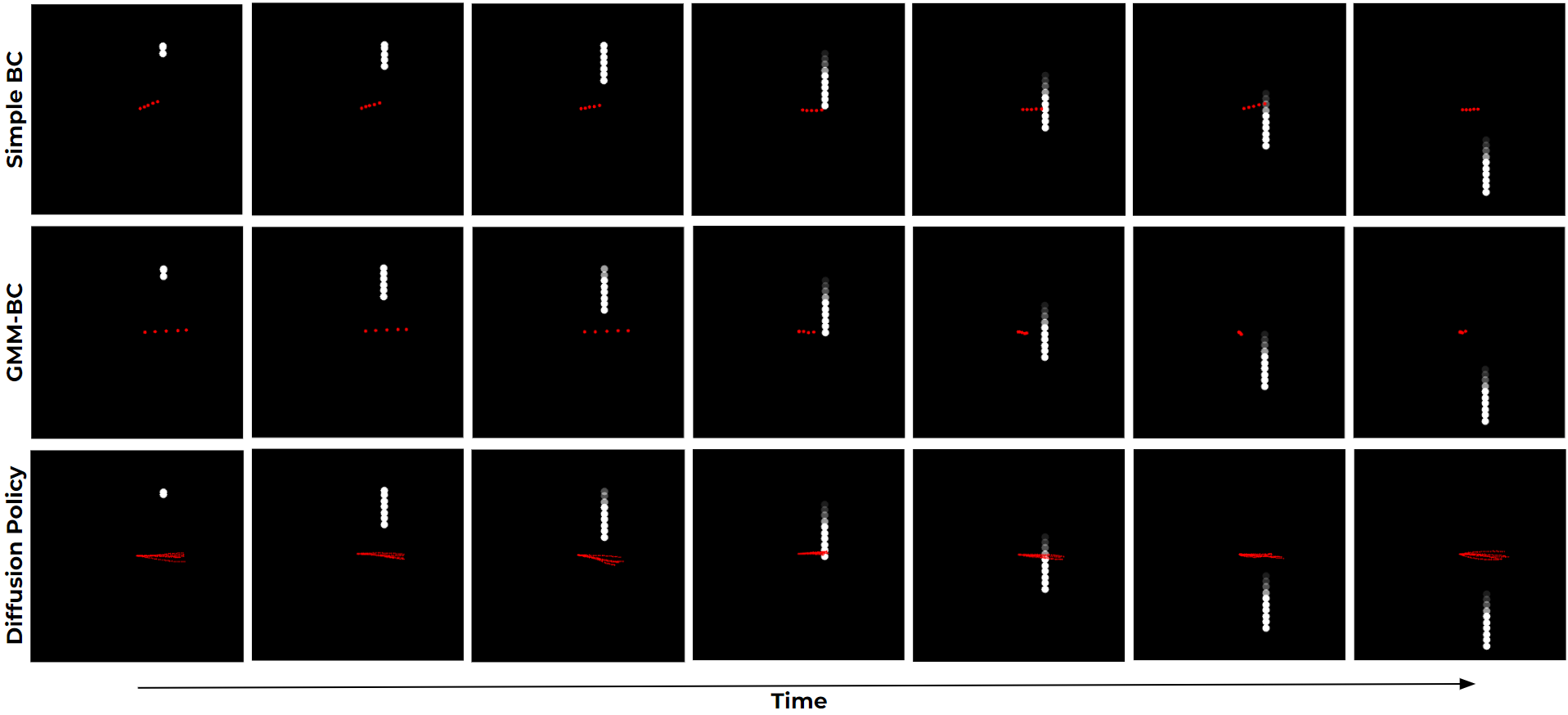}
    \caption{Initial experiments on the T-junction interaction scene comparing simple behavior cloning (SimpleBC) trained using the mean-squared error loss \citep{behaviorcloning}, behavior cloning with a gaussian mixture model (GMM-BC) with negative log-likelihood loss \citep{jaquier2020learninggmm} and Diffusion Policy \citep{chi2023diffusion} on the task of predicting a socially compliant motion path for a mobile robot, trained on the \scand{} dataset. We see that simpleBC does not predict a straight path to the goal initially. In comparison, while GMM-BC predict a straight path, and does yield for the human, it fails to predict a straight path after the human has passes, suggesting that it suffers from mode-collapse. Diffusion Policy overcomes the limitations of SimpleBC and GMM-BC by predicting a straight path when the human is far away, and yielding to the human and resuming navigation appropriately once the human passes, leading to socially acceptable behavior.}
    \label{fig:diffusion_expts}
\end{figure*}

\subsubsection{Learning to Balance Navigation Costs from Demonstrations}

In the off-road navigation work presented in Section \ref{chap:pref_learning_offroad_nav}, we utilized a sampling-based motion planner that greedily picks the optimal trajectory arc, guided by an objective function integrating two distinct costs: geometric and preference-based. During the physical robot experiments in Sections \ref{sec:experimental_results_sterling} and \ref{sec:expts_patern}, we manually adjusted the coefficient that balances these costs, a process that necessitates specialized human expertise. An interesting idea for future work in the short-term is to identify the ideal cost-balancing coefficient through imitation learning from expert demonstrations, a less cognitively demanding form of human feedback, thereby enabling the alignment of off-road behavior with operator intentions.

\subsection{Utilizing Foundation Models in Robot Navigation}
\label{sec:foundationmodels_future_work}
In this section, I present several short-term ideas for future work that expand upon the contributions of this dissertation, by utilizing recently successful multi-modal foundation models trained with internet-scale data.  

\subsubsection{Scene Description-based Visual Alignment}

Unlike \voila{}, rather than estimating reward feedback by matching video-only demonstrations and imitator observations, an alternative approach for aligning with the operator's intentions involves computing rewards through the correlation of language descriptions of the scene between the imitator and the expert's experience. To facilitate imitating video-only demonstrations from physically different experts, \voila{} utilized a demonstration-dependent reward function that was optimized using reinforcement learning. \voila{} enables alignment by matching the imitator and expert observations using a local feature detector that is robust to viewpoint mismatches and then maximizing the match density. Recent advancements have seen large-scale multi-modal models, especially vision-language models \citep{radford2021learningclip, li2023blip, gemini2023}, excel in reasoning by integrating visual scene observations with corresponding natural language descriptions \citep{yukefoundation, embodimentcollaboration2023open, bommasani2022opportunities, touvron2023llama}. A related approach that does not involve online learning, but utilizes vision-language models such as \textsc{blip}v2~\citep{li2023blip} was explored by \cite{yokoyama2023vlfm}. Combining \voila{}'s reinforcement-learning-based approach for visual imitation learning with the emerging developments in large-scale multi-modal vision-language models is an interesting research trajectory for future investigations.  

\subsubsection{Application of Foundation Models to Off-Road Navigation}

Foundation models, characterized by their extensive parameterization and trained on massive datasets, offer a versatile backbone for diverse downstream tasks. In the domain of off-road navigation, recently, there have been several datasets made publically available that contain robot experience collected in diverse off-road environments \citep{rugd, jiang2021rellis, jiang2020rellis3ddataset, triest2022tartandrive}. An interesting direction to explore is to either pre-train a large-scale foundation model or fine-tune an existing model that can aid in visual terrain awareness of a mobile robot's navigation behavior on several downstream tasks such as preference-aligned navigation \citep{sterling, patern} or high-speed off-road driving \citep{viikd}. 

Historically, self-supervised learning in off-road navigation has been narrowly focused on task-specific representations. However, the \sterling{} approach represents a significant advance, offering general, task-agnostic terrain representations through non-contrastive self-supervised learning. This approach could serve as an effective pretraining strategy for a foundation model geared toward off-road navigation. Currently, existing vision-encoder models, including \sterling{}, are limited to a handful of terrains and lack training on extensive, internet-scale datasets. By pretraining a foundation model for terrain-aware navigation, we could vastly improve adaptability across various terrains, making it instrumental in a range of off-road applications such as agriculture, high-speed racing, and search and rescue missions.

\subsubsection{Application of Foundation Models to Social Navigation}

To enable autonomous mobile robots to navigate in a socially compliant manner, they must effectively perceive their surroundings, infer the goals and potential intentions of nearby agents such as humans and other robots, and balance their objectives with those of others, employing communication when required to generate navigation behaviors that are perceived to be socially compliant. Recent foundation models, such as LLaVA \citep{liu2023improvedllava}, have exhibited remarkable performance in perception and language-based scene comprehension, essential for everyday tasks. Utilizing these large-scale foundation models pretrained from internet-scale data to bootstrap from and perform scene understanding and human intent recognition, one of the critical features necessary for social navigation, represents a promising avenue for future work.

\section{Concluding Remarks}

The increasing deployment of autonomous mobile robots in human-centric environments underscores the critical need for their navigational objectives and behaviors to be in alignment with those of their human operators. Learning from human demonstrations and human preference feedback is a powerful tool in imitation learning to address the value alignment problem. In this dissertation, I introduced several novel machine-learning algorithms and a large-scale dataset to align the behaviors of mobile robots with the intentions and preferences of humans. The findings demonstrate a substantial improvement in aligning robot navigation behaviors with human intentions and preferences, marking a significant advancement in the field of mobile robot navigation.

%


\chapter*{Acronyms}
\begin{acronym}[longest acronym here]
\acro{AMRL}{Autonomous Mobile Robotics Laboratory}
\acro{BC}{Behavior Cloning}
\acro{CNN}{Convolutional Neural Networks}
\acro{DNN}{Deep Neural Networks}
\acro{DQN}{Deep Q Networks}
\acro{EM}{Expectation Maximization}
\acro{FCN}{Fully Connected Networks}
\acro{GAIL}{Generative Adversarial Imitation Learning}
\acro{GAIfO}{Generative Adversarial Imitation from Observation}
\acro{GAN}{Generative Adversarial Networks}
\acro{GANAV}{Group-wise Attention for Navigation}
\acro{GARAT}{Generative Adversarial Reinforced Action Transformation}
\acro{GAT}{Grounded Action Transformation}
\acro{GMU}{George Mason University}
\acro{ICRA}{International Conference on Robotics and Automation}
\acro{IL}{Imitation Learning}
\acro{IRL}{Inverse Reinforcement Learning}
\acro{IROS}{International Conference on Intelligent Robots and Systems}
\acro{JS}{Jensen-Shannon}
\acro{KL}{Kullback Leibler}
\acro{LAND}{Learning to Navigate from Disengagements}
\acro{LARG}{Learning Agents Research Group}
\acro{LfD}{Learning from Demonstrations}
\acro{LfP}{Learning from Preferences}
\acro{MDP}{Markov Decision Process}
\acro{MLE}{Maximum Likelihood Estimate}
\acro{POMDP}{Partially Observable Markov Decision Process}
\acro{PPO}{Proximal Policy Optimization}
\acro{RCA}{Ride Comfort Aware Navigation}
\acro{RA-L}{Robotics and Automation Letters}
\acro{RL}{Reinforcement Learning}
\acro{SAC}{Soft Actor Critic}
\acro{TPIL}{Third Person Imitation Learning}
\acro{TRPO}{Trust Region Policy Optimization}
\acro{UTA}{The University of Texas at Austin}

\end{acronym}
\bibliographystyle{plainnat}  
\bibliography{references}        




\end{document}